\documentclass[twoside,11pt]{article}

\usepackage{jmlr2e}

\usepackage{tikz}
\usetikzlibrary{matrix,chains,positioning,decorations.pathreplacing,arrows}
\usetikzlibrary{shapes, snakes, positioning, arrows,calc}
\usepackage[utf8x]{inputenc} 

\usepackage{algorithm}
\usepackage{algpseudocode}

\usepackage{wrapfig}
\usepackage{dblfloatfix}

\usepackage{caption}
\usepackage[labelformat=simple]{subcaption}

\usepackage{hyperref}
\usepackage{url}
\usepackage{amsmath}
\usepackage{amsfonts}
\usepackage{dsfont}
\usepackage{bm}
\usepackage{xcolor}
\usepackage{cleveref}
\usepackage{cases}

\usepackage{etoolbox}
\usepackage{nicefrac}

\newcommand{\mbf}[1]{\mathbf{#1}}
\newcommand{\zbhat}{\mbf{\hat{z}}}
\newcommand{\zb}{\mbf{z}}

\newcommand{\mub}{\bm{\mu}}

\newcommand{\lambdab}{\bm{\lambda}}
\newcommand{\lambdabhat}{\bm{\rho}}

\newcommand{\ub}{\mbf{u}}
\newcommand{\lb}{\mbf{l}}
\newcommand{\xb}{\mbf{x}}
\newcommand{\tb}{\mbf{t}}

\newcommand{\xbhat}{\mbf{\hat{x}}}

\newcommand{\sig}[1]{\sigma\left(#1\right)}
\newcommand{\phiub}{\phi_{\hat{\ub}_k, \hat{\lb}_k}}
\newcommand{\psiub}{\psi_{\hat{\ub}_k, \hat{\lb}_k}}
\newcommand{\ulslope}{\frac{\sig{\hat{\ub}_k}-\sig{\hat{\lb}_k}}{\hat{\ub}_k - \hat{\lb}_k}}
\newcommand{\tlslope}{\frac{\sig{\tb_k} - \sig{\hat{\lb}_k}}{\tb_k - \hat{\lb}_k}}
\newcommand{\auglag}{\mathcal{L}\left( \xbhat,\lambdabhat\right)}
\newcommand{\usigmab}{\bar{\bm{\sigma}}}
\newcommand{\lsigmab}{\ubar{\bm{\sigma}}}
\newcommand{\pib}{\bm{\pi}}
\DeclareMathOperator*{\argmax}{\arg\!\max}
\DeclareMathOperator*{\argmin}{\arg\!\min}

\usepackage{accents}
\newcommand{\ubar}[1]{\underaccent{\bar}{#1}}

\allowdisplaybreaks

\usepackage{enumitem}

\usepackage{stmaryrd}

\usepackage{booktabs}
\usepackage{adjustbox}
\newcommand\Tstrut{\rule{0pt}{2.6ex}}       
\newcommand\Bstrut{\rule[-0.9ex]{0pt}{0pt}} 
\newcommand{\TBstrut}{\Tstrut\Bstrut} 
\usepackage{siunitx}

\jmlrheading{}{2021}{}{}{}{}{Alessandro De Palma, Rudy Bunel, Alban Desmaison, Krishnamurthy Dvijotham, Pushmeet Kohli, Philip H.S. Torr and M. Pawan Kumar}

\ShortHeadings{Improved Branch and Bound for NN Verification}{De Palma, Bunel, Desmaison, Dvijotham, Kohli, Torr, and Kumar}
\firstpageno{1}

\begin{document}

\title{Improved Branch and Bound for Neural Network Verification via Lagrangian Decomposition}

\author{\name Alessandro De Palma \email adepalma@robots.ox.ac.uk\\
			\name Rudy Bunel \email bunel.rudy@gmail.com\\
			\name Alban Desmaison \email desmaison.alban@gmail.com\\\
			\name Philip H.S. Torr \email phst@robots.ox.ac.uk \\
			\name M. Pawan Kumar \email pawan@robots.ox.ac.uk \\
	        \addr Department of Engineering Science\\
	        University of Oxford\\
	        Oxford OX1 3PJ
			\AND    
			\name Krishnamurthy (Dj) Dvijotham \email dvij@google.com \\
			\name Pushmeet Kohli \email pushmeet@google.com \\
			Deepmind\\
			London N1C 4AG
}

\editor{}

\maketitle

\begin{abstract}
%
We improve the scalability of Branch and Bound (BaB) algorithms for formally proving input-output properties of neural networks.
First, we propose novel bounding algorithms based on Lagrangian Decomposition. 
Previous works have used off-the-shelf solvers to solve relaxations at each node of the BaB tree, or constructed weaker relaxations that can be solved efficiently, but lead to unnecessarily weak bounds.
Our formulation restricts the optimization to a subspace of the dual domain that is guaranteed to contain the optimum, resulting in accelerated convergence.
Furthermore, it allows for a massively parallel implementation, which is amenable to GPU acceleration via modern deep learning frameworks.
Second, we present a novel activation-based branching strategy. By coupling an inexpensive heuristic with fast dual bounding, our branching scheme greatly reduces the size of the BaB tree compared to previous heuristic methods. Moreover, it performs competitively with a recent strategy based on learning algorithms, without its large offline training cost.
%
Finally, we design a BaB framework, named Branch and Dual Network Bound (BaDNB), based on our novel bounding and branching algorithms. 
We show that BaDNB outperforms previous complete verification systems by a large margin, cutting average verification times by factors up to $50$ on adversarial robustness properties.
\end{abstract}


\section{Introduction}

As deep learning powered systems become more and more common, the lack of
robustness of neural networks and their reputation for being ``black boxes" has become
increasingly worrisome. 
In order to deploy them in critical scenarios where safety and robustness would be a prerequisite, techniques that can prove formal guarantees for neural network behavior are needed.
A particularly desirable property is resistance to adversarial
examples~\citep{Goodfellow2015,Szegedy2014}: perturbations maliciously crafted
with the intent of fooling even extremely well performing models. After
several defenses were proposed and subsequently broken \citep{Athalye2018, Uesato2018}, some progress has been made in being able to formally verify whether there exist any adversarial examples in the
neighborhood of a data point~\citep{Tjeng2019, Wong2018}.

Verification algorithms fall into three categories: unsound (some false properties are proven false), incomplete (some true properties are proven true), and complete (all properties are correctly verified as either true or false).
Unsound verification, which relies on approximate non-convex optimization, 
is not related to the topic of this work. Instead, we focus on incomplete verification and its role in complete verification.
An incomplete verifier can be obtained via the computation of lower and upper bounds on the output of neural networks.
Many complete verifiers can be seen as branch and bound algorithms~\citep{Bunel2018}, which operate by dividing the property into subproblems (branching) for which incomplete verifiers are more likely to provide a definite answer (bounding).
\citet{Bunel2020} have recently proposed a branch and bound framework that scales to medium-sized convolutional networks, outperforming state-of-the-art complete verifiers~\citep{Katz2017,Wang2018b,Tjeng2019}. 
The aim of this work is to significantly improve their design choices, in order to scale up the applicability of complete verifiers to larger networks. In the remainder of this section, we provide a high-level overview of our proposed~improvements. 

\paragraph{Bounding} Most previous algorithms for computing bounds are either computationally
expensive~\citep{Ehlers2017} or sacrifice tightness in order to
scale~\citep{Gowal2018b,Mirman2018,Wong2018,Singh2018a,Zhang2018}. 
%
%
%
%
Within complete verification \citet{Bunel2018,Bunel2020} chose tightness over scalability, employing off-the-shelf solvers~\citep{gurobi-custom} to solve a network relaxation obtained by replacing activation functions by their convex hull~\citep{Ehlers2017}. In the context of incomplete verification, better speed-accuracy trade-offs were achieved by designing specialized solvers for such relaxation~\citep{Dvijotham2018}.
In this work, we design a novel dual formulation for the bounding problem and two corresponding solvers, which we employ as a branch and bound subroutine.
Our approach offers the following~advantages:
\begin{itemize}
	\item While previous bounding algorithms operating on the same network relaxation~\citep{Dvijotham2018} are based on Lagrangian relaxations, we derive a new family of optimization problems for neural network bounds through Lagrangian Decomposition, which in general yields duals at least as strong as those obtained through Lagrangian relaxation \citep{Guignard1987}. 
	For our bounding problem, the optimal solutions of both the Lagrangian Decomposition and Lagrangian relaxation will match.
	However we prove that, in the context of ReLU networks, for any dual bound from the approach by \citet{Dvijotham2018} obtained in the process of dual optimization, the corresponding bounds obtained by our dual are at least as tight. 
	Geometrically, our dual corresponds to a reduction of the dual space of the Lagrangian relaxation that always contains the optimum.
	We demonstrate empirically that our derivation computes tighter bounds in the same time when using supergradient methods, improving the quality of incomplete verification. 
	We further refine performance by devising a proximal solver for the problem, which decomposes the task into a series of strongly convex subproblems. For each subproblem, we use an iterative method that lends itself to analytical optimal step sizes, thereby resulting in faster convergence.
	\item Both the supergradient and the proximal method hinge on linear operations similar to those
	used during network forward/backward passes.  As a consequence, we can leverage the convolutional structure when necessary, while standard solvers are often restricted to treating it as a general linear operation.
	Moreover, both methods are easily parallelizable:
	when computing bounds on the activations at layer $k$, we need to solve two problems for each hidden unit of the network (for the upper and lower bounds). These can all be solved in parallel.
	Within branch and bound, we need to compute bounds for several different problem domains at once: we solve these problems in parallel as well.
	Our GPU implementation thus allows us to solve several hundreds of linear programs at once on a single GPU, a level of parallelism that would be hard to match on CPU-based systems.
\end{itemize}

\paragraph{Branching}
While bounding is often the computational bottleneck within each branch and bound iteration, a high quality branching strategy is crucial to reduce the branch and bound search tree~\citep{Achterberg2013}. 
Strategies used for neural network verification typically split the domain on a coordinate of the network input~\citep{Wang2018a,Bunel2018,Royo2019}, or on a given network activation~\citep{Ehlers2017,Katz2017,Wang2018b}.
It was recently shown~\citep{Bunel2020} that, for convolutional networks with around one thousand neurons, it is preferable to split on the network activations (activation splitting). As this search space is significantly larger, the best-performing heuristic strategy favors computational efficiency over accuracy~\citep{Bunel2020}. 
In order to improve performance without significantly increasing branching costs, strategies based on learning algorithms were proposed recently~\citep{Lu2020Neural}.
We present a novel branching strategy that, by coupling an inexpensive heuristic with fast dual bounds, greatly improves upon previous approaches strategies~\citep{Bunel2020} and performs competitively with learning algorithms without incurring large training costs.

\paragraph{BaDNB}
We design a massively parallel, GPU-accelerated, branch and bound framework around our bounding and branching algorithms.
We conduct detailed ablation studies over the various components of the framework, named BaDNB, and show that it yields substantial complete verification speed-ups over the state-of-the-art~algorithms.

\vspace{10pt}
A preliminary version of this work, centered around the novel bounding algorithms, appeared in the proceedings of the 36th Conference on Uncertainty in Artificial Intelligence~\citep{BunelDP20}.
The present article significantly extends it by 
(i) presenting a new heuristic branching scheme for activation splitting ($\S$\ref{sec:branching});
(ii) improving on various components of the branch and bound framework employed in the preliminary version ($\S$\ref{sec:badnb}), resulting in large complete verification improvements;
(iii) refining the analysis linking our dual problem to previous dual approaches ($\S$\ref{sec:bounding}, $\S$\ref{sec:dual_comp}), which now includes initialization via any propagation-based algorithm and a geometric explanation of the effectiveness of our dual compared to the one by \citet{Dvijotham2018};
and (iv) expanding the experimental analysis to include both new benchmarks, and new baselines such as CROWN~\citep{Zhang2018}, ERAN~\citep{eran}, nnenum~\citep{Bak2020} and VeriNet~\citep{Henriksen20}. 

The paper is organized as follows: in section \ref{sec:background}, we state the neural network verification problem and describe the technical background necessary for the understanding of our approach. 
Section \ref{sec:decomposition} presents our novel formulation for neural network bounding, yielding efficient incomplete verifiers. 
Section \ref{sec:branching} presents our branching scheme, to be used within branch and bound for complete verification. Technical and implementation details of BaDNB are outlined in section \ref{sec:further-impr}. 
In section \ref{sec:related}, we discuss related work in the context of our contributions.
Finally, sections \ref{sec:experiments-incomplete} and \ref{sec:experiments-complete} present an experimental evaluation of both our bounding algorithms and the branch and bound framework.

\section{Neural Network Verification} \label{sec:background}
Throughout this paper, we will use bold lower case letters (for example, $\xb$)  to represent vectors and upper case letters (for example, $W$) to represent matrices.
Brackets are used to indicate intervals ($[\hat{\lb}_k,\hat{\ub}_k]$) and vector or matrix entries ($\xb[i]$ or $W[i,j]$). 
Moreover, we use $\odot$ for the Hadamard product, $\llbracket \cdot,\cdot \rrbracket$ for integer ranges, $\mathds{1}_{{\mbf{a}}}$ for the indicator vector on condition~$\mbf{a}$. Finally, we write $\text{Conv}(f, \mbf{a}, \mbf{b})$ and $\text{Conv}(S)$ respectively for the convex hull of function $f$ defined in $[\mbf{a}, \mbf{b}]$, and for the convex hull of set $S$.

We begin by formally introducing the problem of neural network verification ($\S$\ref{sec:problem}), followed by an outline of two popular solution strategies ($\S$\ref{sec:bounding}, $\S$\ref{sec:bab}).

\subsection{Problem Specification} \label{sec:problem}

Given a $d$-layer feedforward neural network $f : \mathbb{R}^{n_0} \rightarrow \mathbb{R}^{n_d}$, an input domain $\mathcal{C}$, and a property $P$, verification problem $\left(f, \mathcal{C}, P\right)$ is defined as follows:
\begin{equation*}
	\xb_{0} \in \mathcal{C}\ \wedge\ \xb_{d} = f(\xb_{0}) \implies P(\xb_{d}).
\end{equation*}
%
%
Under the assumption that $P$ is a Boolean formula over linear inequalities (for instance, robustness to adversarial examples), we can represent both $f$ and $P$ as an $n$-layer neural network $f_P : \mathbb{R}^{n_0} \rightarrow \mathbb{R}$, which is said to be in canonical form if, for any $\xb_{0} \in \mathbb{R}^{n_0}$, $P\left(f(\xb_{0})\right) \implies f_P(\xb_{0}) \geq 0$~\citep{Bunel2018,Bunel2020}. 
Verifying $\left(f, \mathcal{C}, P\right)$ then reduces to finding the sign of the minimum of the following optimization~problem:
\begin{subequations}
	\begin{alignat}{2}
	\smash{\min_{\xb, \xbhat}}\quad \hat{x}_{n} \qquad
	\text{s.t. }\quad& \xb_0 \in \mathcal{C},\label{eq:verif-inibounds}\span\\
	& \xbhat_{k+1} = W_{k+1} \xb_{k} + \mathbf{b}_{k+1} \quad && k \in \left\llbracket0,n-1\right\rrbracket,\label{eq:verif-linconstr}\\
	& \xb_{k} = \sigma_k\left(\xbhat_k\right) \quad &&k \in \left\llbracket1,n-1\right\rrbracket,\label{eq:verif-relu}
	\end{alignat}
	\label{eq:noncvx_pb}%
\end{subequations} 
where constraints \eqref{eq:verif-linconstr} implement the linear layers of $f_P$ (fully connected or convolutional), while constraints \eqref{eq:verif-relu} implement its non-linear activation functions. We call $\xbhat_k \in \mathbb{R}^{n_k}$ pre-activations at layer $k$ and $n_k$ denotes the layer's width.
In line with \citet{Dvijotham2018}, we assume that linear functions can be easily optimized over $\mathcal{C}$. In the following, we will first describe how to solve problem \eqref{eq:noncvx_pb} approximately ($\S$\ref{sec:bounding}), then exactly ($\S$\ref{sec:bab}).


\subsection{Neural Network Bounding}  \label{sec:bounding}
The non-linearity of constraint~\eqref{eq:verif-relu} makes problem \eqref{eq:noncvx_pb} non-convex and NP-HARD~\citep{Katz2017}. 
Therefore, many authors \citep{Wong2018,Dvijotham2018,Zhang2018,Raghunathan2018,Singh2019} have instead focused on the computation of a lower bound on the minimum, which significantly simplifies the optimization problem thereby yielding an efficient incomplete verification method.
Here, we are concerned with approaches that allow for a dual interpretation (see $\S$\ref{sec:related} for an overview).

\subsubsection{Propagation-based methods} \label{sec:propagation}

Assume we have access to upper and lower bounds (respectively $\hat{\ub}_k$ and $\hat{\lb}_k$) on the value that $\xbhat_k$ can take, for $k \in \left\llbracket1, n-1\right\rrbracket$. We call these \emph{intermediate bounds}: we detail how to compute them in $\S$\ref{sec:intermediate}.
Moreover, let $\lsigmab_{k}(\xbhat_{k}) = \ubar{\mbf{a}}_k \odot \xbhat_{k} + \ubar{\mbf{b}}_k$ and $\usigmab_{k}(\xbhat_{k}) = \bar{\mbf{a}}_k \odot \xbhat_{k} + \bar{\mbf{b}}_k$ be two linear functions that bound $\sigma_k(\xbhat_{k})$ from below and above, respectively.
Then, problem~\eqref{eq:noncvx_pb} can be replaced by the following convex outer approximation:
\begin{equation}
\begin{aligned}
\smash{\min_{\xb, \xbhat}}\quad \hat{x}_{n} \qquad
\text{s.t. }\quad& \xb_0 \in \mathcal{C}, \\
& \xbhat_{k+1} = W_{k+1} \xb_{k} + \mathbf{b}_{k+1} && k \in \left\llbracket0,n-1\right\rrbracket,\\
& \lsigmab_{k}(\xbhat_{k}) \leq \xb_{k} \leq \usigmab_{k}(\xbhat_{k}) && k \in \left\llbracket1,n-1\right\rrbracket, \\
& \xbhat_k \in [\hat{\lb}_k, \hat{\ub}_k] && k \in \left\llbracket1, n-1\right\rrbracket.
\end{aligned}
\label{eq:primal-propagation}
\end{equation}
A popular and inexpensive class of bounding algorithms solves a relaxation of problem~\eqref{eq:primal-propagation} by back-propagating $\lsigmab_{k}(\xbhat_{k})$ and $\usigmab_{k}(\xbhat_{k})$ through the network \citep{Wong2018,Weng2018,Singh2018a,Zhang2018,Singh2019}.
In the dual space, these methods correspond to evaluating the Lagrangian relaxation of problem \eqref{eq:primal-propagation} at a specific dual point. 
Let us denote $[\mbf{a}]_-=\min(0, \mbf{a})$, $[\mbf{a}]_+=\max(0, \mbf{a})$. 
The Lagrangian relaxation of problem \eqref{eq:primal-propagation} can be written in the following unconstrained form \citep[equations (8), (9), (38)]{Salman2019}:  
\begin{equation}
\begin{aligned}
\max_{\mub,\lambdab}\enskip & d_{P}(\mub,\lambdab),  \quad \text{where: } \\
\hspace{-5pt}d_{P}(\mub,\lambdab) &=  \min_{\xb, \xbhat} \left[ \begin{array}{l}
W_n\xb_{n-1} + b_n + \sum_{k=1}^{n-1} \mub_k^T \left( \xbhat_k - W_k \xb_{k-1} - \mbf{b}_k \right) \\[4pt]
\sum_{k=1}^{n-1} [\lambdab_k]_-^T \left( \xb_k - \lsigmab_{k}(\xbhat_k) \right) + \sum_{k=1}^{n-1} [\lambdab_k]_+^T \left( \xb_k - \usigmab_{k}(\xbhat_k) \right) \end{array} \right.\\
\hspace{-5pt}\text{s.t. }\enskip& \xb_0 \in \mathcal{C}, \\
& \xbhat_k \in [\hat{\lb}_k, \hat{\ub}_k] \qquad k \in \left\llbracket1, n-1\right\rrbracket.
\end{aligned}
\label{eq:dual-propagation}
\end{equation}
\citet{Salman2019} show that propagation-based bounding algorithms are equivalent to evaluating problem \eqref{eq:dual-propagation} at a suboptimal point $(\bar{\mub},\bar{\lambdab})$, given by:
\begin{equation}
\begin{aligned}
&\bar{\lambdab}_{n-1} = -W_{n}^T\mbf{1},\\
&\bar{\mub}_k = \bar{\mbf{a}}_k \odot [\bar{\lambdab}_k]_+ +  \ubar{\mbf{a}}_k \odot [\bar{\lambdab}_k]_- && k \in \left\llbracket1, n-1\right\rrbracket,\\
&\bar{\lambdab}_{k-1} = W_k^T\bar{\mub}_k && k \in \left\llbracket2, n-1\right\rrbracket.
\end{aligned}
\label{eq:salman-assignment}
\end{equation}
The dual assignment \eqref{eq:salman-assignment} is obtained via a single \emph{backward pass} through the network, an operation analogous to the gradient backpropagation employed for neural network training.
Moreover, exploiting the structure of equation \eqref{eq:salman-assignment}, the objective value of problem \eqref{eq:dual-propagation} at such dual point can be conveniently computed as:
\begin{equation}
\begin{aligned}
\hspace{-5pt}d_P(\bar{\mub},\bar{\lambdab}) &=  \min_{\xb_0 \in \mathcal{C}} 
\left(- \bar{\mub}_{1}^T W_0 \xb_{0}\right) +  b_n
- \sum_{k=1}^{n-1} \left( [\bar{\lambdab}_k]_-^T  \ubar{\mbf{b}}_k + [\bar{\lambdab}_k]_+^T\bar{\mbf{b}}_k +  \bar{\mub}_k^T\mbf{b}_k\right). 
\end{aligned}
\label{eq:bounds-propagation}
\end{equation}

\subsubsection{Lagrangian Relaxation of the non-convex formulation} \label{sec:dj}

Propagation-based methods provide a lower bound to problem~\eqref{eq:primal-propagation}, which is a rather loose approximation to problem \eqref{eq:noncvx_pb}.
An alternative approach, presented by \citet{Dvijotham2018}, relies on taking the dual of non-convex problem \eqref{eq:noncvx_pb} directly and solving it via supergradient methods.
By relaxing \eqref{eq:verif-linconstr} and \eqref{eq:verif-relu} via Lagrangian multipliers, and exploiting intermediate bounds, \citet{Dvijotham2018} obtain the following dual:
\begin{equation}
\begin{aligned}
\max_{\mub,\lambdab}\enskip & d_{O}(\mub,\lambdab),  \quad \text{where: } \\
\hspace{-5pt}d_{O}(\mub, \lambdab) &=  \min_{\xb, \xbhat} \left[ \begin{array}{l}
\sum_{k=1}^{n-1} \mub_k^T \left( \xbhat_k - W_k \xb_{k-1} - \mbf{b}_k \right) 
+ \sum_{k=1}^{n-1} \lambdab_k^T \left( \xb_k - \sigma_k(\xbhat_k) \right) \\[4pt]
+ W_n\xb_{n-1} + b_n  \end{array} \right.\\
\text{s.t. }\quad& \xb_0 \in \mathcal{C},\\
& (\xb_k, \xbhat_k) \in [\sigma_k(\hat{\lb}_k),\sigma_k(\hat{\ub}_k)]  \times  [\hat{\lb}_k, \hat{\ub}_k] \qquad k \in \left\llbracket1, n-1\right\rrbracket.
\end{aligned}
\label{eq:dj_prob-main}
\end{equation}
For $\sigma_k(\xbhat_k) = \max(\xbhat_k, 0)$, \citet{Dvijotham2018}, prove that problem \eqref{eq:dj_prob-main} is equivalent to a dual of the following convex problem\footnote{\citet{Dvijotham2018} write $\text{Conv}(\sigma_k, \hat{\lb}_k, \hat{\ub}_k) =  \left\{\left(\xb_k, \xbhat_k\right) | \lsigmab_{k, \text{opt}}(\xbhat_{k}) \leq \xb_{k} \leq \usigmab_{k, \text{opt}}(\xbhat_{k})\right\}$, and consider the dual that results from relaxing both inequalities, along with equality \eqref{eq:verif-linconstr}.}:
\begin{equation}
\begin{aligned}
\smash{\min_{\xb, \xbhat}}\quad \hat{x}_{n} \qquad
\text{s.t. }\quad& \xb_0 \in \mathcal{C}, \\
& \xbhat_{k+1} = W_{k+1} \xb_{k} + \mathbf{b}_{k+1} && k \in \left\llbracket0,n-1\right\rrbracket,\\
& (\xb_k, \xbhat_{k}) \in \text{Conv}(\sigma, \hat{\lb}_k, \hat{\ub}_k) && k \in \left\llbracket1,n-1\right\rrbracket, \\
& (\xb_k, \xbhat_k) \in [\sigma_k(\hat{\lb}_k),\sigma_k(\hat{\ub}_k)]  \times  [\hat{\lb}_k, \hat{\ub}_k]   && k \in \left\llbracket1,n-1\right\rrbracket,
\end{aligned}
\label{eq:planet-relax}
\end{equation}
where $\text{Conv}(\sigma_k, \hat{\lb}_k, \hat{\ub}_k)$ is the convex hull of constraint \eqref{eq:verif-relu}.
\citet{Salman2019} generalize the result to any activation function that acts element-wise on $\xbhat_k$ and prove that, under mild assumptions, strong duality holds for problem \eqref{eq:planet-relax}. 
Therefore, as $\text{Conv}(\sigma_k, \hat{\lb}_k, \hat{\ub}_k) \subseteq  \left\{\left(\xb_k, \xbhat_k\right) | \lsigmab_{k}(\xbhat_{k}) \leq \xb_{k} \leq \usigmab_{k}(\xbhat_{k})\right\}$, the bounding algorithm by \citet{Dvijotham2018} will converge to tighter bounds than propagation-based algorithms. 

\subsubsection{Intermediate bounds} \label{sec:intermediate}

Convex relaxations (for instance, problems \eqref{eq:primal-propagation}, \eqref{eq:planet-relax}) and dual problems (for instance, problem \eqref{eq:dj_prob-main}) are often defined as a function of intermediate bounds $\hat{\lb}_j \text{ and } \hat{\ub}_j$ on the values of $\xbhat_j \in \mathbb{R}^{n_j}$.
These values are computed by running a bounding algorithm over subnetworks. 
Specifically, we are looking for bounds to versions of problem \eqref{eq:noncvx_pb} for which, instead of defining the objective function on the activation of layer $n$, we define it over $\xbhat_j[i]$. As we need to repeat this process for $j \in \left\llbracket1,n-1\right\rrbracket$ and $i \in \left\llbracket1,n_j\right\rrbracket$, intermediate bound computations can easily become the computational bottleneck for neural network bounding.
Therefore, typically, intermediate bounds are computed with inexpensive propagation-based algorithms ($\S$\ref{sec:propagation}), whereas the lower bounding of the network output $\hat{x}_n$ relies on more costly convex relaxations (outlined in $\S$\ref{sec:dj})~\citep{Bunel2020}.

\subsection{Branch and Bound} \label{sec:bab}

Neural network bounding is concerned with solving an approximation of problem \eqref{eq:noncvx_pb}, and may verify a subset of the properties: those for which the computed lower bound is positive.
However, in order to guarantee that any given property will be verified, we need to solve problem \eqref{eq:noncvx_pb} exactly.
The lack of convexity rules out local optimization algorithms such as gradient descent, which will not provably converge to the global optimum. Therefore, many complete verification methods are akin to global optimization algorithms such as branch and bound (see $\S$\ref{sec:related})~\citep{Bunel2018}. 

\subsubsection{Operating principle}
In the context of our verification problem \eqref{eq:noncvx_pb}, branch and bound starts by computing bounds on the minimum: a lower bound is obtained via a \emph{bounding algorithm} (see $\S$\ref{sec:bounding}), while an upper bound can be determined heuristically, as any feasible point yields a valid upper bound. 
If the property cannot yet be verified (that is, the lower bound is negative and the upper bound is positive), the property's feasible domain is divided into a number of smaller problems via some \emph{branching strategy}. 
The algorithm then proceeds by computing bounds for each subproblem, exploiting the fact that a subproblem's lower bound is guaranteed to be at least as tight as the one for its parent problem (that is, before the branching).
Subproblems which cannot contain the global lower bound are progressively discarded: in the canonical form (see $\S$\ref{sec:problem}), this happens if a local lower bound is positive. An incumbent solution to problem \eqref{eq:noncvx_pb} is defined as the smallest encountered upper bound. The order in which subproblems are explored is determined by a \emph{search strategy}. Finally, the verification procedure terminates when either no subproblem has a negative lower bound, or when the incumbent becomes positive.

\subsubsection{Branch and bound for piecewise-linear networks} \label{sec:babsr}

We now turn our attention to the class of piecewise-linear networks. For simplicity, we assume all the activation functions are ReLUs, as other common piecewise-linear activations such as MaxPooling units can be converted into a series of ReLU-based layers~\citep{Bunel2020}.
We describe BaBSR from \citet{Bunel2020}, a specific instantiation of branch and bound that proved particularly effective in the context of larger piecewise-linear networks.

\begin{figure}[b!]
	\vskip -12pt
	\centering
	\noindent\resizebox{.28\textwidth}{!}{
		\begin{tikzpicture}
  \tikzset{dummy/.style= {inner sep=0, outer sep=0}}
  \tikzset{cross/.style={circle, draw,
      minimum size=5*(#1-\pgflinewidth),
      inner sep=0pt, outer sep=0pt,
      ultra thick, color=red}}

  \draw[-, ultra thick](-1, 0) to (0, 0) to (1, 1);

  \draw[dashed](-1, -0.5) to (-1, 1.5);
  \draw[dashed](1, -0.5) to (1, 1.5);

  \draw[fill=green](-1, 0) -- (0,0) -- (1, 1);

  \node[dummy](lb-lab) at (-1.3, -0.3) {$\hat{l}_{k[j]}$\hspace{3pt}};
  \node[dummy](ub-lab) at (1.35, -0.3) {\hspace{3pt}$\hat{u}_{k[j]}$};

  \node[cross=2pt] at (-1, 0) {};
  \node[cross=2pt] at (1, 1) {};
  \node[cross=2pt] at (0, 0) {};

  \draw[-latex](-1.5,0) to (2, 0);
  \node[dummy](x-label) at (2.3, 0) {\hspace{3pt} $\xbhat_{k[j]}$};
  \draw[-latex](0,-0.5) to (0, 1.5);
  \node[dummy](x-label) at (0, 1.8) {$\xb_{k[j]}$};

\end{tikzpicture}
	}
	\caption{Feasible domain of the convex hull for an ambiguous ReLU. Red
		circles indicate the vertices of the feasible region.}
	\label{fig:relucvx}
\end{figure}
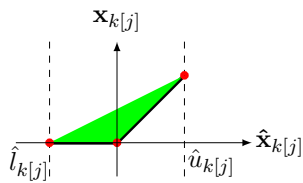

\vspace{3pt}
Let us classify ReLU activations depending on the signs of pre-activation bounds $\hat{\lb}_k$ and $\hat{\ub}_k$. A given ReLU $\sigma_k(\xbhat_k[i])=\max(\xbhat_k[i], 0)$ is passing if $\hat{\lb}_k[i] \geq 0$, blocking if $\hat{\ub}_k[i] \leq 0$, and ambiguous otherwise. Note that non-ambiguous ReLUs can be replaced by linear functions.
At every iteration, BaSBR picks the subproblem with the lowest lower bound, and branches by separating an ambiguous ReLU into its two linear phases (\emph{ReLU branching}). 
The ReLU on which to branch is selected according to a heuristic that estimates the effect of the split on the subproblem lower bound, based on an inexpensive approximation of the bounding algorithm by \citet{Wong2018} (for details, see $\S$\ref{sec:branching}).
Lower bounding is performed by solving the Linear Program (LP) corresponding to the ReLU version of problem \eqref{eq:planet-relax}, where the convex hull is defined as follows~\citep{Ehlers2017}:
\begin{subnumcases}
{\label{eq:planet-relu-hull}\hspace{-30pt}\text{Conv}(\sigma_k, \hat{\lb}_k, \hat{\ub}_k):=}
\hspace{-5pt}\begin{array}{l}
\xb_{k} \geq 0, \quad \xb_k \geq \xbhat_k \\[3pt]
\xb_{k} \leq \frac{ \hat{\ub}_k \odot (\xbhat_k - \hat{\lb}_k)}{ \hat{\ub}_k-\hat{\lb}_k}
\end{array} & if $\hat{\lb}_k \leq  0$ and $\hat{\ub}_k \geq 0$, \label{eq:planet-reluamb} \\[5pt]
\xb_k = 0 & if  $\hat{\ub}_k \leq 0$, \label{eq:planet-relublock}\\[5pt]
\xb_k = \xbhat_k & if  $\hat{\lb}_k \geq 0$. \label{eq:planet-relupass}
\end{subnumcases}
Upper bounds are computed by evaluating the neural network at the solution of the LP.
Finally, the intermediate bounds for each LP are obtained by taking the layer-wise best bonds between interval bound propagation~\citep{Gowal2018b,Mirman2018} and the propagation-based method by \citet{Wong2018}. 

In the remainder of this paper, we present improvements to both the bounding algorithm ($\S$\ref{sec:decomposition}) and the branching strategy ($\S$\ref{sec:branching}), then outline the details of the resulting branch and bound framework ($\S$\ref{sec:further-impr}).

\section{Better Bounding: Lagrangian Decomposition} \label{sec:decomposition}

We will now describe a novel dual approach to obtain a lower bound to problem~\eqref{eq:noncvx_pb} and relate it to the duals described in section \ref{sec:bounding}.
We present two bounding algorithms: a supergradient method ($\S$\ref{subsec:subgradient_method}), and a solver based on proximal maximization ($\S$\ref{sec:prox}).

\subsection{Problem Derivation}

Our approach is based on Lagrangian Decomposition, also known as variable splitting~\citep{Guignard1987}. Due to
the compositional structure of neural networks, most constraints involve only a limited number of variables. As a result, we can split the
problem into meaningful, easy to solve subproblems. We then impose
constraints that the solutions of the subproblems should agree.


\vspace{3pt}
We start from problem \eqref{eq:planet-relax}, a convex outer approximation of the original non-convex problem \eqref{eq:noncvx_pb} where activation functions are replaced by their convex hull.
In the following, we will use ReLU activation functions as an example: their convex hull is defined in equation \eqref{eq:planet-relu-hull}. We stress that the derivation can be extended to other
non-linearities. For example, appendix \ref{sec:sigmoid_act} describes the case
of sigmoid activation function.
In order to obtain an efficient decomposition, we divide the constraints into subsets that allow for easy optimization subtasks. 
Each subset will correspond to a pair of an activation layer, and the linear layer coming after it. 
The only exception is the first linear layer which is combined with the restriction of the input domain to $\mathcal{C}$. 
Using this grouping of the constraints, we can concisely write problem \eqref{eq:planet-relax} as:
\begin{equation}
\begin{aligned}
&\smash{\min_{\xb, \xbhat}} \quad \hat{x}_n \qquad \text{s.t. }\quad && \mathcal{P}_0(\xb_0, \xbhat_{1}), \\
&&&\mathcal{P}_k(\xb_{k}, \xbhat_{k}, \xbhat_{k+1}) \qquad k \in \left\llbracket1,n-1\right\rrbracket, \\[5pt]
\end{aligned}
\label{eq:nondec_pb}
\end{equation}
where the constraint subsets are defined as:
\begin{equation*}
\begin{aligned}
\mathcal{P}_0(\xb_{0}, \xbhat_{1}) & := \begin{cases} \xb_{0} \in \mathcal{C}\\
\xbhat_{1} = W_{1} \xb_{0} + \mathbf{b}_{1},
\end{cases} \\
\mathcal{P}_k(\xb_k, \xbhat_{k}, \xbhat_{k+1}) &:=
\begin{cases}
(\xb_k, \xbhat_{k}) \in \text{Conv}(\sigma_k, \hat{\lb}_k, \hat{\ub}_k), \\
(\xb_k, \xbhat_k) \in [\sigma_k(\hat{\lb}_k),\sigma_k(\hat{\ub}_k)]  \times  [\hat{\lb}_k, \hat{\ub}_k] , \\
\xbhat_{k+1} = W_{k+1} \xb_{k} + \mathbf{b}_{k}.
\end{cases}
\end{aligned}
\end{equation*}
To obtain a Lagrangian Decomposition, we duplicate the variables so that each
subset of constraints has its own copy of the variables in which it is involved. Formally, we
rewrite problem~\eqref{eq:nondec_pb} as follows:
\begin{subequations}
	\begin{alignat}{2}
	\smash{\min_{\xb, \xbhat}} \quad \hat{x}_n \qquad \text{s.t. }\quad & \mathcal{P}_0(\xb_0, \xbhat_{A, 1}),  \\
	&\mathcal{P}_k(\xb_{k}, \xbhat_{B, k}, \xbhat_{A, k+1}) &&\qquad k \in \left\llbracket1,n-1\right\rrbracket,\\
	&\xbhat_{A, k} = \xbhat_{B, k} &&\qquad k \in \left\llbracket1,n-1\right\rrbracket. \label{eq:matcheq}%
	\end{alignat}
	\label{eq:primal_lag_problem}%
\end{subequations}
The additional equality constraints \eqref{eq:matcheq} impose agreements between
the various copies of variables. We introduce the dual variables $\lambdabhat$
and derive the Lagrangian dual:
\begin{equation}
\begin{split}
\max_{\lambdabhat}\enskip q(\lambdabhat),  \quad &\text{where: } \qquad q\left( \lambdabhat \right) = \min_{\xb, \xbhat}\ \hat{x}_{A, n}
+ \sum_{k=1}^{n-1} \lambdabhat_k^T \left( \xbhat_{B, k} - \xbhat_{A, k} \right)\\[5pt]
\text{s.t. }\quad & \mathcal{P}_0(\xb_0, \xbhat_{A,1}), \\
&\mathcal{P}_k(\xb_k, \xbhat_{B, k}, \xbhat_{A, k+1}) \qquad k \in \left\llbracket1,n-1\right\rrbracket.
\end{split}
\label{eq:nonprox_problem}
\end{equation}
Any value of $\lambdabhat$ provides a valid lower bound by
virtue of weak duality. While we will maximize over the choice of dual variables
in order to obtain as tight a bound as possible, we will be able to
interrupt the optimization process at any point and obtain a valid bound by
evaluating $q$. It remains to show how to solve problem \eqref{eq:nonprox_problem} efficiently in practice: this is the subject of $\S$\ref{subsec:subgradient_method} and $\S$\ref{sec:prox}. 
In $\S$\ref{sec:dual_comp}, we analyze the relationship between our dual \eqref{eq:nonprox_problem}, propagation-based methods and problem \eqref{eq:dj_prob-main}.

\subsection{Supergradient Solver}
\label{subsec:subgradient_method}
In line with the work by \citet{Dvijotham2018}, who use supergradient methods on their dual \eqref{eq:dj_prob-main}, 
we present a supergradient-based solver (Algorithm \ref{alg:subg}) for problem \eqref{eq:nonprox_problem}.

\begin{figure}[b!]
	\centering
	\begin{minipage}{.99\textwidth}
		\begin{algorithm}[H]
		\caption{Supergradient method}\label{alg:subg}
		\begin{algorithmic}[1]
			\algrenewcommand\algorithmicindent{1.0em}%
			\Function{supergradient\_compute\_bounds}{Problem \eqref{eq:planet-relax}}
			\State Initialize dual variables $\lambdabhat^0$ via proposition \ref{prop:init}
			\For{$t \in \llbracket0, T-1\rrbracket$}
			\State $\xbhat^{*,t} \in \argmin_{\xb, \xbhat} q(\lambdabhat)$ as detailed in $\S$\ref{sec:innerminp0} and $\S$\ref{sec:innerminpk} \Comment{inner minimization}
			\State $\nabla_{\lambdabhat} q(\lambdabhat^t) \leftarrow \xbhat_B^{*,t} - \xbhat_A^{*,t} $  \Comment{compute supergradient}
			\State $\lambdabhat^{t+1} \leftarrow \lambdabhat^{t} + \alpha^t \nabla_{\lambdabhat} q(\lambdabhat^t)$ \Comment{supergradient step} \label{alg:adam-update}
			\EndFor\label{alg:inner_loopend}
			\State\Return $q(\lambdabhat^T)$
			\EndFunction
		\end{algorithmic}
		\end{algorithm}
\end{minipage}%
\end{figure}

At a given point $\lambdabhat$, obtaining the supergradient requires us to know
the values of $\xbhat_{A}$ and $\xbhat_{B}$ for which the inner minimization is
achieved. Based on the identified values of $\xbhat_{A}^*$ and $\xbhat_{B}^*$, we can then compute
the supergradient \mbox{$\nabla_{\lambdabhat} q= \xbhat_{B}^* - \xbhat_{A}^*$} and move in its direction:
\begin{equation}
\lambdabhat^{t+1} = \lambdabhat^{t} + \alpha^t \nabla_{\lambdabhat} q(\lambdabhat^t),
\label{eq:subgradient_update}
\end{equation}
where $\alpha^t$ corresponds to a step size schedule that needs to be provided.
It is also possible to use any variant of gradient descent, such as
Adam~\citep{Kingma2015}.

It remains to show how to perform the inner minimization over the primal variables.
By design, each of the
variables is only involved in one subset of constraints. As a result, the
computation completely decomposes over the
subproblems, each corresponding to one of the subset of constraints. We
therefore simply need to optimize linear functions over one subset of
constraints at a time. 

\subsubsection{Inner minimization: $\mathcal{P}_0$ subproblems} \label{sec:innerminp0}
To minimize over $\xb_{0}, \xbhat_{A, 1}$, the variables constrained by $\mathcal{P}_0$, we need to solve:
\begin{equation}
\begin{split}
(\xb_{0}^*, \xbhat_{A, 1}^*) &= \argmin_{\xb_{0}, \xbhat_{A, 1}}\ -\lambdabhat_1^T \xbhat_{A, 1} \\
\text{s.t } \quad &\xb_0 \in \mathcal{C}, \qquad \xbhat_{A, 1} = W_{1} \xb_0 + \mathbf{b}_{0}.
\end{split}
\label{eq:typeA_cgd}
\end{equation}
Rewriting the problem as a linear function of $\xb_{0}$ only, problem
\eqref{eq:typeA_cgd} is simply equivalent to minimizing $-\lambdabhat_1^T W_1 \xb_0$
over $\mathcal{C}$. We assumed that the optimization of a linear function
over $\mathcal{C}$ was efficient. We now give examples of $\mathcal{C}$ where problem
\eqref{eq:typeA_cgd} can be solved efficiently.

\paragraph{Bounded Input Domain}
If $\mathcal{C}$ is defined by a set of lower bounds $\lb_0$ and upper bounds
$\ub_0$ (as in the case of $\ell_{\infty}$ adversarial examples),
optimization will simply amount to choosing either the lower or upper bound
depending on the sign of the linear function. The optimal solution is:
\begin{equation}
\xb_{0} =\  \mathds{1}_{\lambdabhat_1^T W_1 < 0} \odot \hat{\lb}_{0} +\mathds{1}_{\lambdabhat_1^T W_1 \geq 0} \odot \hat{\ub}_0, \qquad
\xbhat_{A, 1} =\ W_{1} \xb_{0} + \mbf{b}_{1}.
\label{eq:box_bounds_sol}
\end{equation}

\paragraph{$\ell_2$ Balls} If $\mathcal{C}$ is defined by an
$\ell_2$ ball of radius $\epsilon$ around a point $\mbf{\bar{x}}$
($\left\lVert\xb_0 - \mbf{\bar{x}}\right\rVert_2 \leq \epsilon$), optimization amounts to choosing
the point on the boundary of the ball such that the vector from the center to it
is opposed to the cost function. Formally, the optimal solution is given by:
\begin{equation}
\xbhat_{A, 1} =\  W_{1} \xb_{0} + \mbf{b}_{1}, \qquad
\xb_0 =\ \mbf{\bar{x}} + \left(\nicefrac{\epsilon}{ \left\lVert \lambdabhat_1 \right\rVert_2}\right) \lambdabhat_1.
\end{equation}

\subsubsection{Inner minimization: $\mathcal{P}_k$ subproblems} \label{sec:innerminpk}
For the variables constrained by subproblem $\mathcal{P}_k$ ($\xb_{k}, \xbhat_{B, k}, \xbhat_{A, k+1}$), we need to solve:
\begin{equation}
\begin{aligned}
\hspace{-7pt}(\xbhat_{B, k}^*&, \xbhat_{A, k+1}^*) = \smash{\argmin_{\xbhat_{B, k}, \xbhat_{A, k+1}}} \quad
\lambdabhat_k^T \xbhat_{B, k} - \lambdabhat_{k+1}^T \xbhat_{A, k+1} \\[8pt]
\text{s.t } & \quad (\xb_k, \xbhat_{B, k}) \in \text{Conv}(\sigma_k, \hat{\lb}_k, \hat{\ub}_k),\\
& \quad (\xb_k, \xbhat_{B,k}) \in [\sigma_k(\hat{\lb}_k),\sigma_k(\hat{\ub}_k)]  \times  [\hat{\lb}_k, \hat{\ub}_k] ,\\
& \quad \xbhat_{A, k+1} = W_{k+1} \xb_{k} + \mathbf{b}_{k+1}.
\end{aligned} \label{eq:pk_sub}
\end{equation}
We outline the minimization steps for the case of ReLU activation functions.
However, the following can be generalized to other activations. For example, appendix \ref{sec:sigmoid_act} describes the solution for
the sigmoid function.
For $\sigma_k(\xbhat_k) = \max(\mathbf{0}, \xbhat_k)$, $\text{Conv}(\sigma_k, \hat{\lb}_k, \hat{\ub}_k)$ is given by equation~\eqref{eq:planet-relu-hull}, and we can
find a closed form solution. 
Using the last equality of the problem, and omitting constant terms, we can start by
rewriting the objective function as \mbox{$\lambdabhat_k^T \xbhat_{B, k} - \lambdabhat_{k+1}^T W_{k+1} \xb_{k}$}. 
The optimization can be performed independently for each of the coordinates, over which both the objective function and the constraints decompose completely.
The ensuing minimization will then depend on whether a given ReLU is ambiguous or equivalent to a linear function. 

\paragraph{Ambiguous ReLUs} If the ReLU is ambiguous, the shape of the
convex hull is represented in Figure~\ref{fig:relucvx}. 
For each dimension $i$, problem \eqref{eq:pk_sub} is a linear program, which means that the optimal point will be a vertex. 
The possible vertices for $(\xbhat_{B,k}[i], \xb_{k}[i])$ are \mbox{$( \hat{\lb}_k[i], 0 )$}, \mbox{$\left( 0, 0 \right)$},
and \mbox{$\left( \hat{\ub}_k[i], \hat{\ub}_k[i] \right)$}. In order to find the minimum, we can therefore 
evaluate the objective function at these three points and keep the one with the smallest value. Denoting the vertex set by $V_{k, i}$:
\begin{equation}
\argmin_{(\xbhat_{B,k}[i], \xb_{k}[i]) \in V_{k, i}} \enskip \bigg(\lambdabhat_k[i] \xbhat_{B, k}[i] - (\lambdabhat_{k+1}^T W_{k+1} )\hspace{-1pt}[i]\xb_{k}[i]\bigg).
\label{eq:min-pk}
\end{equation}

\paragraph{Non-ambiguous ReLUs}
If for a ReLU we have $\lb_k[i] \geq 0$ or $\ub_k[i] \leq 0$, $\text{Conv}(\sigma_k, \hat{\lb}_k, \hat{\ub}_k)$ 
is a simple linear equality constraint. For those coordinates, the
problem is analogous to the one solved by equation~\eqref{eq:box_bounds_sol},
with the linear function being minimized over the $\xbhat_{B,k}[i]$ box bounds
being $\lambdabhat_k^T[i] \xbhat_{B,k}[i]$ in the case of blocking ReLUs or \mbox{$\left(
	\lambdabhat_{k}^T - \lambdabhat_{k+1}^TW_{k+1}\right)\hspace{-3pt}[i] \xbhat_{B, k}[i]$} in the case of
passing ReLUs.

\subsection{Proximal Solver} \label{sec:prox}
We now present a second solver (Algorithm \ref{alg:prox}) for problem \eqref{eq:nonprox_problem}, relying on proximal maximization rather than supergradient~methods (as in $\S$\ref{subsec:subgradient_method}). 


\subsubsection{Augmented Lagrangian}
Applying proximal maximization to the dual function $q$ results in the Augmented
Lagrangian Method, which is also known as the method of multipliers.  
Let us indicate the value of a variable at the
$t$-th iteration via superscript $t$.
For our problem\footnote{We refer the reader to \citet{Bertsekas1989} for the derivation of the update steps.}, the method of multipliers
will correspond to alternating between the following updates to the dual variables
$\lambdabhat$:
\begin{equation}
\lambdabhat_k^{t+1} = \lambdabhat_k^{t} + \frac{\xbhat^{t}_{B,k} - \xbhat^{t}_{A,k}}{\eta_k} \label{eq:lambda_updates},
\end{equation}
and updates to the primal variables $\xbhat$, which are carried out as follows:
\begin{equation}
\begin{aligned}
\left( \xb^{t}, \xbhat^{t} \right) \ &= \argmin_{\xb, \xbhat} \mathcal{L}\left( \xbhat, \lambdabhat^t\right), \qquad \text{where:} \\
&\hspace{-80pt}\mathcal{L}\left( \xbhat, \lambdabhat^t\right):= \xbhat_{A, n} + \sum_{k=1}^{n-1} \lambdabhat_k^T \left( \xbhat_{B, k} - \xbhat_{A, k} \right) 
+ \sum_{k=1}^{n-1} \frac{1}{2 \eta_k}  \| \xbhat_{B, k} - \xbhat_{A, k}\|^2 \\[3pt]
\text{s.t. }\quad & \mathcal{P}_0(\xb_0, \xbhat_{A,1}), \\
&\mathcal{P}_k(\xb_k, \xbhat_{B, k}, \xbhat_{A, k+1}) \qquad k \in \left\llbracket1,n-1\right\rrbracket.
\end{aligned}
\label{eq:z_updates}
\end{equation}
The term $\auglag$ is the Augmented Lagrangian of problem \eqref{eq:primal_lag_problem}.
The additional quadratic term in \eqref{eq:z_updates}, compared to the objective of $q(\lambdabhat)$,
arises from the proximal terms on $\lambdabhat$. It has the advantage of
making the problem strongly convex, and hence easier to optimize. Later on,
we will show that this allows us to derive optimal
step-sizes in closed form. The weight $\eta_k$ is a hyperparameter of the algorithm. A high
value will make the problem more strongly convex and therefore quicker to solve,
but it will also limit the ability of the algorithm to perform large updates.

While obtaining the new values of $\lambdabhat$ is trivial using
equation~\eqref{eq:lambda_updates}, problem \eqref{eq:z_updates} does not have a
closed-form solution. We show how to solve it efficiently
nonetheless.

\begin{figure}[t!]
	\centering
	\begin{minipage}{.99\textwidth}
		\begin{algorithm}[H]
			\caption{Proximal method}\label{alg:prox}
			\begin{algorithmic}[1]
				\algrenewcommand\algorithmicindent{1.0em}%
				\Function{Proximal\_compute\_bounds}{Problem \eqref{eq:planet-relax}}
				\State Initialize dual variables $\lambdabhat^0$ via proposition \ref{prop:init}
				\State $\xbhat^0 \in \argmin_{\xb, \xbhat} q(\lambdabhat^0)$ as detailed in $\S$\ref{sec:innerminp0} and $\S$\ref{sec:innerminpk} \Comment{initialize primals}
				\For{$t \in \llbracket0, T-1\rrbracket$}
				\For{$j \in \llbracket0, J-1\rrbracket$} \Comment{inner minimization loop}
				\For{$k \in \llbracket0, n-1\rrbracket$}  \Comment{block-coordinate loop}
				\State $(\xbhat_{B, k}^j, \xbhat_{A, k+1}^j) \in \argmin_{\xbhat_{B, k}, \xbhat_{A, k+1}} q(\lambdabhat_k^{t} + \frac{\xbhat^{t}_{B,k} - \xbhat^{t}_{A,k}}{\eta_k})$ \Comment{conditional gradient}
				\State Compute layer optimal step size $\gamma_k^*$ using \eqref{eq:optimal-gamma-gauss}
				\State $(\xbhat_{B, k}^{t}, \xbhat_{A, k+1}^{t}) \leftarrow \gamma_k^* (\zbhat_{B, k}^j, \zbhat_{A, k+1}^j) + (1 - \gamma_k^*) (\xbhat_{B, k}^{t}, \xbhat_{A, k+1}^{t})$
				\EndFor
				\EndFor
				\State Compute $\lambdabhat^{t+1}$ using equation~\eqref{eq:lambda_updates} or \eqref{eq:momentum_updates} \Comment{dual update}
				\EndFor
				\State\Return $q(\lambdabhat^T)$
				\EndFunction
			\end{algorithmic}
		\end{algorithm}
	\end{minipage}%
\end{figure}

\subsubsection{Frank-Wolfe Algorithm}
\label{subsec:FW}
Problem~\eqref{eq:z_updates} can be optimized using the conditional gradient
method, also known as the Frank-Wolfe algorithm~\citep{Frank1956}. The advantage
it provides is that there is no need to perform a projection step to remain in
the feasible domain. Indeed, the different iterates remain in the feasible
domain by construction as convex combination of points in the feasible domain.
At each time step, we replace the objective by a linear approximation and
optimize this linear function over the feasible domain to get an update
direction, named conditional gradient. 
We then take a step in this direction. As the Augmented Lagrangian is smooth over the primal variables, there is no need to take the Frank-Wolfe step for all the network layers at once. We can in fact do it in a block-coordinate fashion, where a block is a network layer, with the goal of speeding up convergence.

\paragraph{Conditional Gradient Computation}
Let us denote iterations for the inner problem by the superscript $j$.
Obtaining the conditional gradient requires minimizing a linearization of $\auglag$ on the primal variables, restricted to the feasible domain:
\begin{equation*}
\begin{aligned}
\left( \zb^{j}, \zbhat^{j} \right) \ &= \argmin_{\xb, \xbhat} \nabla_{\xbhat} \mathcal{L}\left( \xbhat, \lambdabhat^t\right)^T \xbhat \\
\text{s.t. }\quad & \mathcal{P}_0(\xb_0, \xbhat_{A,1}), \\
&\mathcal{P}_k(\xb_k, \xbhat_{B, k}, \xbhat_{A, k+1}) \qquad k \in \left\llbracket1,n-1\right\rrbracket.
\end{aligned}
\end{equation*}
This computation corresponds exactly to the one we do to perform the inner minimization of problem \eqref{eq:nonprox_problem} over $\xb$ and $\xbhat$ in order to compute the supergradient (cf. $\S$\ref{sec:innerminp0}, $\S$\ref{sec:innerminpk}). To make this equivalence clearer, we point out that the linear coefficients of the primal variables will maintain the same form (with the difference that the dual variables are represented as their closed-form update for the following iteration), as \mbox{$\nabla_{\xbhat_{B, k}} \mathcal{L}\left( \xbhat,\lambdabhat^t\right) = \lambdabhat_{k}^{t+1}$} and \mbox{$\nabla_{\xbhat_{A, k+1}}\mathcal{L}\left( \xbhat,\lambdabhat^t\right) = -\lambdabhat_{k}^{t+1}$}.
The equivalence of conditional gradient and supergradient is not particular to our problem. A more general description can be found in the work of~\citet{Bach2015}.

\paragraph{Block-Coordinate Steps}
As the Augmented Lagrangian \eqref{eq:z_updates} is smooth in the primal variables, we can perform the Frank-Wolfe steps in a block-coordinate fashion \citep{Jaggi2013b}.
The conditional gradient computation decomposes over the subproblems~\eqref{eq:pk_sub}, it is therefore natural to consider each $(\xbhat_{B, k}, \xbhat_{A, k+1})$ as a separate variable block.
As the values of the primals at following layers are inter-dependent through the gradient of the Augmented Lagrangian, these block-coordinate updates will result in faster convergence.
For each $k \in \left\llbracket0,n-1\right\rrbracket$, denoting again conditional gradients as $\zbhat$, the Frank-Wolfe steps for the $k$-th block take the following~form:
\begin{equation}
(\xbhat_{B, k}^{j}, \xbhat_{A, k+1}^{j}) = \gamma_k (\zbhat_{B, k}^j, \zbhat_{A, k+1}^j) + (1 - \gamma_k) (\xbhat_{B, k}^{j-1}, \xbhat_{A, k+1}^{j-1}).
\label{eq:fw-update}
\end{equation}
Let us denote by $\xb_{\gamma_k}^j$ a vector of all $0$s, except for the $\xbhat_{A}$ and $\xbhat_{B}$ entries of the $k$-th block, which are set to equation~\eqref{eq:fw-update}.
Due to the structure of problem \eqref{eq:z_updates}, we can compute an optimal step size $\gamma_k^*$ by solving a one dimensional
quadratic problem:
\begin{equation*}
\gamma_k^* \in \argmin_{\gamma_k \in \left[ 0, 1 \right]} \mathcal{L}(\xb_{\gamma_k}^j, \lambdabhat^t).
\end{equation*}
Let us denote by $\text{Clip}_{[0, 1]}$ an operator clipping a value into $[0,1]$, and let us cover corner cases through dummy assignments: $\eta_n=\infty$, and $\xbhat_{B,0} = \zbhat_{B,0} = 0$.
Then, $\gamma_k^*$ is given~by:
\begin{equation}
\gamma_{k} ^*= \underset{[0, 1]}{\text{Clip}}\left(\frac{
	\nabla_{\xbhat_{B, k}} \mathcal{L}\left( \xbhat^j,\lambdabhat^t\right)^T (\xbhat_{B,k}^{j-1} - \zbhat_{B,k}^{j})  + \nabla_{\xbhat_{A, k+1}}\mathcal{L}\left( \xbhat^j,\lambdabhat^t\right)^T (\xbhat_{A,k+1}^{j-1}-\zbhat_{A,k+1}^{j} ) 
}{
	\frac{1}{\eta_k} \left\Vert \zbhat_{B,k}^{j} - \xbhat_{B,k}^{j-1} \right\Vert^2 + \frac{1}{\eta_{k+1}}  \left\Vert\zbhat_{A,k+1}^{j} - \xbhat_{A,k+1}^{j-1}  \right\Vert^2
}\right)
\label{eq:optimal-gamma-gauss}
\end{equation}
Finally, inspired by previous work on accelerating proximal methods \citep{Lin2017,Salzo12}, we apply momentum on the dual updates to accelerate convergence; for details we refer the reader to appendix \ref{sec:prox-appendix}.

\subsection{Comparison to Previous Dual Problems} \label{sec:dual_comp}

We conclude this section by comparing our dual problem \eqref{eq:nonprox_problem} to the duals presented in~$\S$\ref{sec:bounding}, focusing on ReLU activation functions. 

\subsubsection{Lagrangian Relaxation of the non-convex formulation}
We first consider problem \eqref{eq:dj_prob-main} by \citet{Dvijotham2018}. From the high level perspective, our decomposition considers larger subsets of constraints and hence results in a smaller number of dual variables to optimize over. 

\begin{proposition} 
	Assume $\sigma_k(\xbhat_k) = \max(\mathbf{0}, \xbhat_k)$ $\forall\ k \in \left\llbracket1,n-1\right\rrbracket$. Then,  $\max_{\mub, \lambdab} d_O(\mub, \lambdab) = \max_{\lambdabhat} q(\lambdabhat)$, that is, the dual solutions of problem \eqref{eq:nonprox_problem} and problem \eqref{eq:dj_prob-main} coincide.
	Additionally, both dual solutions coincide with the solution of problem \eqref{eq:planet-relax}.
	\label{prop:convergence}
\end{proposition}
\begin{proof}
	Recall that the ReLU version of problem \eqref{eq:planet-relax} is an LP (see \eqref{eq:planet-relu-hull}).
	Due to linear programming duality~\citep{lemarechal2001lagrangian}, $\max_{\lambdabhat} q(\lambdabhat)=p^*$, where $p^*$ denotes the solution of problem \eqref{eq:planet-relax}.
	Moreover, Theorem 2 by \citet{Dvijotham2018} shows that problem \eqref{eq:dj_prob-main} corresponds to the Lagrangian dual of problem \eqref{eq:planet-relax}. Therefore, invoking linear programming duality again, $\max_{\mub, \lambdab} d_O(\mub, \lambdab) = \max_{\lambdabhat} q(\lambdabhat)=p^*$.
\end{proof}
While proposition \ref{prop:convergence} states that problems \eqref{eq:nonprox_problem} and \eqref{eq:dj_prob-main} will yield the same bounds at optimality, this does not imply that the two derivations yield solvers with the same efficiency. 
In fact, we will next prove that, for ReLU-based networks, our
formulation dominates problem \eqref{eq:dj_prob-main}, producing bounds at least as tight based on the same dual variables. 
In fact, problem \eqref{eq:nonprox_problem} operates on a subset of the dual space of problem \eqref{eq:dj_prob-main} that always contains the dual~optimum.

\begin{theorem} 
	Let us assume $\sigma_k(\xbhat_k) = \max(\mathbf{0}, \xbhat_k)$ $\forall\ k \in \left\llbracket1,n-1\right\rrbracket$. For dual point $(\mub, \lambdab)$ of problem \eqref{eq:dj_prob-main} by \citet{Dvijotham2018} yielding bound $d(\mub, \lambdab)$, it holds that $q(\mub) \geq d(\mub, \lambdab)$.
	Furthermore, if $\lambdab'_{n-1} = -W_n^T \mbf{1}$ and $\lambdab'_{k-1} = W_k^T \mub_k$ for $k \in \left\llbracket2,n-1\right\rrbracket$, then $q(\mub) = d(\mub, \lambdab')$.
	\label{theorem}
\end{theorem}
\begin{proof}
	See appendix~\ref{sec:dual_comp_proof}.
\end{proof}
Theorem \ref{theorem} motivates the use of dual \eqref{eq:nonprox_problem} over the general form of problem \eqref{eq:dj_prob-main}. Moreover, it shows that, for ReLU activations, this application of Lagrangian Decomposition coincides with a modified version of problem \eqref{eq:dj_prob-main} with additional equality constraints. 
Such a modification can be also found in the work by \citet[appendix G.1]{Salman2019}. However, its advantages for iterative bounding algorithms and connections to Lagrangian Decomposition were not investigated in their work.

\subsubsection{Propagation-based methods}

We now turn our attention to propagation-based methods (see $\S$\ref{sec:propagation}).

\begin{proposition}
	Let  $\bar{d}_P$ be a lower bound to problem \eqref{eq:noncvx_pb} obtained via a propagation-based bounding algorithm. Then, if $\sigma_k(\xbhat_k) = \max(\mathbf{0}, \xbhat_k)$ $\forall\ k \in \left\llbracket1,n-1\right\rrbracket$, there exist some dual points $\bar{\lambdabhat}$ and $\left(\bar{\mub}, \bar{\lambdab}\right)$ such that $q(\bar{\lambdabhat})=d_O\left(\bar{\mub}, \bar{\lambdab}\right)=\bar{d}_P$, and both $\bar{\lambdabhat}$ and $\left(\bar{\mub}, \bar{\lambdab}\right)$ can be computed at the cost of a backward pass through the network.
	\label{prop:init}
\end{proposition}
\begin{proof} 
	See appendix~\ref{sec:dual-propagation}.
\end{proof}
Proposition \ref{prop:init} shows that both problem \eqref{eq:dj_prob-main} and problem \eqref{eq:nonprox_problem} can be inexpensively initialized via propagation-based algorithms such as CROWN~\citep{Zhang2018} or the one by \citet{Wong2018}. 
We exploit this result in our computational evaluation ($\S$\ref{sec:experiments-incomplete}, $\S$\ref{sec:experiments-complete}).

\section{Better Branching} \label{sec:branching}

In this section, we present a novel branching strategy aimed at branch and bound for neural network verification. 

\subsection{Preliminaries: Approximations of Strong Branching}
Recall that branch and bound discards subproblems when the available lower bound on their minimum becomes positive (see $\S$\ref{sec:bab}). 
Let us denote the employed bounding algorithm by $\mathcal{A}$, and its lower bound for subproblem $p$ as $l_\mathcal{A}(p)$. Moreover, let $c_{d\{h\}}(p)$ be the $h$-th children of $p$ according to branching decision $d$. 
Ideally, we would like to take the branching decision that maximizes the chances that some $c_{d\{h\}}(p)$ is discarded, in order to minimize the size of the branch and bound tree. 
In order to do so, we could compute $l_\mathcal{A}(c_{d\{h\}}(p))$ according to every possible branching decision $d$ and each relative child $h$. In the context of branch and bound for integer programming, this branching strategy is traditionally referred to as \emph{full strong branching}~\citep{Morrison2016}. 
As full strong branching is impractical on the large search spaces encountered in neural network verification, it is usually replaced by an approximation. 
For branching based on input domain splitting, each branching decision $d$ corresponds to an index of the input space, that is: $d := i \in \mathbb{R}^{n_0}$.
In this context, \citet{Bunel2018} replace $\mathcal{A}$ (the bounding algorithm used for subproblem lower bounds\footnote{in the case of \citet{Bunel2018}, this means solving the LP in problem \eqref{eq:planet-relax}.}) by a looser yet inexpensive method. 
Specifically, they rely on the bounding algorithm by \citet{Wong2018}, denoted WK, which is a propagation-based method for ReLUs, where $\bar{\sigma}_k(\xbhat_k) =  \frac{ [\hat{\ub}_k]_+ \odot (\xbhat_k - [\hat{\lb}_k]_-)}{ [\hat{\ub}_k]_+-[\hat{\lb}_k]_-}$, and $\ubar{\sigma}_k(\xbhat_k) =  \frac{ [\hat{\ub}_k]_+ \odot \xbhat_k}{ [\hat{\ub}_k]_+-[\hat{\lb}_k]_-}$ (see $\S$\ref{sec:propagation}).
The resulting branching strategy, termed Smart Branching (SB) by~\citet{Bunel2018}, branches on the input coordinate $i$ such that:
$$i \in \argmax_{i \in \mathbb{R}^{n_0}} \left(\min_{h \in \left\llbracket1,2\right\rrbracket} \bigg\{l_{\text{WK}}(c_{i\{h\}}(p))\bigg\}\right).$$ 
%
However, input-based branching was found to be ineffective on large convolutional networks~\citep{Bunel2020}.
With this in mind, in BaBSR, \citet{Bunel2020} rely on a branching strategy that operates by splitting a ReLU into its two linear phases (see $\S$\ref{sec:babsr}).
The original SB heuristic is unsuitable for ReLU branching, as it requires a number of backward passes linear in the size of the space of branching decisions: in general, $\left(\sum_{k=1}^{n-1}n_k\right) \gg n_0$. 
Therefore, Smart ReLU (SR) branching, the heuristic adopted within BaBSR, approximates strong branching even further.  At the cost of a single backward pass, it assigns scores $\mbf{s}_{\text{SR}}$ and $\mbf{t}_{\text{SR}}$ to all possible branching decisions (that is, to each ReLU):
\begin{equation}
\begin{aligned}
&\bar{\lambdab}_{n-1} = -W_{n}^T\mbf{1}, \hspace{70pt} \bar{\lambdab}_{k-1} = W_k^T \left( \frac{ [\hat{\ub}_k]_+}{ [\hat{\ub}_k]_+-[\hat{\lb}_k]_-}\odot \bar{\lambdab}_k\right) && k \in \left\llbracket2, n-1\right\rrbracket,\\[5pt]
&\mbf{s}_{\text{SR}, k} = 
\left| \begin{array}{l}
\max\left\{ 
0, \bar{\lambdab}_{k} \odot\mbf{b}_k \right\} \\
- \frac{\hat{\ub}_k}{ \hat{\ub}_k -\hat{\lb}_{k}} \odot \bar{\lambdab}_{k} \odot \mbf{b}_k + \frac{ \hat{\ub}_k\odot\hat{\lb}_k}{ \hat{\ub}_k -\hat{\lb}_k} \odot [\bar{\lambdab}_{k}]_+ \end{array} 
\right| \odot \mathds{1}_{\hat{\lb}_k < 0, \hat{\ub}_k > 0} && k \in \left\llbracket1,n-1\right\rrbracket, \\[5pt]
&\mbf{t}_{\text{SR}, k} = 
\left. \begin{array}{l}
\frac{ -\hat{\ub}_k\odot\hat{\lb}_k}{ \hat{\ub}_k -\hat{\lb}_k} \odot [\bar{\lambdab}_{k}]_+ \end{array} 
\right. \hspace{-5pt} \odot \mathds{1}_{\hat{\lb}_k < 0, \hat{\ub}_k > 0} && k \in \left\llbracket1,n-1\right\rrbracket.
\end{aligned}
\label{eq:babsr-scores}
\end{equation}
Then, SR branches on the ReLU having the largest $\mbf{s}_{\text{SR}, k}$ scores or, if such a ReLU's score is below a threshold, the largest backup scores $\mbf{t}_{\text{SR}, k}$. 
This strategy can be seen as an approximation of SB. In fact, scores $\mbf{s}_{\text{SR}, k}$ approximate the change in the bounds by \citet{Wong2018} that would arise from splitting on the ambiguous ReLUs at layer $k$. 
In more detail, they consider the effect of splitting within equation \eqref{eq:bounds-propagation}, without backpropagating the effect on $\bar{\lambdab}_{j}$ for $j \in \left\llbracket1,k-1\right\rrbracket$ via equation \eqref{eq:salman-assignment}. 
The two arguments of the maximum in equation \eqref{eq:babsr-scores} correspond to the blocking and passing cases, whereas the remaining terms represent the ambiguous case.
Backup scores $\mbf{t}_{\text{SR}, k}$, instead, correspond to the product of the Lagrangian multiplier for $\xb_k \leq \bar{\sigma}_k(\xbhat_k)$, and the maximum distance of $\bar{\sigma}_k(\xbhat_k)$ from $\sigma(\xbhat_k)$ (in fact, $\bar{\mbf{b}}_k=\frac{ -\hat{\ub}_k\odot\hat{\lb}_k}{ \hat{\ub}_k -\hat{\lb}_k}$). 
They hence provide a second estimation of the effect that a given ReLU split, through its associated reduction of the feasible space, would have on~bounding.

\subsection{Filtered Smart Branching}

We now present \emph{Filtered Smart Branching} (FSB), our novel strategy for activation splitting. 
\citet{Bunel2020} show that, in spite of its rougher approximation of strong branching, SR significantly outperforms SB on larger networks.
Therefore, it is natural to investigate whether improving SR's approximation quality would further reduce the size of the branch and bound tree.
First, inspired by SB, we replace the maximization within $\mbf{s}_{\text{SR}, k}$ with a minimization. Keeping $\bar{\lambdab}$ as in equation \eqref{eq:babsr-scores}, we obtain:
\begin{equation}
\mbf{s}_{\text{FSB}, k} = 
\left| \begin{array}{l}
\min\left\{ 
0, \bar{\lambdab}_{k} \odot\mbf{b}_k \right\} \\
- \frac{\hat{\ub}_k}{ \hat{\ub}_k -\hat{\lb}_{k}} \odot \bar{\lambdab}_{k} \odot \mbf{b}_k + \frac{ \hat{\ub}_k\odot\hat{\lb}_k}{ \hat{\ub}_k -\hat{\lb}_k} \odot [\bar{\lambdab}_{k}]_+ \end{array} 
\right| \odot \mathds{1}_{\hat{\lb}_k < 0, \hat{\ub}_k > 0} \qquad  k \in \left\llbracket1,n-1\right\rrbracket.
\label{eq:fsb-scores}
\end{equation}
Compared to $\mbf{s}_{\text{SR}, k}$, $\mbf{s}_{\text{FSB}, k}$ is designed to balance the branch and bound tree, prioritizing branching decisions that yield bounding improvements in both children, rather than one of them. 
As for $\mbf{s}_{\text{SR}, k}$, the $\mbf{s}_{\text{FSB}, k}$ scores owe their computational efficiency to their shortsightedness and can be computed at the cost of a single gradient backpropagation. 
However, considering that our bounding algorithms ($\S$\ref{sec:decomposition}) require multiple forward/backward passes over the network, we can afford to marginally increase the branching cost if doing so benefits the quality of split.
We propose a layered approach: we employ scores $\mbf{s}_{\text{FSB}, k}$ and $\mbf{t}_{\text{SR}, k}$ to select the most promising candidate branching choices for each layer. 
Denoting a branching choice by a pair $(k, i) \in \left\llbracket1,n-1\right\rrbracket \times \mathbb{R}^{n_k}$,
we create a set of $O(n)$~candidate choices $D_{\text{FSB}} := \cup_k D_{\text{FSB}, k}$, where:
\begin{equation}
	 D_{\text{FSB}, k} = \left\{\left(k,\ \argmax_{i \in \mathbb{R}^{n_k}} \mbf{s}_{\text{FSB}, k}[i]\right), \enskip \left(k,\ \argmax_{i \in \mathbb{R}^{n_k}} \mbf{t}_{\text{SR}, k}[i]\right)\right\} \qquad k \in \left\llbracket1,n-1\right\rrbracket.
	\label{eq:fsb-prefilter}
\end{equation}
As, in general, $O(n) < \left(\sum_{k=1}^{n-1}n_k\right)$, we can then afford to compute lower bounds for each of the candidates using a fast dual bounding algorithm $\mathcal{A}_{\text{FSB}}$. In our implementation, $\mathcal{A}_{\text{FSB}}$ returns the tightest bounds between CROWN\footnote{As mentioned in $\S$\ref{sec:propagation}, CROWN is a propagation-based bounding algorithm. In particular, for ReLU activations, it employs $\bar{\sigma}_k(\xbhat_k) =  \frac{ [\hat{\ub}_k]_+ \odot (\xbhat_k - [\hat{\lb}_k]_-)}{ [\hat{\ub}_k]_+-[\hat{\lb}_k]_-}$ and $\ubar{\sigma}_k(\xbhat_k) = \left(\mathds{1}_{-\hat{\lb}_{k} \leq \hat{\ub}_{k}} + \mathds{1}_{\hat{\lb}_{k} \geq 0}\right) \xbhat_{k}$.}~\citep{Zhang2018} and the algorithm by \citet{Wong2018}.
FSB splits on the activation determined by:
\begin{equation}
	d_{\text{FSB}} \in \argmax_{d \in D_{\text{FSB}} } \left(\min_{h \in \left\llbracket1,2\right\rrbracket} \bigg\{l_{\mathcal{A}_{\text{FSB}}}(c_{d\{h\}}(p))\bigg\}\right),
\end{equation}
where the $D_{\text{FSB}}$ candidate set is determined via equation \eqref{eq:fsb-prefilter}.
FSB is both conceptually simple and effective in practice.
In fact, we will show in section \ref{sec:experiments-complete} that FSB significantly improves on SR~\citep{Bunel2020}. Moreover, it is strongly competitive with a strategy that mimics strong branching via learning algorithms~\citep{Lu2020Neural}, without its training costs.
Finally, we point out that, while FSB was presented in the context of ReLU branching, the technique generalizes to other activation functions. Let us divide an activation's domain into intervals: $[\hat{\lb}_k, \hat{\ub}_k] =\cup_j[\hat{\lb}_k^j, \hat{\ub}_k^j]$. It is convenient to branch on the intervals if $\text{Conv}(\sigma_k, \hat{\lb}_k, \hat{\ub}_k)$~satisfies: 
\begin{equation}
	\cup_j \left\{\text{Conv}(\sigma_k, \hat{\lb}_k^j, \hat{\ub}_k^j)\right\} \subset \text{Conv} (\cup_j \left\{\text{Conv}(\sigma_k, \hat{\lb}_k^j, \hat{\ub}_k^j)\right\})= \text{Conv}(\sigma_k, \hat{\lb}_k, \hat{\ub}_k).
	\label{eq:assumption-cvxhull}
\end{equation}
Then, in order to apply FSB, it suffices to: (i) adapt $\mbf{s}_{\text{FSB}, k}$ and $\mbf{t}_{\text{SR}, k}$ to the chosen activation's linear bounding functions $\bar{\sigma}_k(\xbhat_k)$ and $\ubar{\sigma}_k(\xbhat_k)$, (ii) choose an appropriate bounding algorithm $\mathcal{A}_{FSB}$. For instance, the work by \citet{Zhang2018} provides suitable $\bar{\sigma}_k(\xbhat_k)$, $\ubar{\sigma}_k(\xbhat_k)$ and $\mathcal{A}_{FSB}$ for both the hyperbolic tangent and sigmoid, which satisfy equation \eqref{eq:assumption-cvxhull}.

\section{Improved Branch and Bound} \label{sec:further-impr}

Having presented the building blocks of our branch and bound framework ($\S$\ref{sec:decomposition}, $\S$\ref{sec:branching}), we now present further technical and implementation details.

\subsection{Additional Branch and Bound Improvements} \label{sec:badnb}

This section presents the remaining details for Branch and Dual Network Bound (BaDNB), our branch and bound framework for neural network verification designed around dual bounding algorithms ($\S$\ref{sec:decomposition}) and Filtered Smart Branching ($\S$\ref{sec:branching}). 
We start from the treatment of intermediate bounds ($\S$\ref{sec:badnb-ib}), then present a simple heuristic to dynamically adapt the bounding tightness within the branch and bound tree ($\S$\ref{sec:badnb-bt}), and describe how to obtain upper bounds on the minimum of each subproblem~($\S$\ref{sec:badnb-ub}).

\subsubsection{Intermediate Bounds} \label{sec:badnb-ib}

Branching on a ReLU at layer $k$ will potentially influence all $\hat{\lb}_j$ and $\hat{\ub}_j$ for $j \in \left\llbracket k+1, n-1 \right\rrbracket$. 
Therefore, BaBSR by \citet{Bunel2020} updates the relevant intermediate bounds after every branching decision, leading to \mbox{$\sum_{k=2}^{n-1}2n_k$} bounding computations per subproblem in the worst case (that is, when the branching is performed on the first layer). For medium-sized convolutional networks,  \mbox{$\sum_{k=2}^{n-1}2n_k$} will be in the order of thousands.
In order to compensate for the large computational expense, BaBSR relies on relatively loose bounding algorithms for intermediate bounds, taking the layer-wise best bounds between the method by \citet{Wong2018} and Interval Bound Propagation~\citep{Mirman2018}. 

Dual bounding algorithms such as ours ($\S$\ref{sec:decomposition}) or the one by \citet{Dvijotham2018} are significantly less expensive than solving the convex hull LP \eqref{eq:planet-relax} to optimality. Nevertheless, the use of dual iterative algorithms for intermediate bounds would be the bottleneck of each branch and bound iteration.
In order to tighten intermediate bounds without incurring significant expenses, and considering its remarkable performance at the cost of a single backward pass (for instance, see $\S$\ref{sec:experiments-incomplete}), we propose to employ CROWN~\citep{Zhang2018}. In particular, we use the layer-wise best bounds between CROWN and the method by \citet{Wong2018}.
Furthermore, as BaDNB employs cheaper last layer bounding than BaBSR, even inexpensive intermediate bounding can make up a large portion of a branch and bound iteration's runtime.
Therefore, in order to maximize the number of visited nodes within a given time, we only compute intermediate bounds at the root of the branch and bound tree. In $\S$\ref{sec:experiments-complete}, we will show that, while it sacrifices the tightness of the bounds, such a choice pays off experimentally.

\subsubsection{Dynamically Adjusting the Tightness of Bounds} \label{sec:badnb-bt}

BaDNB relies on dual bounding algorithms, whose advantage over black-box LP solvers is to quickly reach close-to-optimal bounds (see $\S$\ref{sec:experiments-incomplete}). 
In addition, they allow for massively parallel implementations, which we exploit by computing bounds for a batch of branch and bound subproblems at once ($\S$\ref{sec:implementation}).
However, the level of tightness required for a batch of possible heterogeneous subproblems is not clear in advance.
Due to the structure of duals \eqref{eq:dj_prob-main} and \eqref{eq:nonprox_problem}, the bounding improvement per iteration will decrease as the solver approaches optimality, both in theory and in practice. Therefore, a straightforward solution is to choose a fixed number iterations near the ``knee point" of the curve plotting bounds over iterations for the root of the branch and bound tree. 
However, a similar ``diminishing returns" law holds for the bounding improvement caused by branching as one moves deeper in the branch and bound tree. Therefore, at some depth in the tree, it will be more convenient to invest the computational resources in tighter bounding rather than branching. 
In order to take this into account, we devise a simple heuristic to dynamically adjust tightness for last layer~bounding.

Let us again denote by $\mathcal{A}$ the bounding algorithm used for the subproblem lower bounds, adding a subscript to indicate the employed number of iterations. 
We denote by $t(\mathcal{A}_T)$ the cost of running $\mathcal{A}$ with $T$ iterations.
We start from a relatively inexpensive setting $\mathcal{A}_{T_0}$, and choose $m$ different speed-accuracy trade-offs: $\mathcal{A}_{T_0}, \mathcal{A}_{T_1}, \dots, \mathcal{A}_{T_{m-1}}$, with $t(\mathcal{A}_{T_j}) <t(\mathcal{A}_{T_{j+1}})$ for $j \in \left\llbracket0, m-2 \right\rrbracket$.
Let us denote by $c^{-1}(p)$ the parent of subproblem $p$ and keep an exponential moving average $i(p)$ of the lower bound improvement from parent to child: $i(p) := \alpha \left(l_{\mathcal{A}_{T_j}}(p) - l_{\mathcal{A}_{T_j}} (c^{-1}(p))\right) + (1 - \alpha) i(c^{-1}(p))$, where $\alpha \in [0,1]$. 
Furthermore, let us estimate the tightness of the bounds given by each $\mathcal{A}_{T_j}$ on some test subproblem\footnote{In our implementation, in order to capture the effect of activation splits (see $\S$\ref{sec:branching}), we set $r$ to the first encountered subproblem at a depth of $4$ in the branch and bound tree.} $r$.
We update the bounding algorithm from $\mathcal{A}_{T_j}$ to $\mathcal{A}_{T_{j+1}}$ when the following condition is satisfied: 
\begin{equation*}
	i(p_{\text{min}}) < \left(l_{\mathcal{A}_{T_{j+1}}}(r) - l_{\mathcal{A}_{T_j}}(r)\right) \frac{t(\mathcal{A}_{T_j})}{t(\mathcal{A}_{T_{j+1}})},
\end{equation*}
where $p_{\text{min}}$ denotes the subproblem with the smallest lower bound within the current subproblem batch.
In other words, we increase the number of iterations when an estimation of the associated bounds tightening, normalized by its runtime overhead, exceeds the current expected branching improvement. 

\subsubsection{Upper Bounds} \label{sec:badnb-ub}

Similarly to BaBSR~\citep{Bunel2020}, we compute an upper bound on the minimum of the current subproblem by evaluating the network at an input point $\xb_{0}$ from the lower bound computation. BaBSR's use of LP solvers allows them to evaluate the network at the primal optimal solution of problem \eqref{eq:planet-relax}. 
However, as explained in $\S$\ref{sec:badnb-bt}, in general BaDNB will not run the dual iterative algorithms presented in $\S$\ref{sec:decomposition} to convergence.
Therefore, we will evaluate the network at some feasible yet suboptimal $\xb_{0}$. 
In practice, for supergradient-type methods like algorithm \ref{alg:subg}, we evaluate the network at the last computed inner minimizer from problem~\eqref{eq:typeA_cgd}. For Algorithm \ref{alg:prox}, instead, we evaluate the network at the last $\xb_{0}$ found while optimizing problem \eqref{eq:z_updates}.

\subsection{Implementation Details} \label{sec:implementation}

The calculations involved in the various components of our branch and bound framework correspond to standard linear algebra operations commonly employed during the forward and backward passes of neural networks.
For instance, operations of the form \mbox{$\smash{\xbhat_{k+1} = W_{k+1} \xb_{k} + \mathbf{b}_{k+1}}$} are exactly forward passes of the network, while operations like $\smash{\lambdabhat_{k+1}^T W_{k+1}}$ are analogous to the backpropagation of gradients. 
This makes it possible for us to leverage the engineering efforts
made to enable fast training and evaluation of neural networks, and easily take
advantage of GPU accelerations. 
As an example, when dealing with convolutional layers, we can employ
specialized implementations rather than building the equivalent $W_k$ matrix, which would contain a lot of~redundancy.
\subsubsection{Bounding Algorithms}
We implement propagation-based algorithms ($\S$\ref{sec:propagation}), the algorithm by \citet{Dvijotham2018}, and our methods based on Lagrangian Decomposition ($\S$\ref{sec:decomposition}) within a unified framework, exploiting their common building blocks.
One of the benefits of these dual bounding algorithms is that they are easily parallelizable. 
In fact, when computing the upper and lower bounds for all
the neurons of a layer, there is no dependency between the different problems,
so we are free to solve them all simultaneously in a batched mode. The approach closely mirrors the batching over samples commonly employed for training neural networks.
\subsubsection{Branch and Bound}
For complete verification, the use of branch and bound opens up yet another stream of parallelism. 
In fact, it is possible to batch over subproblems as well, for both branching and bounding.
In detail, a batch is formed by the $B$ subproblems having the lowest lower bound (in $\S$\ref{sec:experiments-complete}, $B$ ranges from $100$ to $1600$, depending on the given experiment). We first compute and execute branching decisions for the batch, then move on to the batch of children, whose size is $2B$ (the branching is binary for ReLU activations). Then, for the branch and bound specifications which require it, intermediate bounds are updated for the entire batch, parallelizing both over the $2B$ subproblems and the neurons of a layer (leading to up to $45000$ bounding computations at once, in our experiments). Finally, we compute lower bounds for the $2B$ subproblems, and get upper bounds as detailed in $\S$\ref{sec:badnb-ub}.

\section{Related Work} \label{sec:related}

The work by \citet{Bunel2018,Bunel2020} presented an unified view of neural network verification, providing an interpretation of state-of-the-art complete verification methods as branch and bound algorithms (see $\S$\ref{sec:bab}).
Such interpretation holds for a wide array of approaches, including SMT solvers~\citep{Ehlers2017,Katz2017}, Mixed Integer Programming (MIP) formulations~\citep{Tjeng2019}, ReLUVal and Neurify~\citep{Wang2018a,Wang2018b}.
By presenting modifications to the search strategy, the bounding process and the branching algorithm, the methods by \citet{Bunel2020} outperform previous complete verifiers by a significant margin, on a variety of standard datasets such as those from \citet{Ehlers2017,Katz2017}. 
Therefore, building upon its success, we started from the framework by \citet{Bunel2020} and presented various improvements to improve its scaling capabilities.

While many of our contributions ($\S$\ref{sec:branching}, $\S$\ref{sec:further-impr}) are to be employed within branch and bound, bounding algorithms ($\S$\ref{sec:decomposition}) can be additionally seen as stand-alone incomplete verifiers (see~$\S$\ref{sec:background}).
Although they cannot verify properties for all problem instances, incomplete verifiers scale significantly better, as they trade speed for completeness.  
So far, we have focused on approaches presenting a dual interpretation. However, incomplete verifiers can be more generally described as solvers for relaxations of problem \eqref{eq:noncvx_pb}.
In fact, explicitly or implicitly, these are all equivalent to propagating a convex domain through the
network to over-approximate the set of reachable values. 
Some approaches~\citep{Ehlers2017,Salman2019} rely on off-the-shelf solvers to
solve accurate relaxations such as Planet (equation \eqref{eq:planet-relu-hull})~\citep{Ehlers2017}, which is the best known
linear-sized approximation of the problem. On the other hand, as Planet and other
more complex relaxations do not have closed form solutions, some researchers have also
proposed easier to solve, looser formulations. 
Many of these fall into the category of propagation-based methods ($\S$\ref{sec:propagation})~\citep{Wong2018,Weng2018,Singh2018a,Zhang2018,Singh2019}, which solve linear relaxations with only two constraints per activation function, yielding large yet inexpensive over-approximations.
Others relaxed the problem even further in order to obtain faster
solutions, either by propagating intervals~\citep{Gowal2018b}, or through abstract interpretation~\citep{Mirman2018}. 
Our bounding algorithms and the one by \citet{Dvijotham2018} are custom dual solvers for the convex hull of the element-wise activation function (Planet, in the ReLU case). 
Finally we point out that, while tighter convex relaxations exist, they involve a
quadratic number of variables or exponentially many constraints. The semi-definite programming method of
\citet{Raghunathan2018}, or the relaxation by~\citet{Anderson2020}, obtained from relaxing strong Mixed Integer Programming formulations, fall in this category. We do not address them~here.


%

\section{Incomplete Verification Experiments} \label{sec:experiments-incomplete}

In this section, we test the speed-accuracy trade-offs of our bounding algorithms in an incomplete verification setting.
In particular, we compare them with various bounding algorithms on an adversarial robustness task, for images of the CIFAR-10 test set~\citep{Krizhevsky2009}. 

\subsection{Experimental Setup}
For each test image, we compute an upper bound on the robustness margin of a network to each possible misclassification. In other words, we upper bound the difference between the ground truth logit and the remaining $9$ logits, under an allowed perturbation $\epsilon_{\text{ver}}$ in infinity norm of the inputs. 
If for any class the upper bound on the robustness margin is negative, then we are certain that the network is vulnerable against that adversarial perturbation. 
We employ a ReLU-based convolutional network used by \citet{Wong2018} and whose structure corresponds to the ``Wide" architecture in table \ref{tab:oval-nets}. 
We train the network against perturbations of a size up to \mbox{$\epsilon_{\text{train}}=8/255$} in $\ell_{\infty}$ norm, and test for adversarial vulnerability on $\epsilon_{\text{ver}} = 12/255$. Adversarial training is performed via the method by~\citet{Madry2018}, based on an
attacker using 50 steps of projected gradient descent to obtain the
samples. Additional experiments for a network trained using standard stochastic
gradient descent and cross entropy, with no robustness-related term in the objective, can be found in appendix~\ref{sec:experiments-incomplete-appendix}.

\subsection{Bounding Algorithms}
We consider the following bounding algorithms:
\begin{itemize}
	\item {\bf IBP} Interval Bound Propagation~\citep{Gowal2018b,Mirman2018}, whose bounds correspond to setting all dual variables to $0$ in dual problems \eqref{eq:dj_prob-main} and \eqref{eq:nonprox_problem}. 
	\item {\bf WK} and {\bf CROWN}, the propagation-based methods by respectively \citet{Wong2018} and \citet{Zhang2018}. Exploiting proposition \ref{prop:init}, the bounds by both algorithms correspond to a specific dual assignment for both problem \eqref{eq:dj_prob-main} and problem~\eqref{eq:nonprox_problem}.
	\item {\bf DSG+} uses supergradient methods on dual \eqref{eq:dj_prob-main}, the method by \citet{Dvijotham2018}. 
	We use the Adam~\citep{Kingma2015}
	updates rules and decrease the step size linearly between two values, similarly to the experiments of \citet{Dvijotham2018}. We
	experimented with other step size schedules, like constant step size or
	$\frac{1}{t}$ schedules, which all performed worse. 
	\item  {\bf Dec-DSG+} is a direct application of Theorem~\ref{theorem}: it obtains a dual point $(\mub, \lambdab)$ by optimizing problem \eqref{eq:dj_prob-main} via DSG+ and then evaluates $q(\mub)$ in problem~\eqref{eq:nonprox_problem} to obtain the final bounds.
	\item {\bf Supergradient} is the first of the two solvers presented ($\S$\ref{subsec:subgradient_method}), using a supergradient method on
	problem~\eqref{eq:nonprox_problem}. As for DSG+, we use Adam updates and linearly decay the step size.
	\item {\bf Proximal} is the solver presented in $\S$\ref{sec:prox}, performing proximal maximization on problem~\eqref{eq:nonprox_problem}.
	We use a small fixed number of iterations for the inner problems  (specifically, we set $J=2$ in algorithm \ref{alg:prox}).
	\item {\bf Gurobi} is our gold standard method. It employs the commercial black
	box solver Gurobi to solve problem \eqref{eq:planet-relax} to optimality. We make use of LP incrementalism (warm-starting): as the experiment involves computing $9$ different output upper bounds, we warm-start each LP from the LP of the previous neuron.
	\item {\bf Gurobi-TL} is the time-limited version of the above, which stops at the first dual simplex iteration for which the total elapsed time exceeded that required by $400$ iterations of the proximal method.
\end{itemize}
\begin{figure*}[b!]
	\centering
	\begin{subfigure}{\textwidth}
		\includegraphics[width=1\textwidth]{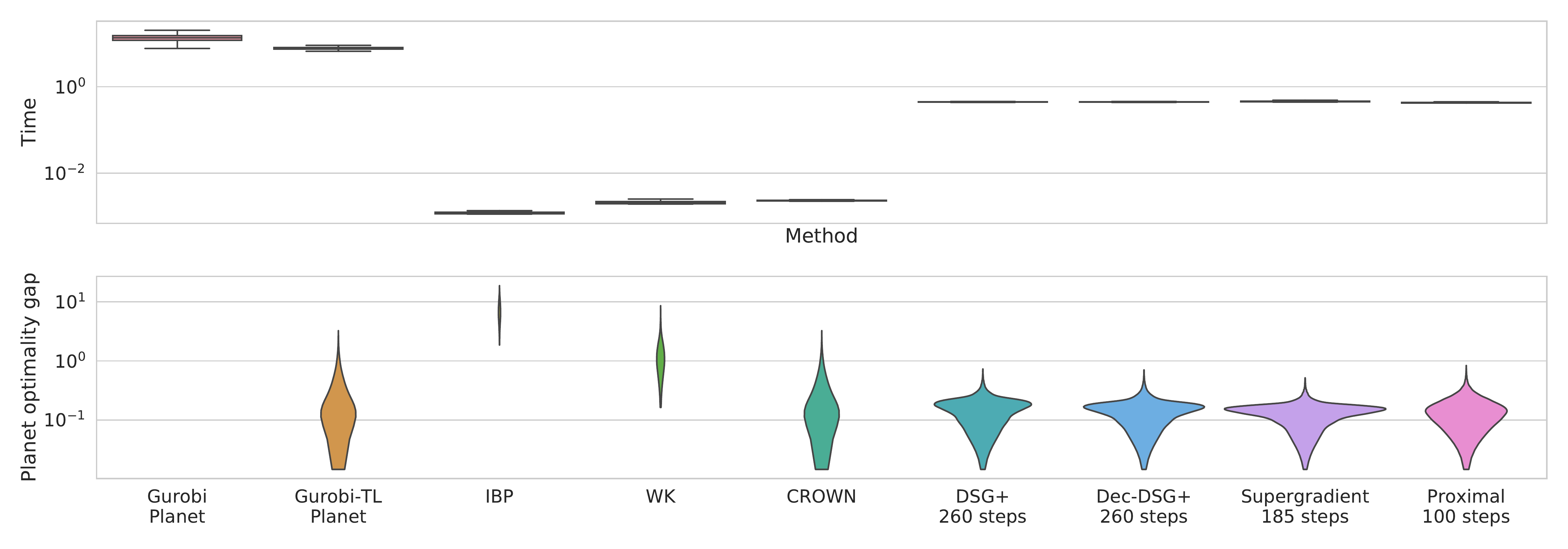}
		\vspace{-15pt}
	\end{subfigure}
	\begin{subfigure}{\textwidth}
		\includegraphics[width=1\textwidth]{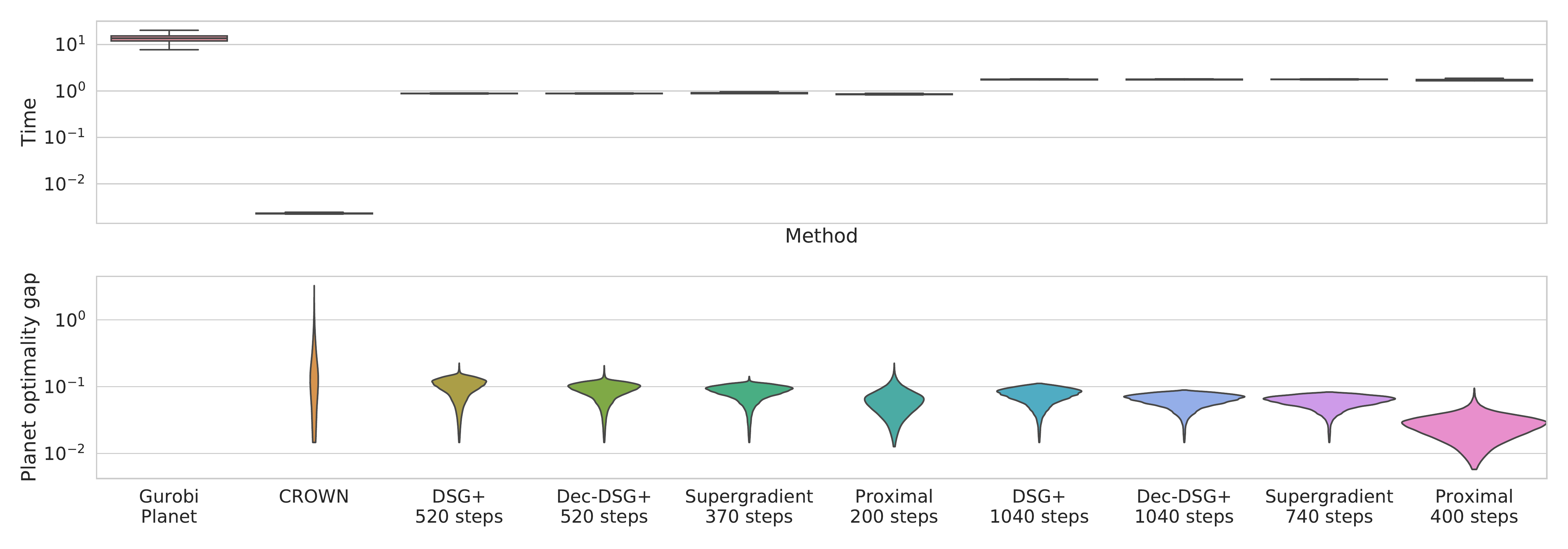}
	\end{subfigure}
	\vspace{-5pt}
	\caption{\label{fig:madry8} 
		Distribution of runtime
		and gap to optimality on a network adversarially trained with the method by \citet{Madry2018}, on CIFAR-10 images. In both
		cases, lower is better. The width at a given value represents the
		proportion of problems for which this is the result. Gurobi Planet always returns the optimal solution to problem \eqref{eq:planet-relax}, at a large computational cost.}
\end{figure*}   
Exploiting proposition \ref{prop:init}, all dual iterative algorithms (Proximal, Supergradient, and DSG+) are initialized from CROWN, which usually outperforms other propagation-based algorithms on ReLU networks~\citep{Zhang2018}.
In all cases, we pre-compute intermediate bounds (see $\S$\ref{sec:intermediate}) using the layer-wise best amongst CROWN and WK. This reflects the bounding schemes used within branch and bound for networks of comparable size~\citep{Bunel2020,Lu2020Neural}.
Hyper-parameters were tuned on a small subset of the CIFAR-10 test set. For both supergradient methods (Supergradient, and DSG+), we decrease the step size linearly from $10^{-2}$ to $10^{-4}$. For Proximal, we employ momentum coefficient $\mu=0.3$ (see appendix \ref{sec:prox-appendix}) and, for all layers, linearly increase the weight of the proximal terms from $10^1$ to $5\times10^2$. 
Because of their small cost per iteration, dual iterative methods allow the user to choose amongst a variety of trade-offs between tightness and speed.                                                                          
In order to perform a fair comparison, we fixed the number of iterations for the various methods so that each of them would take the same average time,. This was done by tuning the iteration ratios on a subset of the images. We report results for three different computational budgets.
Note that the Lagrangian Decomposition has a higher cost per iteration due to its more complex primal feasible set. The cost of the proximal method is even larger, as it requires an iterative procedure for the inner minimization \eqref{eq:z_updates}.
All methods were implemented in PyTorch~\citep{Paszke2019} and run on a single Nvidia Titan V GPU, except those based on Gurobi, which were run on $4$ cores of i9-7900X CPUs.
The amenability of dual methods do GPU acceleration is a big part of their advantages over off-the-shelf solvers.
Experiments were run under Ubuntu 16.04.2~LTS.

\subsection{Results}                                           

\begin{figure*}[t!]
	\centering
	\begin{minipage}{.48\textwidth}
		\centering
		\includegraphics[width=\textwidth]{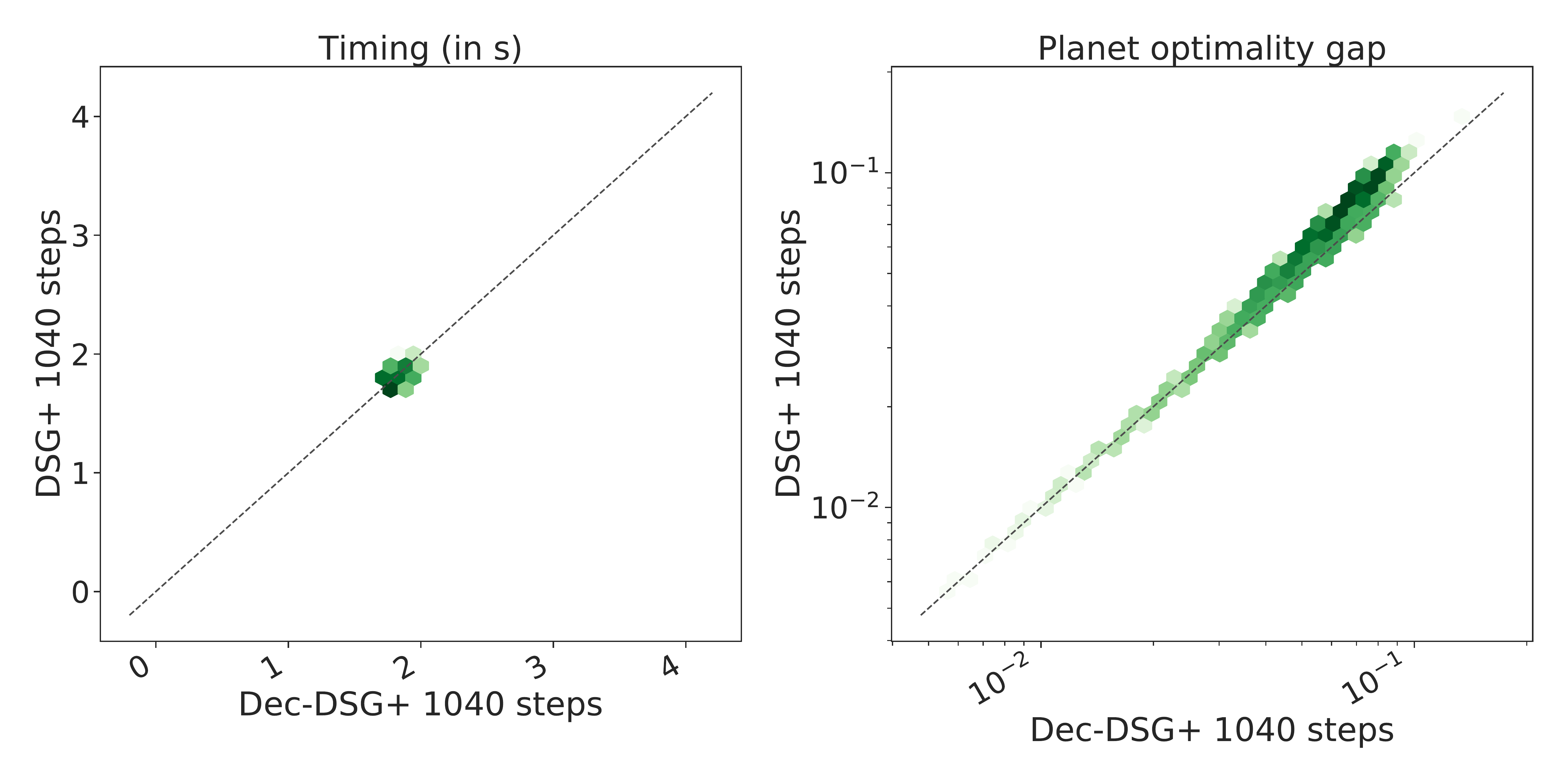}
	\end{minipage}
	\hspace{5pt}
	\begin{minipage}{.48\textwidth}
		\centering
		\includegraphics[width=\textwidth]{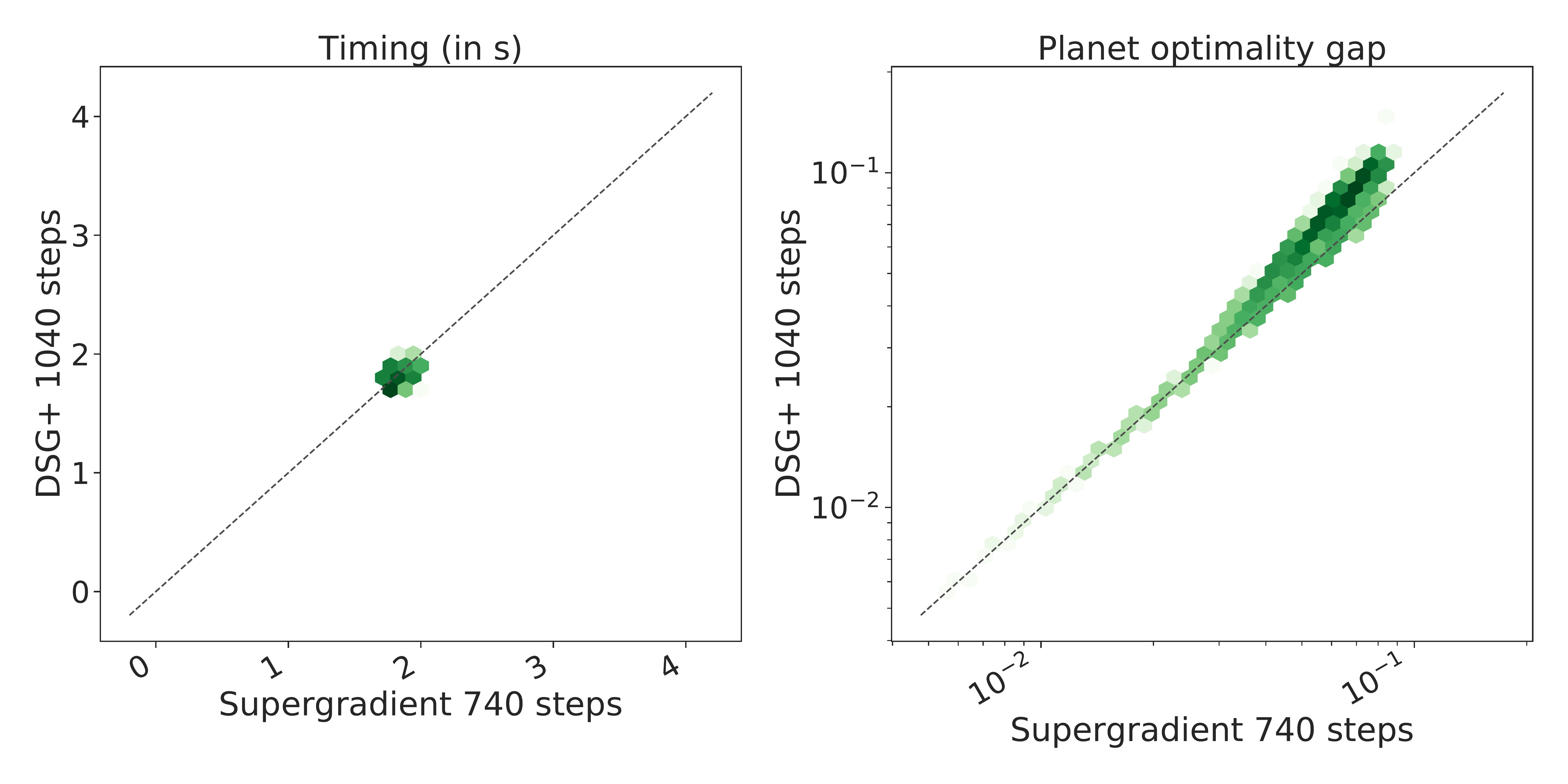}
	\end{minipage}
	\vspace{5pt}
	\begin{minipage}{.48\textwidth}
		\centering
		\includegraphics[width=\textwidth]{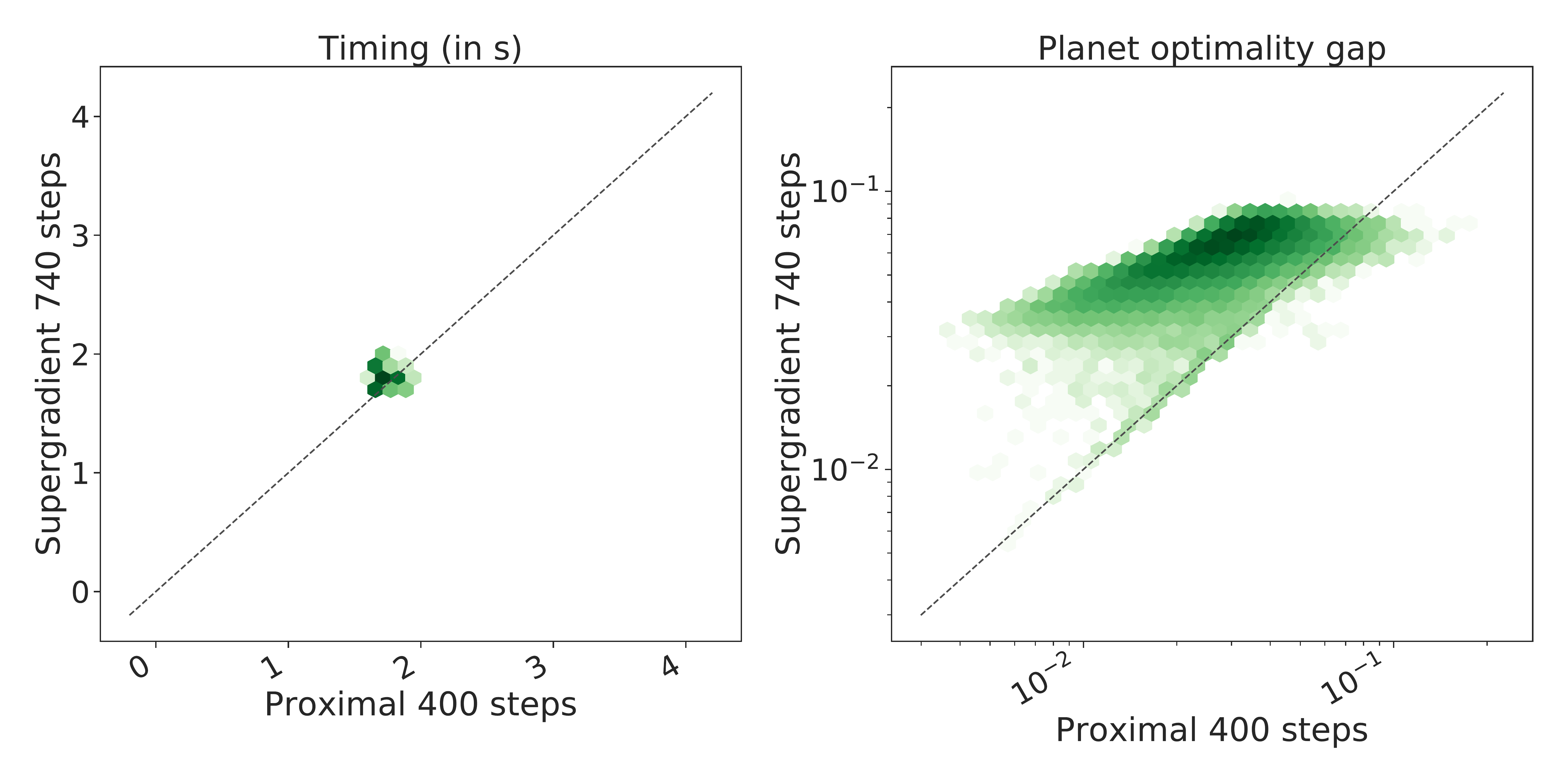}
	\end{minipage}
	\hspace{5pt}
	\begin{minipage}{.48\textwidth}
		\centering
		\includegraphics[width=\textwidth]{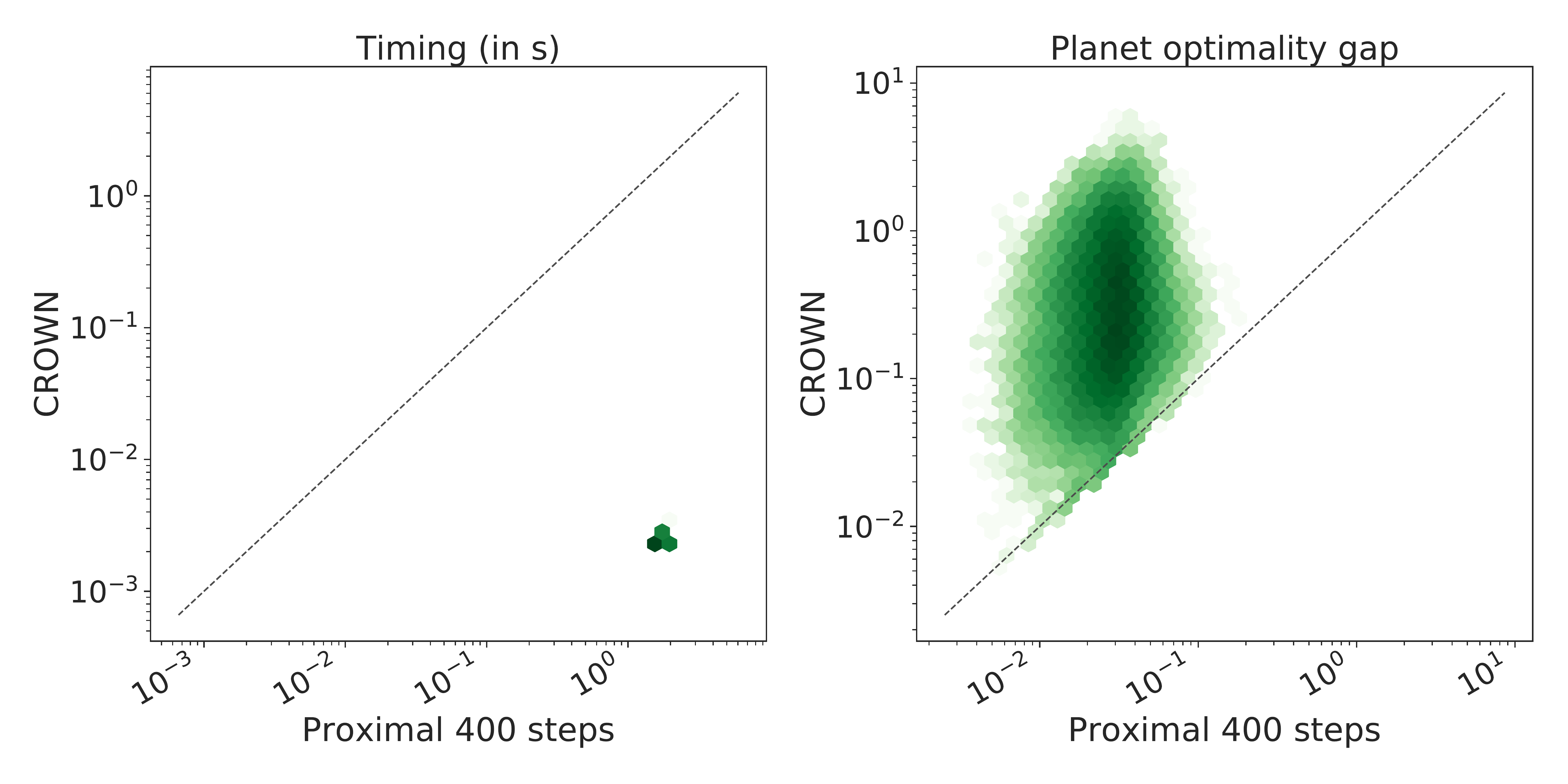}
	\end{minipage}                                                                                                 
	\vspace{-15pt}                                                                                                 
	\caption{Pointwise comparison for a subset of the methods on the data presented in Figure \ref{fig:madry8}. Each datapoint corresponds to a CIFAR image, darker color shades mean higher point density in a logarithmic scale. The dotted line corresponds to the equality and in both graphs, lower is better along both axes.}
	\label{fig:madry8-pointwise}
\end{figure*} 

We measure the time to compute last layer bounds, and report the gap to the optimal solution of problem \eqref{eq:planet-relax}, which is based on the Planet relaxation \eqref{eq:planet-relu-hull} for our ReLU benchmark.                                                                             
Figure \ref{fig:madry8} presents the distribution of results for all bounding algorithms. 
The fastest method is by IP, which requires only a few linear algebra operations over the last network layer. However, the bounds it returns are consistently very loose.
WK and CROWN have a similarly low computational cost: they both require a single backward pass through the network per optimization problem. Nevertheless, CROWN generates much tighter average bounds, and is therefore the best candidate for the initialization of dual iterative algorithms.
At the opposite end of the spectrum, Gurobi is extremely slow but provides the best achievable bounds. 
Furthermore, time-limiting the LP solver significantly worsens the produced bounds without a noticeable cut in runtimes. This is due to the high cost per iteration of the dual simplex algorithm.
For DSG+, Supergradient and Proximal, the improved quality of the bounds as compared to IP, WK and CROWN shows that there are benefits in actually solving the relaxation rather than relying on approximations. In particular, a few iterations of the iterative algorithms cuts away the looser part of CROWN's bounds distribution, and an increased computational budget leads to significantly better average bounds.
While the relative cost of dual solvers over CROWN (three order of magnitudes more time) might seem disproportionate, we will see in $\S$\ref{sec:complete-exp-lastlayer} that the gain in tightness is crucial for faster complete verification.
Furthermore, by looking at the point-wise comparisons in Figure \ref{fig:madry8-pointwise}, we can see that Supergradient yields consistently better bounds than \textsc{DSG+}. As both methods employ the Adam update rule (and the same hyper-parameters, which are optimal for both), we can conclude that operating on the Lagrangian Decomposition dual \eqref{eq:noncvx_pb} produces better speed-accuracy trade-offs compared to the dual \eqref{eq:dj_prob-main} by \citet{Dvijotham2018}.
This is in line with the expectations from Theorem \ref{theorem}. Moreover, while a direct application of the Theorem (Dec-DSG+) does improve on the DSG+ bounds, operating on the Decomposition dual \eqref{eq:noncvx_pb} is empirically more effective.
Finally, on average, the Proximal yields better bounds than those returned by Supergradient, further improving on the DSG+ baseline. In particular, we stress that the support of optimality gap distribution is larger for Proximal, with a heavier tail towards better bounds. 

\section{Complete Verification} \label{sec:experiments-complete}

We now evaluate our branch and bound framework and its building blocks within complete verification.
As detailed in $\S$\ref{sec:complete-exp-setup}, we focus on proving (or disproving) a network's adversarial robustness. 
We first study the effect of each presented branch and bound component ($\S$\ref{sec:complete-exp-lastlayer}, $\S$\ref{sec:complete-exp-branching}, $\S$\ref{sec:complete-exp-ib}), then compare BaDNB with state-of-the-art complete verifiers in~$\S$\ref{sec:complete-exp-final}.

\subsection{Experimental Setup} \label{sec:complete-exp-setup}

\begin{table}[t!]
	\centering
	\small
	\begin{adjustbox}{max width=\textwidth, center}
		\begin{tabular}{|c|c|c|}
			\hline
			\textbf{Network Specifications}\TBstrut & \textbf{Verification Properties} \TBstrut & \textbf{Network Architecture} \TBstrut\\
			\hline
			\begin{tabular}{ll}
				Name: & Base \\
				Activation: & ReLU \\
				Training method: & WK \\
				Total activations: & 3172
			\end{tabular}  & \begin{tabular}{cl}
				$100$ properties \\[7pt]
				$\epsilon_{\text{ver}} \in [ 0.061, 0.253]$
			\end{tabular}  & \begin{tabular}{@{}c@{}} 
				\footnotesize Conv2d(3,8,4, stride=2, padding=1) \Tstrut \\
				\footnotesize Conv2d(8,16,4, stride=2, padding=1)\\
				\footnotesize linear layer of 100 hidden units \\
				\footnotesize linear layer of 10 hidden units\Bstrut
			\end{tabular}\\
			\hline
			\begin{tabular}{ll}
				Name: & Wide \\
				Activation: & ReLU \\
				Training method: & WK \\
				Total activations: & 6244
			\end{tabular} & \begin{tabular}{cl}
				$100$ properties \\[7pt]
				$\epsilon_{\text{ver}} \in [ 0.061, 0.216]$
			\end{tabular}  & \begin{tabular}{@{}c@{}} 
				\footnotesize Conv2d(3,16,4, stride=2, padding=1) \Tstrut\\
				\footnotesize Conv2d(16,32,4, stride=2, padding=1)\\
				\footnotesize linear layer of 100 hidden units \\
				\footnotesize linear layer of 10 hidden units\Bstrut
			\end{tabular}\\
			\hline
			\begin{tabular}{ll}
				Name: & Deep \\
				Activation: & ReLU \\
				Training method: & WK \\
				Total activations: & 6756
			\end{tabular} & \begin{tabular}{cl}
				$100$ properties \\[7pt]
				$\epsilon_{\text{ver}} \in [ 0.061, 0.259]$
			\end{tabular} & \begin{tabular}{@{}c@{}} 
				\footnotesize Conv2d(3,8,4, stride=2, padding=1) \Tstrut \\
				\footnotesize Conv2d(8,8,3, stride=1, padding=1) \\
				\footnotesize Conv2d(8,8,3, stride=1, padding=1) \\
				\footnotesize Conv2d(8,8,4, stride=2, padding=1)\\
				\footnotesize linear layer of 100 hidden units \\
				\footnotesize linear layer of 10 hidden units\Bstrut
			\end{tabular}\\
			\hline
		\end{tabular}
	\end{adjustbox}
	\caption{\label{tab:oval-nets} Specifications of the OVAL dataset, a subset of the CIFAR-10 complete verification dataset from \citet{Lu2020Neural}. Each verification property is associated to a different $\epsilon_{\text{ver}}$ value. WK denotes the adversarial training algorithm by \citet{Wong2018}.}
\end{table}

\begin{table}[b!]
	\centering \small
	\begin{adjustbox}{max width=\textwidth, center}
		\begin{tabular}{|c|c|c|c|}
			\hline
			\textbf{Network Specifications}\TBstrut & \textbf{Verification Properties} \TBstrut & \textbf{Network Architecture} \TBstrut\\
			\hline
			\begin{tabular}{ll}
				Name: & $2/255$ \\
				Activation: & ReLU \\
				Training method: &COLT \\
				Total activations: & 16643
			\end{tabular}
			& \begin{tabular}{c}
				$82$ properties \\[7pt]
				$\epsilon_{\text{ver}} = \frac{2}{255}$
			\end{tabular} & \begin{tabular}{@{}c@{}} 
				\footnotesize Conv2d(3, 32, 3, stride=1, padding=1) \Tstrut \\
				\footnotesize Conv2d(32,32,4, stride=2, padding=1)\\
				\footnotesize Conv2d(32,128,4, stride=2, padding=1)\\
				\footnotesize linear layer of 250 hidden units \\
				\footnotesize linear layer of 10 hidden units \Bstrut
			\end{tabular}\\
			\hline
			\begin{tabular}{ll}
				Name: & $8/255$ \\
				Activation: & ReLU \\
				Training method: &COLT \\
				Total activations: & 49411
			\end{tabular} & \begin{tabular}{c}
				$56$ properties \\[7pt]
				$\epsilon_{\text{ver}} = \frac{8}{255}$
			\end{tabular}  & \begin{tabular}{@{}c@{}} 
				\footnotesize Conv2d(3, 32, 5, stride=2, padding=2) \Tstrut \\
				\footnotesize Conv2d(32,128,4, stride=2, padding=1)\\
				\footnotesize linear layer of 250 hidden units \\
				\footnotesize linear layer of 10 hidden units \Bstrut
			\end{tabular}\\
			\hline
		\end{tabular}
	\end{adjustbox}
	\caption{\label{tab:eth-nets} Specifications of the COLT-based~\citep{Balunovic2020} CIFAR-10 complete verification dataset.}
\end{table}

Tables \ref{tab:oval-nets} and \ref{tab:eth-nets} present the details of the two verification datasets on which we conduct our experimental evaluation. 
For both, the goal is to verify whether a network is robust to $\ell_{\infty}$ norm perturbations of radius $\epsilon_{\text{ver}}$ on images of the CIFAR-10~\citep{Krizhevsky2009} test set. For each complete verifier, we measure the time to termination, limited at one hour. In case of timeouts, the runtime is replaced by such time limit. All experiments were run under Ubuntu 16.04.2 LTS. 
The dataset by \citet{Lu2020Neural}, which we name ``OVAL", was introduced to test their novel GNN-based branching algorithm (table \ref{tab:oval-nets}). 
It consists of three different ReLU-based convolutional networks of varying size, which were adversarially trained using the algorithm by \citet{Wong2018}. 
For each network, it associates an incorrect class and a perturbation radius $\epsilon_{\text{ver}}$ to a subset of the CIFAR-10 test images.
The radii $\epsilon_{\text{ver}}$ are found via a binary search, and are designed to ensure that each problem meets a certain problem difficulty when verified by BaBSR~\citep{Bunel2020}. 
As a consequence, the dataset lacks properties that are easily verifiable regardless of the employed algorithm, or that are hardly verified by any method. Therefore, we believe it is an effective testing ground for complete verifiers.
These properties can be represented in the canonical form of problem \eqref{eq:noncvx_pb} by setting $\hat{x}_n$ to be the difference between the ground truth logit and the target logit.
In order to complement the dataset by \citet{Lu2020Neural}, we additionally experiment on two larger networks trained using COLT, the recent adversarial training scheme by \citet{Balunovic2020} (table \ref{tab:eth-nets}).
In this case, we employ a fixed $\epsilon_{\text{ver}}$, chosen to be the radius employed for training. We focus on the first $100$ elements of the CIFAR-10 testset, excluding misclassified images.
Differently from the dataset by \citet{Lu2020Neural}, the goal is to verify robustness with respect to any misclassification. The properties can be converted to the canonical form by suitably adding auxiliary layers~\citep{Bunel2020}.

\subsection{Bounding Algorithms} \label{sec:complete-exp-lastlayer}

The computational bottleneck of BaBSR was found to be bounding algorithm~\citep{Bunel2020}.
Therefore, we start our experimental evaluation by examining the effect of replacing Gurobi Planet, the bounding algorithm from BaBSR, with some of the dual methods evaluated in $\S$\ref{sec:experiments-incomplete}.
Specifically, we will employ DSG+, Supergradient and Proximal amongst dual iterative algorithms. 
For each of these three methods, we initialize the problem relative to each subproblem with the dual variables from the parent's bounding, and with CROWN for the root subproblem.
Additionally, we compare against \textbf{WK + CROWN}, which returns the best bounds amongst WK and CROWN, as representative of propagation-based methods.
As explained in section \ref{sec:implementation}, due to the highly parallelizable nature of the dual algorithms, we are able to compute lower bounds for multiple subproblems at once for DSG+, Supergradient, Proximal, and WK + CROWN. 
In detail, the number of simultaneously solved subproblems is $300$ for the Base network, and $200$ for the Wide and Deep networks.
For Gurobi, which does not support GPU acceleration, we improve on the original BaBSR implementation by \citet{Bunel2020} by computing bounds relative to different subproblems in parallel over the CPUs.
For both supergradient methods (our Supergradient, and DSG+), we decrease the step size linearly from $\alpha=10^{-3}$ to $\alpha=10^{-4}$: the initial step size is smaller than in $\S$\ref{sec:experiments-incomplete} to account for the parent initialization.
For Proximal, we do not employ momentum and keep the weight of the proximal terms fixed to $\eta=10^2$ for all layers throughout the iterations. As in the previous section, the number of iterations for the bounding algorithms are tuned to employ roughly the same time: we use $100$ iterations for Proximal, $160$ for Supergradient, and $260$ for DSG+.
Dual bounding computations (for both intermediate and subproblem bounds) were run on a single Nvidia Titan Xp GPU. 
Gurobi was run on i7-6850K CPUs, using $6$ cores.

\begin{figure}[t!]
	\sisetup{detect-weight=true,detect-inline-weight=math,detect-mode=true}
	\begin{subtable}{\textwidth}
		\centering
		\label{table:lastlayer}
		
		\scriptsize
		\setlength{\tabcolsep}{4pt}
		\aboverulesep = 0.1mm  
		\belowrulesep = 0.2mm  
		\renewcommand{\bfseries}{\fontseries{b}\selectfont}
		\newrobustcmd{\B}{\bfseries}
		\newcommand{\boldentry}[2]{%
			\multicolumn{1}{S[table-format=#1,
				mode=text,
				text-rm=\fontseries{b}\selectfont
				]}{#2}}
		\begin{adjustbox}{max width=\textwidth, center}
			\begin{tabular}{
					l
					S[table-format=4.3]
					S[table-format=4.3]
					S[table-format=4.3]
					S[table-format=4.3]
					S[table-format=4.3]
					S[table-format=4.3]
					S[table-format=4.3]
					S[table-format=4.3]
					S[table-format=4.3]
				}
				& \multicolumn{3}{ c }{Base} & \multicolumn{3}{ c }{Wide} & \multicolumn{3}{ c }{Deep} \\
				\toprule
				
				\multicolumn{1}{ c }{Method} &
				\multicolumn{1}{ c }{time [s]} &
				\multicolumn{1}{ c }{subproblems} &
				\multicolumn{1}{ c }{$\%$Timeout} &
				\multicolumn{1}{ c }{time [s]} &
				\multicolumn{1}{ c }{subproblems} &
				\multicolumn{1}{ c }{$\%$Timeout} &
				\multicolumn{1}{ c }{time [s]} &
				\multicolumn{1}{ c }{subproblems} &
				\multicolumn{1}{ c }{$\%$Timeout} \\
				
				\cmidrule(lr){1-1} \cmidrule(lr){2-4} \cmidrule(lr){5-7} \cmidrule(lr){8-10}
				
				\multicolumn{1}{ c }{\textsc{DSG+}}
				& 812.84 &    135686.10 &    20.00 &   638.63 &     73591.82 &    13.00 &   257.01 &     21928.52 &     4.00 \\
				
				\multicolumn{1}{ c }{\textsc{Supergradient}}
				&  \B 776.35 &    147347.26 &    \B 19.00 &   561.50 &     74274.48 &    12.00 &   228.67 &     14851.40 &    \B   3.00 \\
				
				\multicolumn{1}{ c }{\textsc{Proximal}}
				& 808.62 &    166478.44 &    20.00 &   \B 498.69 &     80703.96 &   \B   10.00 &   \B 206.54 &     18139.62 &   \B    3.00 \\
				
				\multicolumn{1}{ c }{\textsc{wk + crown}}
				& 2417.83 &    725627.96 &    66.00 &  3042.91 &    736415.20 &    84.00 &  2055.09 &    382353.30 &    53.00 \\
				
				\multicolumn{1}{ c }{\textsc{Gurobi Planet}}
				& 1352.15 &      7013.72 &    25.00 &  1236.54 &       2737.36 &    17.00 &   782.84 &      848.32 &     7.00 \\
				
				\bottomrule
				
			\end{tabular}
		\end{adjustbox}
		\subcaption{\small Comparison of average solving time, average number of solved subproblems and the percentage of timed out properties. The best performing method is highlighted in bold.}
		\vspace{5pt}
	\end{subtable}
	\begin{subfigure}{\textwidth}
		\centering
		\begin{subfigure}{.32\textwidth}
			\centering
			\includegraphics[width=\textwidth]{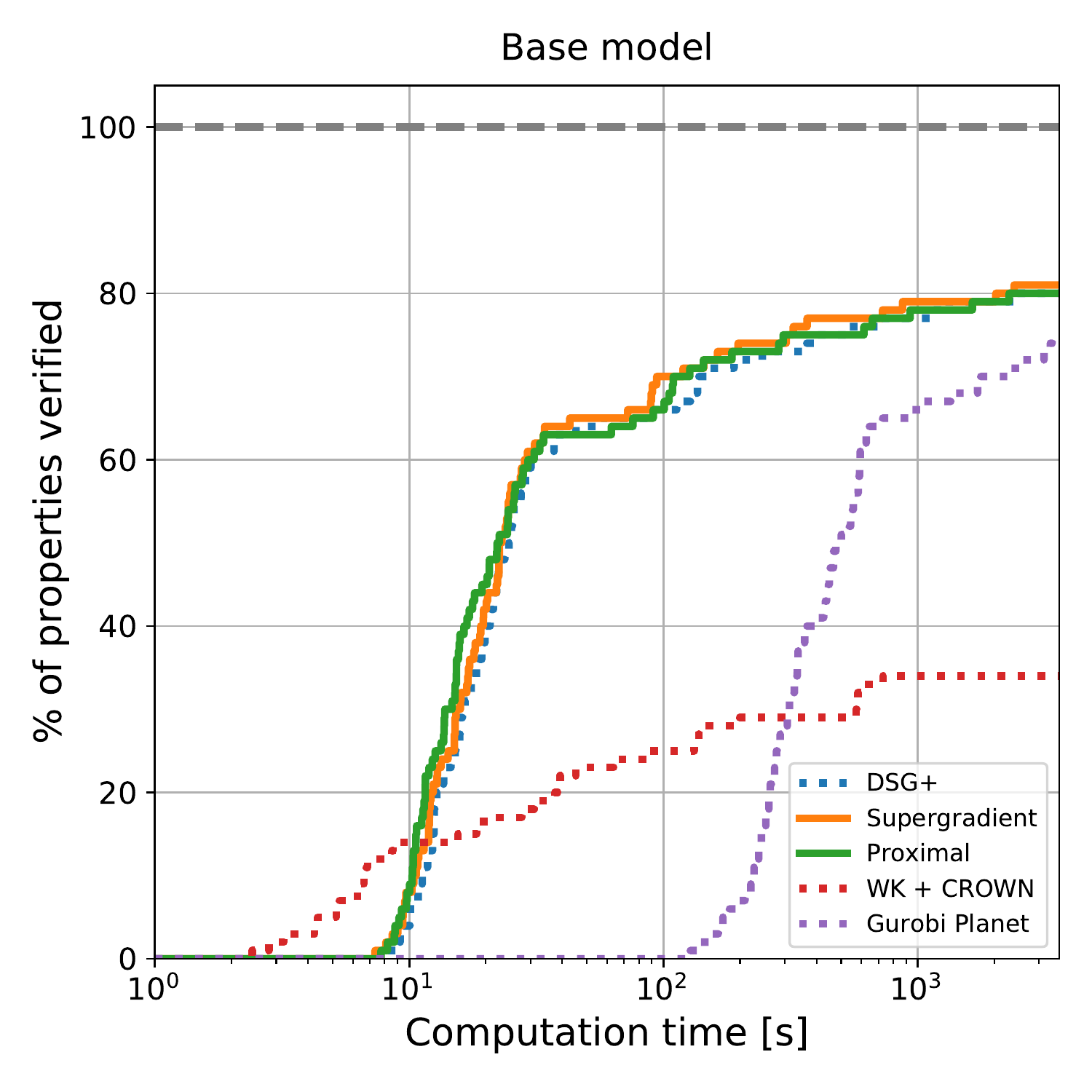}
		\end{subfigure}
		\begin{subfigure}{.32\textwidth}
			\centering
			\includegraphics[width=\textwidth]{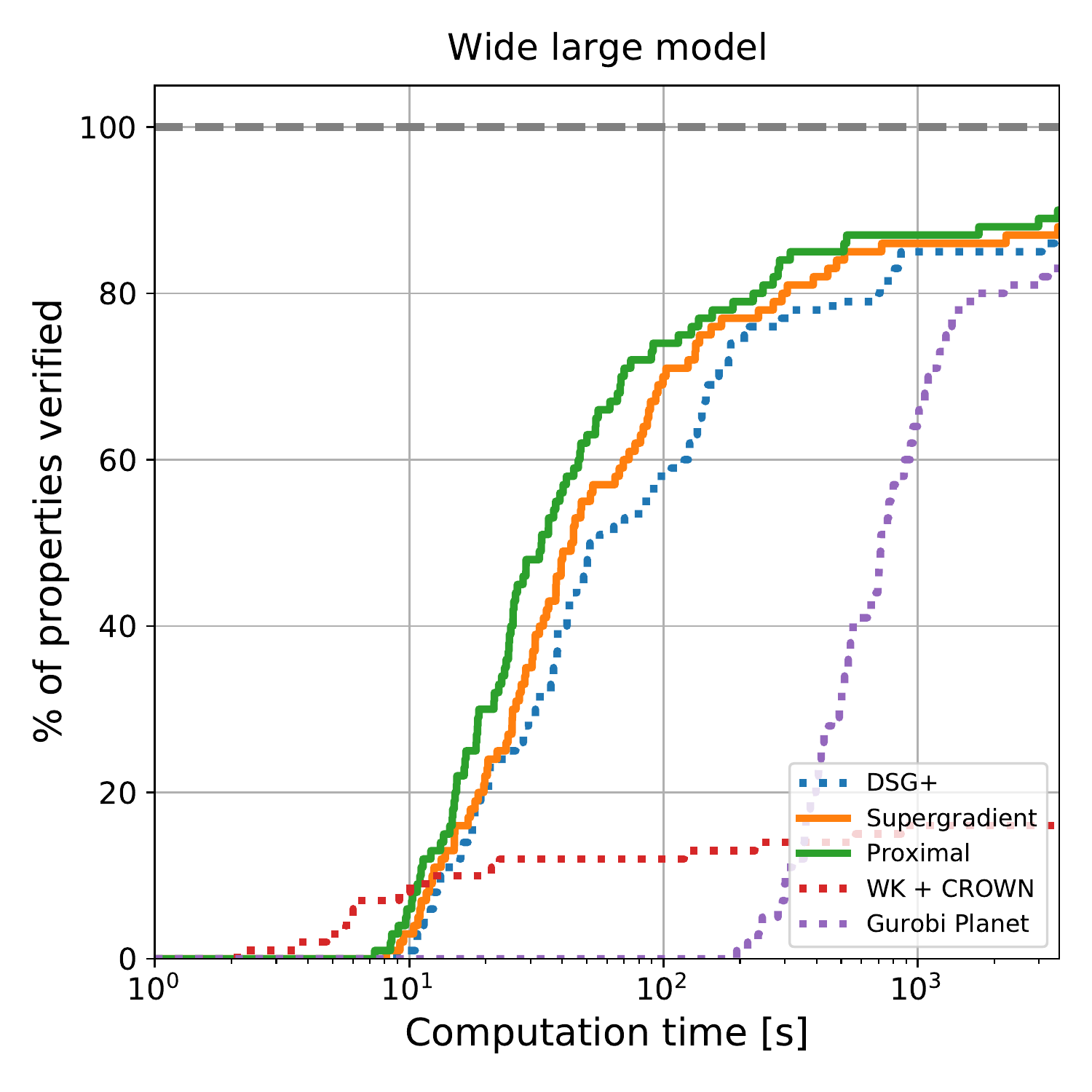}
		\end{subfigure}
		\begin{subfigure}{.32\textwidth}
			\centering
			\includegraphics[width=\textwidth]{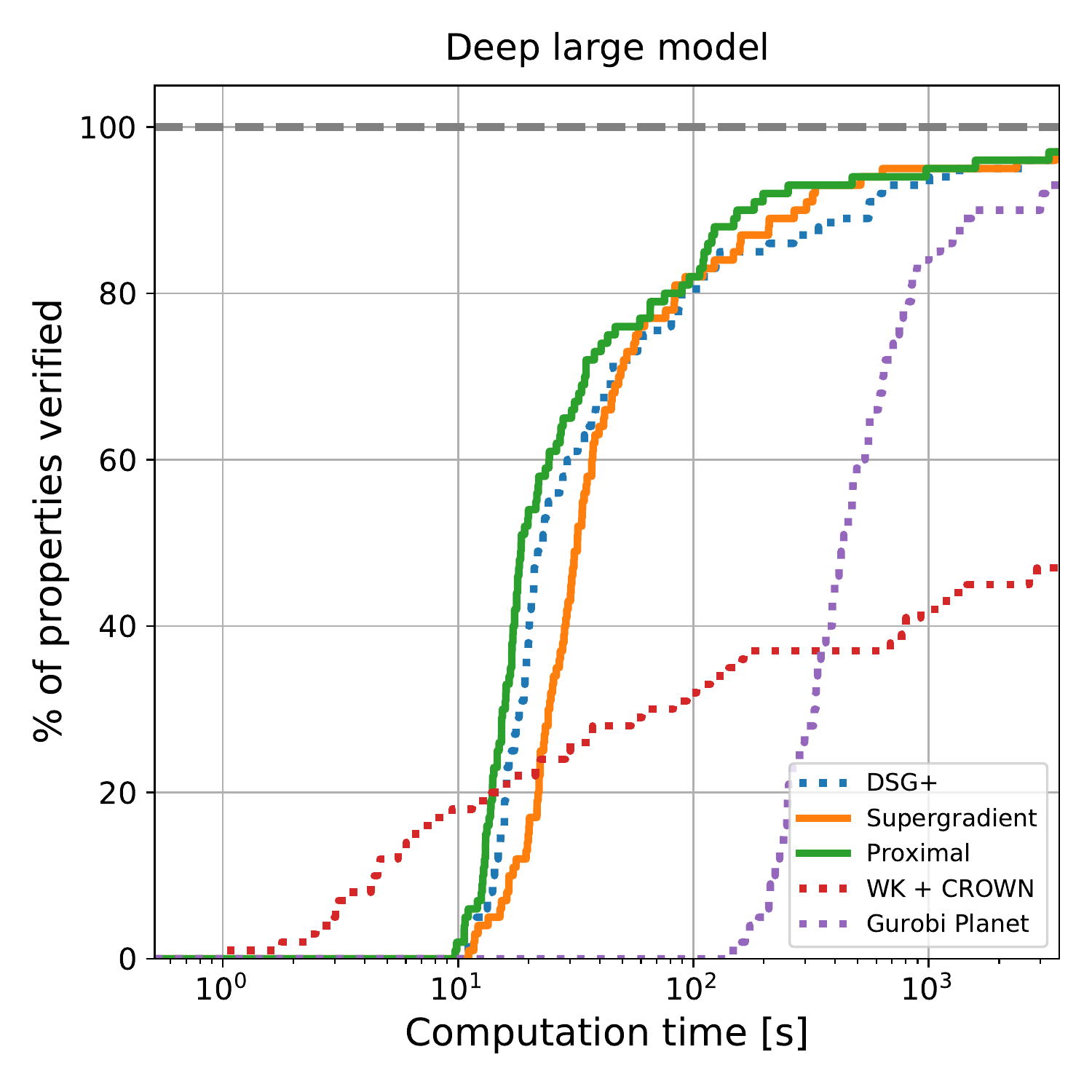}
		\end{subfigure}
		\vspace{-5pt}
		\subcaption{Cactus plots: percentage of solved properties as a function of runtime. Baselines are represented by dotted lines.}
	\end{subfigure}
	\caption{Complete verification performance of different bounding algorithms within BaBSR, on the OVAL dataset. }
	\label{fig:lastlayer}
\end{figure}

Figure \ref{fig:lastlayer} shows that the use of GPU-accelerated dual iterative algorithms is highly beneficial within complete verification. In fact, in spite of the loss in tightness compared to Gurobi, the efficiency of the implementation and the convenient speed-accuracy trade-offs significantly speed up the verification process.
On the other hand, the bounds returned by propagation-based methods are too loose to be effectively employed for the last layer bounding, and lead to a very large number of timed out properties. 
The larger performance difference with respect to incomplete verification ($\S$\ref{sec:experiments-incomplete}) is explained by the fact that dual information cannot be propagated from parent to child. Dual solvers, instead, compared to one-off approximations like WK + CROWN, can more effectively exploit the change in subproblem specifications linked to activation splitting.
Furthermore, consistently with the incomplete verification results in the last section, Figure \ref{fig:lastlayer} shows that the Supergradient outperforms DSG+ on average, thus confirming the benefits of our Lagrangian Decomposition approach. 
Furthermore, Proximal provides an additional increase in average performance over DSG+ on the larger networks, which is visible especially over the properties that are easier to verify. 
The gap between competing bounding methods increases with the size of the employed network, both in terms of layer width and network depth.
We have seen that, by exploiting dual bounding algorithms, the performance of BaBSR can be significantly improved. We will now move on to studying the role of other branch and bound~components.

\subsection{Branching} \label{sec:complete-exp-branching}

The use of our Proximal solver improves the average verification performance within BaBSR (see $\S$\ref{sec:complete-exp-lastlayer}).
We now keep the bounding algorithm fixed to Proximal and study the effect of various branching strategies on verification time. 
As in BaBSR, we update intermediate bounds after each activation split with the best bounds between IBP and WK.
We consider the following branching schemes:
\begin{itemize}
	\item \textbf{SR} denotes the original branching scheme from BaBSR~\citep{Bunel2020}, relying on the activation scores found in equation \eqref{eq:babsr-scores}.
	\item \textbf{mSR}  denotes our modification of the SR branching scheme (min-based SR), in which $\mbf{s}_{\text{SR}}$ scores are replaced by $\mbf{s}_{\text{FSB}}$ as seen in equation \eqref{eq:fsb-scores}.
	\item \textbf{GNN} is the learning-based approach from \citet{Lu2020Neural}. It exploits a trained Graph Neural Network, which takes the topology of the network to be verified as input, to perform the branching decision, and falls back to SR whenever the decision from the GNN is deemed unsatisfactory. We re-train the network using Proximal as bounding algorithm, omitting the online fine-tuning due to its modest empirical impact~\citep{Lu2020Neural}. 
	\item \textbf{FSB} is our novel branching scheme (see $\S$\ref{sec:branching}), which exploits the mSR scores to select a subset of the branching choices, and then approximates strong branching using bounds from WK + CROWN.
\end{itemize}
\begin{figure}[t!]
	\sisetup{detect-weight=true,detect-inline-weight=math,detect-mode=true}
	\begin{subtable}{\textwidth}
		\centering
		\label{table:branching}
		
		\scriptsize
		\setlength{\tabcolsep}{4pt}
		\aboverulesep = 0.1mm  
		\belowrulesep = 0.2mm  
		\renewcommand{\bfseries}{\fontseries{b}\selectfont}
		\newrobustcmd{\B}{\bfseries}
		\newcommand{\boldentry}[2]{%
			\multicolumn{1}{S[table-format=#1,
				mode=text,
				text-rm=\fontseries{b}\selectfont
				]}{#2}}
		\begin{adjustbox}{max width=\textwidth, center}
			\begin{tabular}{
					l
					S[table-format=4.3]
					S[table-format=4.3]
					S[table-format=4.3]
					S[table-format=4.3]
					S[table-format=4.3]
					S[table-format=4.3]
					S[table-format=4.3]
					S[table-format=4.3]
					S[table-format=4.3]
				}
				& \multicolumn{3}{ c }{Base} & \multicolumn{3}{ c }{Wide} & \multicolumn{3}{ c }{Deep} \\
				\toprule
				
				\multicolumn{1}{ c }{Method} &
				\multicolumn{1}{ c }{time [s]} &
				\multicolumn{1}{ c }{subproblems} &
				\multicolumn{1}{ c }{$\%$Timeout} &
				\multicolumn{1}{ c }{time [s]} &
				\multicolumn{1}{ c }{subproblems} &
				\multicolumn{1}{ c }{$\%$Timeout} &
				\multicolumn{1}{ c }{time [s]} &
				\multicolumn{1}{ c }{subproblems} &
				\multicolumn{1}{ c }{$\%$Timeout} \\
				
				\cmidrule(lr){1-1} \cmidrule(lr){2-4} \cmidrule(lr){5-7} \cmidrule(lr){8-10}
				
				\multicolumn{1}{ c }{\textsc{SR}}
				& 822.61 &    146700.30 &    20.00 &   524.24 &     67495.74 &    11.00 &   234.23 &     15443.30 &     4.00 \\
				
				\multicolumn{1}{ c }{\textsc{mSR}}
				&  704.39 &    125989.32 &    17.00 &   299.59 &     38477.18 &     6.00 &   256.17 &     13573.06 &     4.00 \\
				
				\multicolumn{1}{ c }{\textsc{GNN}}
				& 633.71 &     92904.44 &    14.00 &   274.57 &     25337.36 &     6.00 &    88.43 &      3959.28 &     1.00 \\
				
				\multicolumn{1}{ c }{\textsc{FSB}}
				& \B546.22 &   \B  85433.94 &  \B  12.00 &   \B247.13 &    \B 20324.06 &  \B   6.00 &  \B  32.29 &     \B 1702.02 &   \B  0.00 \\
				
				\bottomrule
				
			\end{tabular}
		\end{adjustbox}
		\subcaption{\small Comparison of average runtime, average number of solved subproblems and the percentage of timed out properties. The best performing method is highlighted in bold.}
		\vspace{5pt}
	\end{subtable}
	\begin{subfigure}{\textwidth}
		\centering
		\begin{subfigure}{.32\textwidth}
			\centering
			\includegraphics[width=\textwidth]{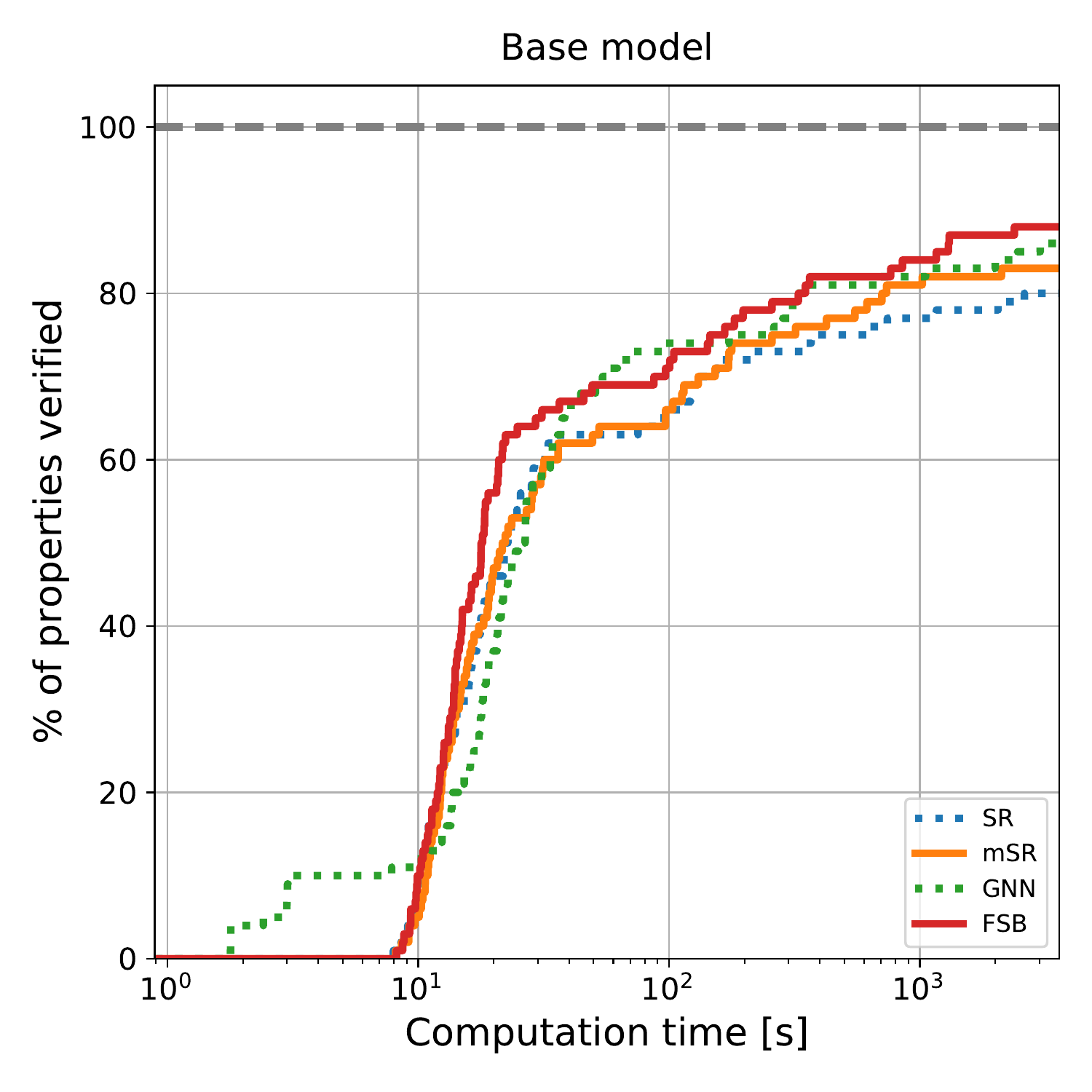}
		\end{subfigure}
		\begin{subfigure}{.32\textwidth}
			\centering
			\includegraphics[width=\textwidth]{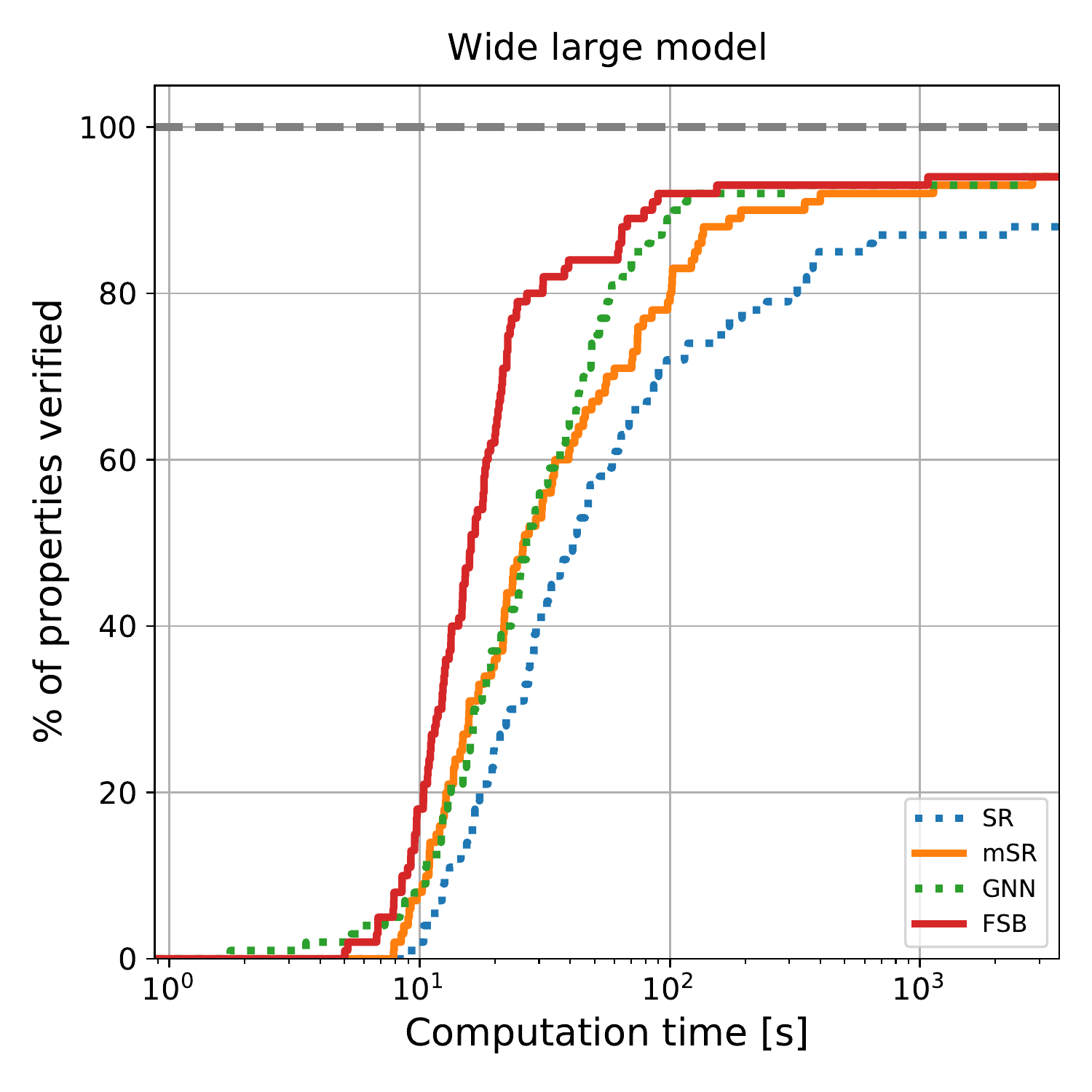}
		\end{subfigure}
		\begin{subfigure}{.32\textwidth}
			\centering
			\includegraphics[width=\textwidth]{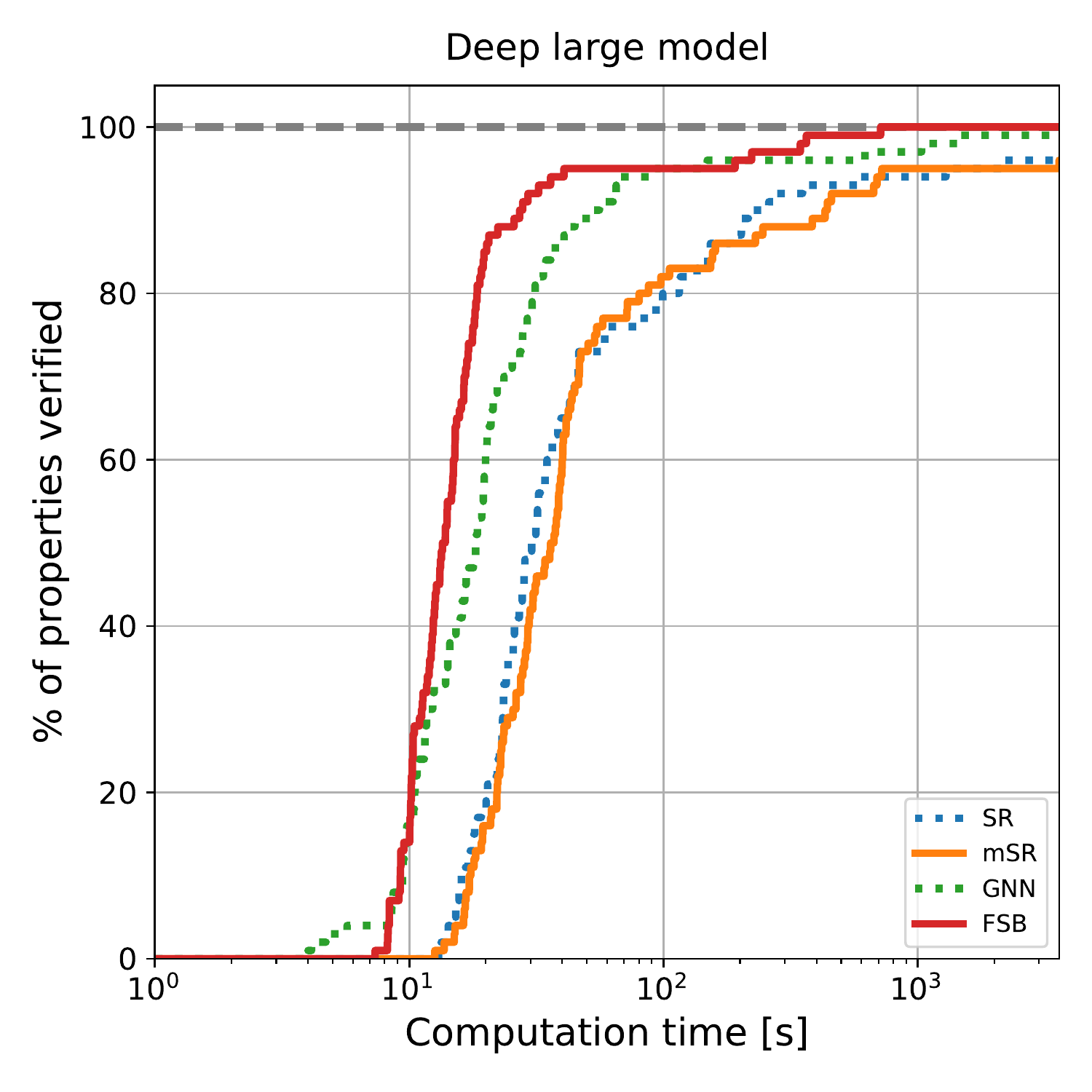}
		\end{subfigure}
		\vspace{-5pt}
		\subcaption{Cactus plots: percentage of solved properties as a function of runtime. Baselines are represented by dotted lines.}
	\end{subfigure}
	\caption{ Using Proximal as bounding algorithm, complete verification performance of different branching strategies, on the OVAL dataset. }
	\label{fig:branching}
\end{figure}
Similarly to dual bounding, branching computations were parallelized over batches of subproblems, and run on a single Nvidia Titan Xp GPU for all considered methods.  
In the context of our implementation, the time required for branching is negligible with respect to the cost of the bounding procedure, for all considered branching schemes. 
For this set of experiments, Proximal is additionally employed to compute upper bounds to a modified version of problem \eqref{eq:noncvx_pb}, where the minimization is replaced by a maximization, for each subproblem.
This maximizes the dual information available to the GNN, and provides an additional infeasibility check\footnote{A given activation split might empty the feasible region of problem \eqref{eq:planet-relax}.} for all methods. In fact, as infeasible subproblems will result in an unbounded dual, the subproblem can be discarded whenever the values of the upper and lower bounds cross. Empirically, this results in a modest decrease in the size of the enumeration tree and a minor increase in runtime (compare the results for SR in figure \ref{fig:branching}, with those for Proximal in figure~\ref{fig:lastlayer}).

Figure \ref{fig:branching} shows that our simple modification of SR successfully reduces the average number of subproblems to termination. This results in faster verification for both the Base and Wide network. For the Deep network, however, mSR tends to branch on earlier layers, thus involving a larger average number of intermediate bounding computations (see $\S$\ref{sec:badnb-ib}). 
This overhead is not matched by a significant reduction of the enumeration tree, slowing down overall verification.
The results for FSB demonstrate that coupling the mSR scores with fast dual algorithms yields a larger and more consistent reduction of subproblems with respect to SR.
As a consequence, FSB produces significant verification speed-ups to SR, which range from roughly $35\%$ on the Base network to an order of magnitude on the Deep network.
Furthermore, FSB is strongly competitive with GNN. On the considered networks, FSB outperforms the learned approach both in terms of average verification time and number of subproblems. 
Differently from GNN, strong verification performance is achieved without incurring large training-related offline costs.

\subsection{Intermediate Bounds} \label{sec:complete-exp-ib}

In this section, we consider the effect of various intermediate bounding strategies (see $\S$\ref{sec:intermediate}) on final verification performance. 
For this set of experiments, we use Proximal as bounding algorithm (for the subproblem bounds), and FSB for the branching strategy, computing intermediate bounds with the following algorithms:
\begin{itemize}
	\item \textbf{IBP + WK} denotes the layer-wise best bounds returned by IBP and WK (see $\S$\ref{sec:experiments-incomplete}), updated after each activation split (see $\S$\ref{sec:badnb-ib}).
	This is the intermediate bounding strategy employed in BaBSR~\citep{Bunel2020}.
	\item \textbf{WK + CROWN} denotes the layer-wise best bounds returned by WK and CROWN, updated after each activation split. 
	\item \textbf{WK + CROWN @root} restricts WK + CROWN to the root of the branch and bound tree, foregoing any possible tightening after activation splits.
\end{itemize}

\begin{figure}[t!]
	\sisetup{detect-weight=true,detect-inline-weight=math,detect-mode=true}
	\begin{subtable}{\textwidth}
		\centering
		
		\scriptsize
		\setlength{\tabcolsep}{4pt}
		\aboverulesep = 0.1mm  
		\belowrulesep = 0.2mm  
		\renewcommand{\bfseries}{\fontseries{b}\selectfont}
		\newrobustcmd{\B}{\bfseries}
		\newcommand{\boldentry}[2]{%
			\multicolumn{1}{S[table-format=#1,
				mode=text,
				text-rm=\fontseries{b}\selectfont
				]}{#2}}
		\begin{adjustbox}{max width=\textwidth, center}
			\begin{tabular}{
					l
					S[table-format=4.3]
					S[table-format=4.3]
					S[table-format=4.3]
					S[table-format=4.3]
					S[table-format=4.3]
					S[table-format=4.3]
					S[table-format=4.3]
					S[table-format=4.3]
					S[table-format=4.3]
				}
				& \multicolumn{3}{ c }{Base} & \multicolumn{3}{ c }{Wide} & \multicolumn{3}{ c }{Deep} \\
				\toprule
				
				\multicolumn{1}{ c }{Method} &
				\multicolumn{1}{ c }{time [s]} &
				\multicolumn{1}{ c }{subproblems} &
				\multicolumn{1}{ c }{$\%$Timeout} &
				\multicolumn{1}{ c }{time [s]} &
				\multicolumn{1}{ c }{subproblems} &
				\multicolumn{1}{ c }{$\%$Timeout} &
				\multicolumn{1}{ c }{time [s]} &
				\multicolumn{1}{ c }{subproblems} &
				\multicolumn{1}{ c }{$\%$Timeout} \\
				
				\cmidrule(lr){1-1} \cmidrule(lr){2-4} \cmidrule(lr){5-7} \cmidrule(lr){8-10}
				
				\multicolumn{1}{ c }{\textsc{IBP + WK}}
				& 526.77 &     94959.20 &    11.00 &   245.67 &     21429.28 &   \B  6.00 &    31.46 &      1704.10 &  \B   0.00 \\
				
				\multicolumn{1}{ c }{\textsc{WK + CROWN}}
				&  384.60 &     56823.50 &    \B 7.00 &   248.23 &     15750.48 &   \B  6.00 &    17.95 &       490.16 &  \B   0.00 \\
				
				
				\multicolumn{1}{ c }{\textsc{WK + CROWN @root}}
				& \B 329.46 &     97219.08 &   \B  7.00 &  \B 230.30 &     43952.94 &   \B  6.00 &   \B 11.41 &       533.90 &   \B  0.00 \\
				
				\bottomrule
				
			\end{tabular}
		\end{adjustbox}
		\subcaption{\label{table:ib}\small Comparison of average runtime, average number of solved subproblems and the percentage of timed out properties. The best performing method is highlighted in bold.}
		\vspace{5pt}
	\end{subtable}
	\begin{subfigure}{\textwidth}
		\centering
		\begin{subfigure}{.32\textwidth}
			\centering
			\includegraphics[width=\textwidth]{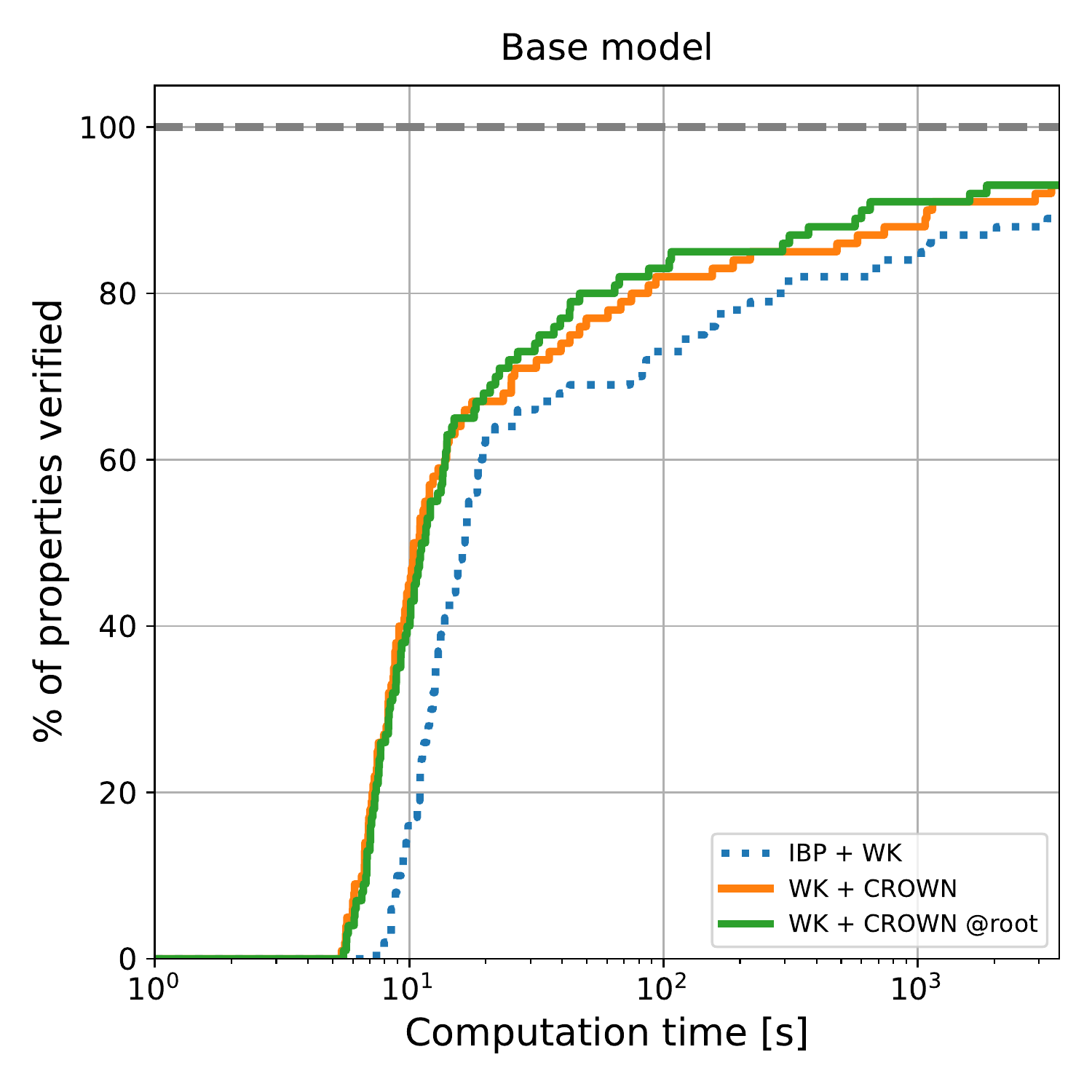}
		\end{subfigure}
		\begin{subfigure}{.32\textwidth}
			\centering
			\includegraphics[width=\textwidth]{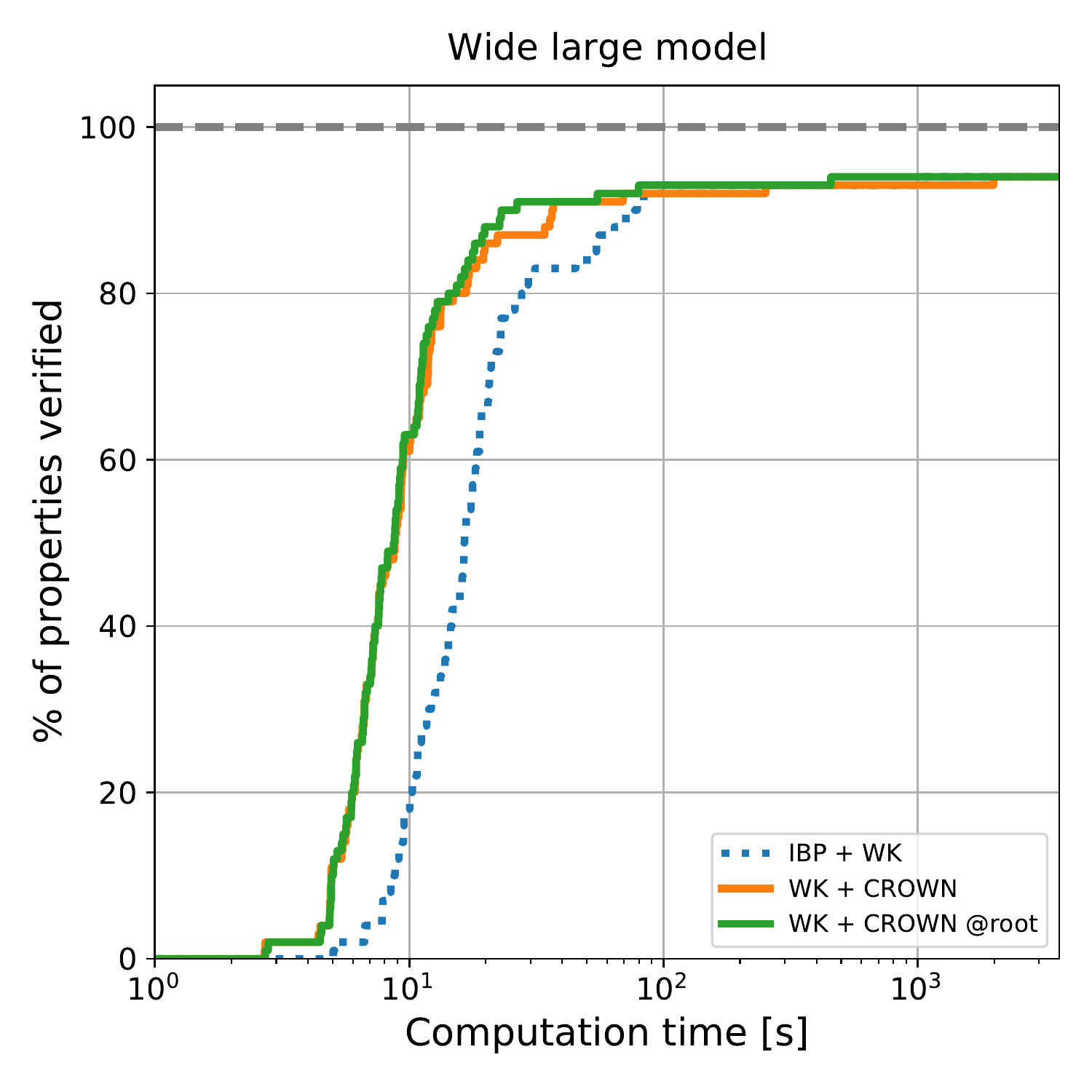}
		\end{subfigure}
		\begin{subfigure}{.32\textwidth}
			\centering
			\includegraphics[width=\textwidth]{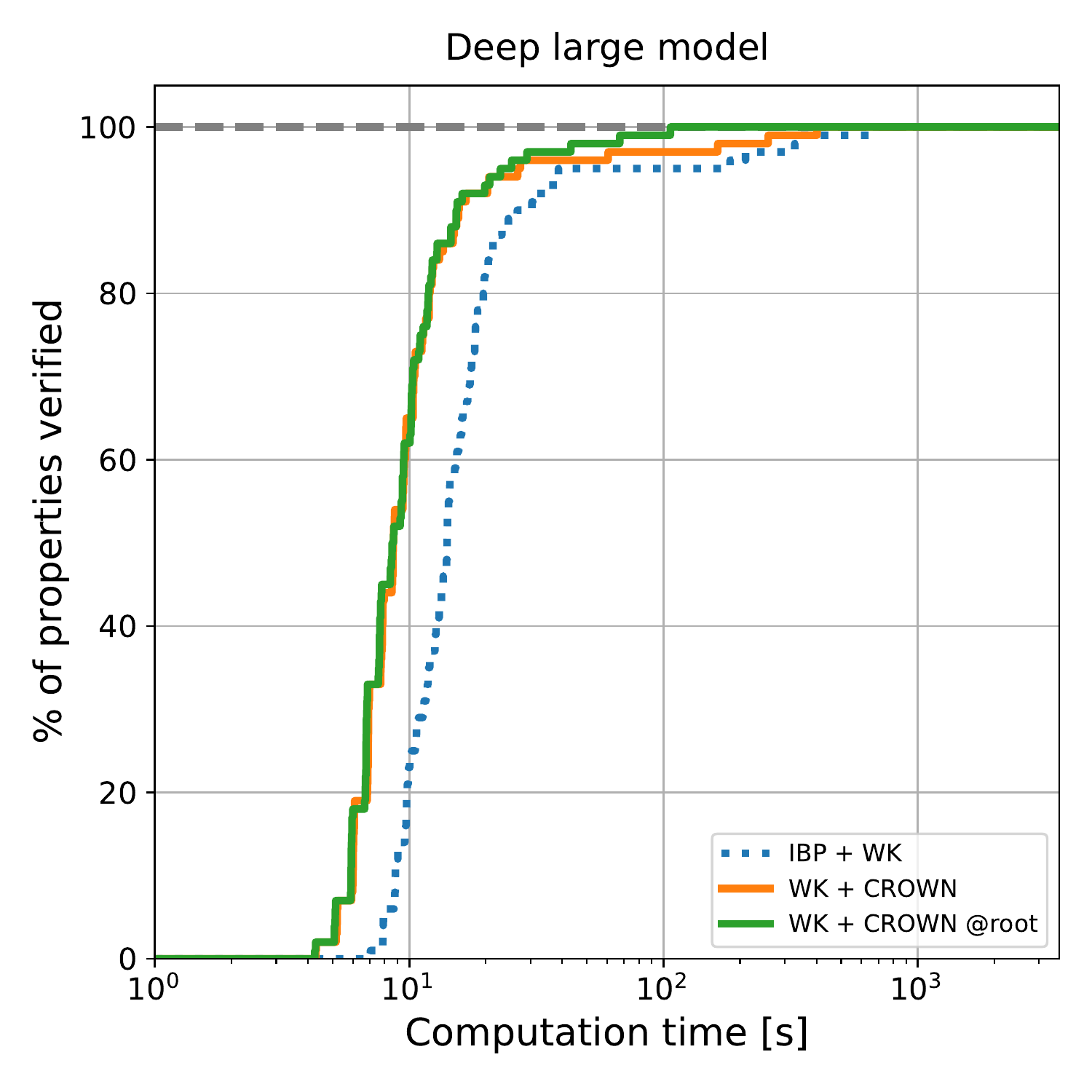}
		\end{subfigure}
		\vspace{-5pt}
		\subcaption{Cactus plots: percentage of solved properties as a function of runtime. Baselines are represented by dotted lines.}
	\end{subfigure}
	\caption{ Using Proximal for the last layer bounding, complete verification performance of different intermediate bounding schemes, on the OVAL dataset. }
	\label{fig:ib}
\end{figure}

As expected from the incomplete verification experiments in $\S$\ref{sec:experiments-incomplete}, replacing IBP with CROWN markedly tightens intermediate bounds.
This is testified by the decrease in the number of average subproblems to termination visible in figure \ref{fig:ib}. As a consequence, WK + CROWN reduces verification time for all the three considered networks.
Moreover, restricting WK + CROWN to the branch and bound root results in an increase in the average number of subproblems to termination.
This is due to both the reduced cost per branch and bound iteration, and a loss in bounding tightness. 
As evidenced by the decrease in verification time on all the three considered networks, WK + CROWN @root produces better speed-accuracy trade-offs than the other intermediate bounding strategies, and is particularly convenient for larger networks.

\subsection{Comparison of Complete Verifiers}  \label{sec:complete-exp-final}

In $\S$\ref{sec:complete-exp-lastlayer}, $\S$\ref{sec:complete-exp-branching}, and $\S$\ref{sec:complete-exp-ib}, we have studied the effect of isolated branch and bound components on the OVAL dataset.
We conclude our experimental evaluation by comparing our branch and bound framework with state-of-the-art complete verifiers on both the OVAL and COLT datasets.
For both benchmarks, we consider the following algorithms:
\begin{itemize}
	\item \textbf{MIP} solves problem \eqref{eq:noncvx_pb} as a Mixed-Integer linear Program (MIP) via Gurobi, exploiting the representation of ReLU activations used by \citep{Tjeng2019}. Gurobi was run on i7-6850K CPUs, using $6$ cores.
	In order to minimize pre-processing time, intermediate bounds are pre-computed with the layer-wise best bounds between IBP and WK, on an Nvidia Titan Xp GPU.
	\item \textbf{Gurobi BaBSR} is a multi-core adaptation of the BaBSR branch and bound framework~\citep{Bunel2020}. Gurobi is run on $6$ cores from i7-6850K CPUs (see ``Gurobi Planet" in $\S$\ref{sec:complete-exp-lastlayer}), branching and intermediate bounding were run on an Nvidia Titan Xp GPU.  
	\item \textbf{Proximal BaBSR} replaces the Gurobi-based Planet solver used in BaBSR with our Proximal bounding algorithm (see ``Proximal" in $\S$\ref{sec:complete-exp-lastlayer}), run on an Nvidia Titan Xp~GPU.
	\item \textbf{BaDNB} is our novel branch and bound framework (see $\S$\ref{sec:further-impr}). Compared to Proximal BaBSR, it employs FSB as branching strategy, dynamically determines the number of Proximal iterations, and computes intermediate bounds via WK + CROWN @root (see $\S$\ref{sec:complete-exp-ib}).
	BaDNB was run on an Nvidia Titan Xp GPU.
	\item \textbf{ERAN} is the complete verification toolbox by \citet{eran}, based on several works combining abstract interpretation, propagation-based methods, LP and MILP solvers~\citep{Singh2018a, Singh2019b, Singh2019, Singh2019a}. Results are taken from the recent VNN-COMP competition~\citep{vnn-comp}. 
	On the OVAL dataset, ERAN was executed on a 2.6 GHz Intel Xeon CPU E5-2690, using $14$ cores. On the COLT dataset, it was run on a $10$ Core Intel i9-7900X Skylake CPU.
\end{itemize}

\subsubsection{OVAL Dataset}

\begin{figure}[t!]
	\sisetup{detect-weight=true,detect-inline-weight=math,detect-mode=true}
	\begin{subtable}{\textwidth}
		\centering
		
		\scriptsize
		\setlength{\tabcolsep}{4pt}
		\aboverulesep = 0.1mm  
		\belowrulesep = 0.2mm  
		\renewcommand{\bfseries}{\fontseries{b}\selectfont}
		\newrobustcmd{\B}{\bfseries}
		\newcommand{\boldentry}[2]{%
			\multicolumn{1}{S[table-format=#1,
				mode=text,
				text-rm=\fontseries{b}\selectfont
				]}{#2}}
		\begin{adjustbox}{max width=\textwidth, center}
			\begin{tabular}{
					l
					S[table-format=4.3]
					S[table-format=4.3]
					S[table-format=4.3]
					S[table-format=4.3]
					S[table-format=4.3]
					S[table-format=4.3]
					S[table-format=4.3]
					S[table-format=4.3]
					S[table-format=4.3]
				}
				& \multicolumn{3}{ c }{Base} & \multicolumn{3}{ c }{Wide} & \multicolumn{3}{ c }{Deep} \\
				\toprule
				
				\multicolumn{1}{ c }{Method} &
				\multicolumn{1}{ c }{time [s]} &
				\multicolumn{1}{ c }{subproblems} &
				\multicolumn{1}{ c }{$\%$Timeout} &
				\multicolumn{1}{ c }{time [s]} &
				\multicolumn{1}{ c }{subproblems} &
				\multicolumn{1}{ c }{$\%$Timeout} &
				\multicolumn{1}{ c }{time [s]} &
				\multicolumn{1}{ c }{subproblems} &
				\multicolumn{1}{ c }{$\%$Timeout} \\
				
				\cmidrule(lr){1-1} \cmidrule(lr){2-4} \cmidrule(lr){5-7} \cmidrule(lr){8-10}
				
				\multicolumn{1}{ c }{\textsc{BaDNB}}
				& \B  309.30 &     38496.52 &     7.00 & \B  165.54 &     11258.56 &    \B 4.00 &   \B 10.50 &       368.16 &    \B 0.00 \\
				
				\multicolumn{1}{ c }{\textsc{Proximal BaBSR}}
				&  808.62 &    166478.44 &    20.00 &   498.69 &     80703.96 &    10.00 &   206.54 &     18139.62 &     3.00 \\
				
				\multicolumn{1}{ c }{\textsc{MIP}}
				&   2582.30 &     13929.44 &    57.00 &  1702.88 &      4170.95 &    38.00 &  1831.44 &      6268.08 &    25.00 \\
				
				\multicolumn{1}{ c }{\textsc{Gurobi BaBSR}}
				& 1352.15 &     7013.72 &    25.00 &  1236.54 &     2737.36 &    17.00 &   782.84 &       848.32 &     7.00 \\
				
				\multicolumn{1}{ c }{\textsc{ERAN}}
				& 805.89 &        {-}  &  \B   5.00 &   632.12 &        {-}  &     9.00 &   545.72 &       {-}  &    \B 0.00 \\
				
				\bottomrule
				
			\end{tabular}
		\end{adjustbox}
		\subcaption{\label{table:bab}\small Comparison of average runtime, average number of solved subproblems and the percentage of timed out properties. The best performing method is highlighted in bold.}
		\vspace{5pt}
	\end{subtable}
	\begin{subfigure}{\textwidth}
		\centering
		\begin{subfigure}{.32\textwidth}
			\centering
			\includegraphics[width=\textwidth]{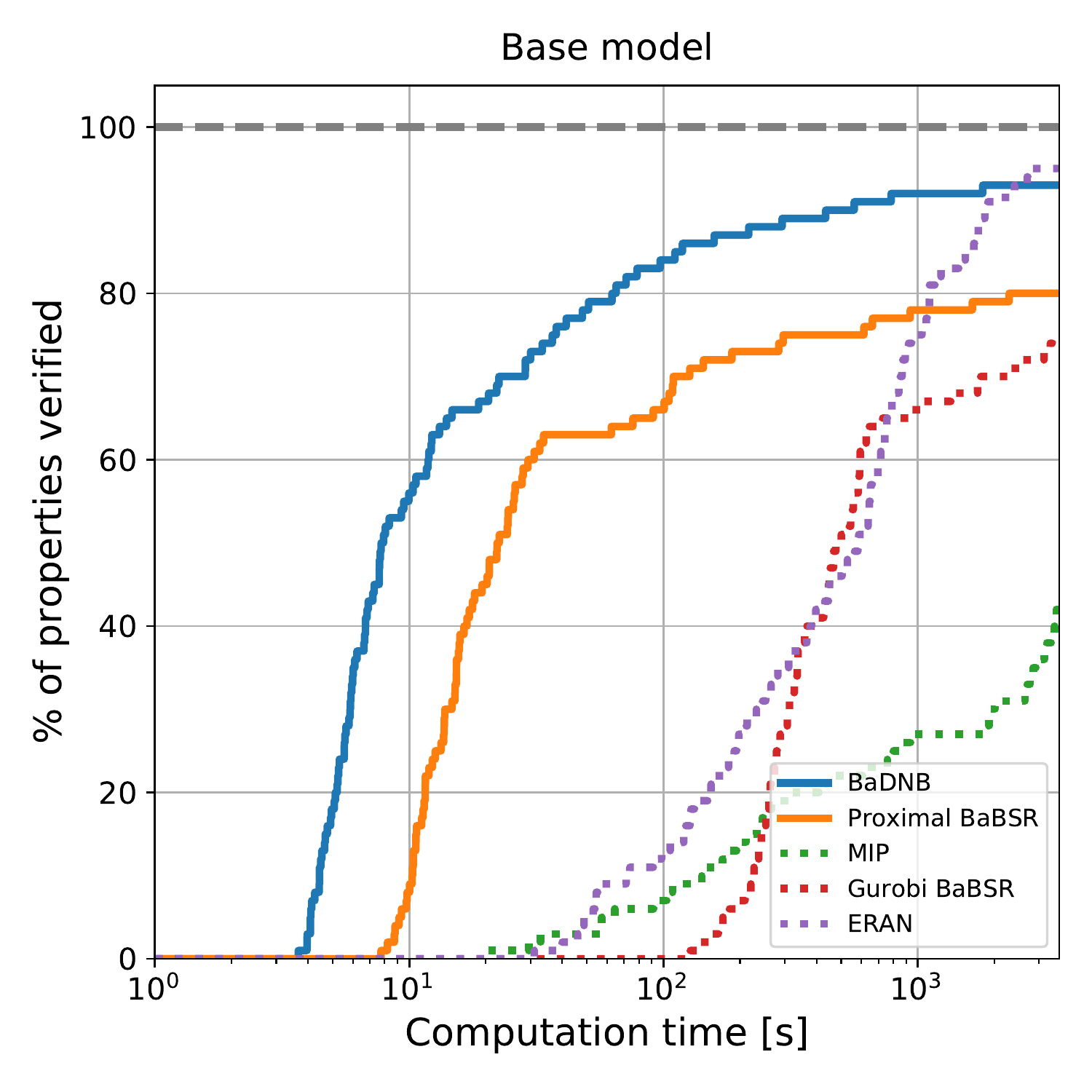}
		\end{subfigure}
		\begin{subfigure}{.32\textwidth}
			\centering
			\includegraphics[width=\textwidth]{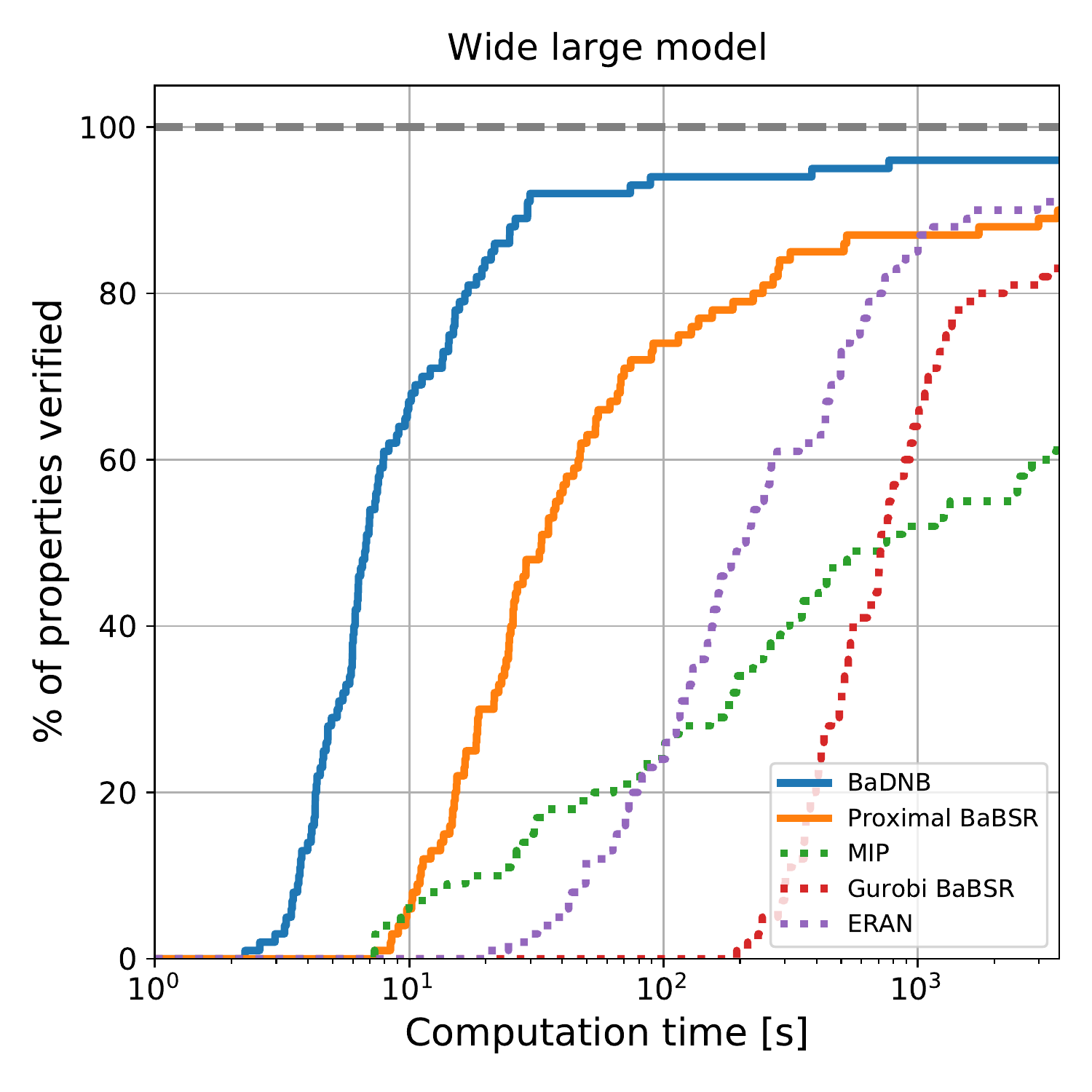}
		\end{subfigure}
		\begin{subfigure}{.32\textwidth}
			\centering
			\includegraphics[width=\textwidth]{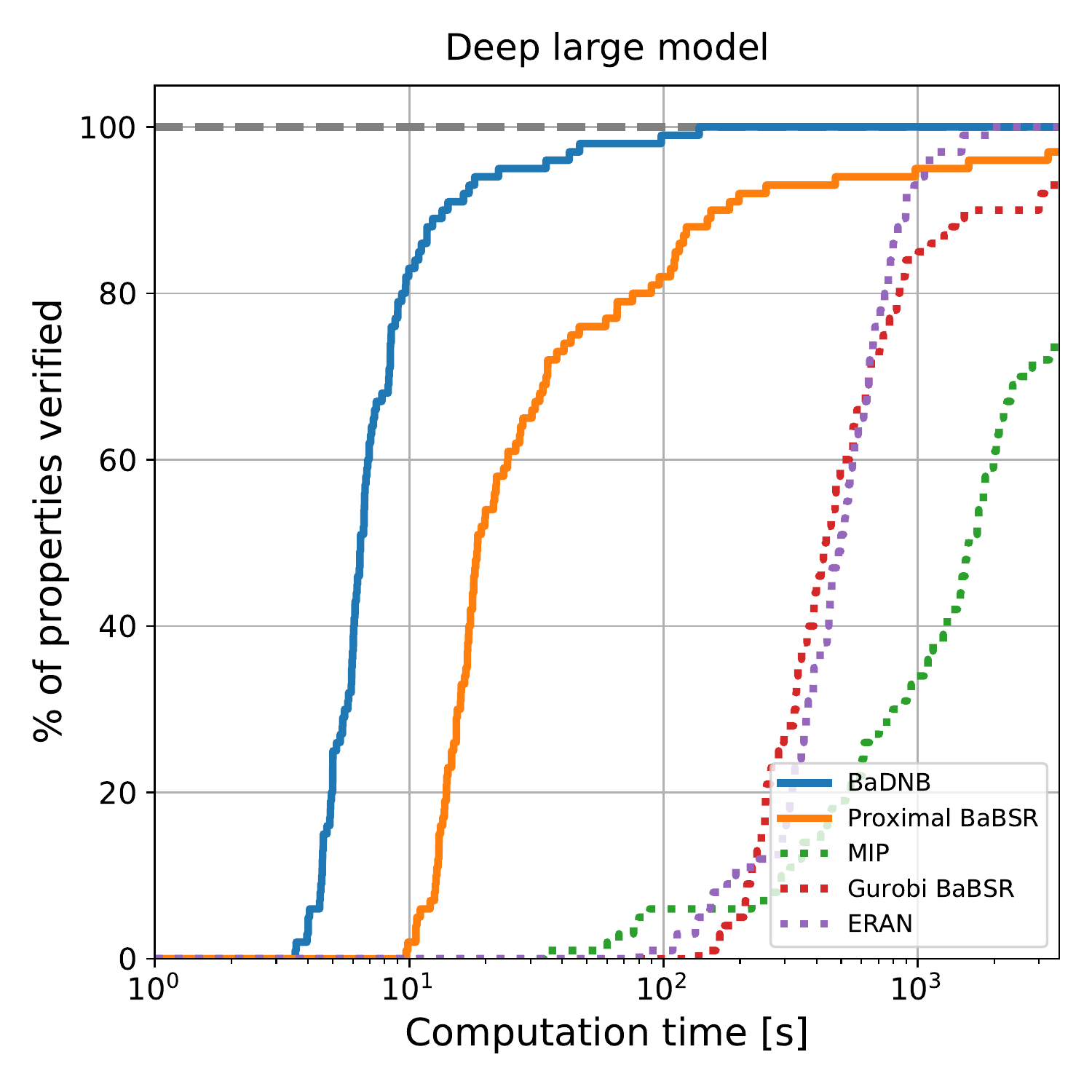}
		\end{subfigure}
		\vspace{-5pt}
		\subcaption{Cactus plots: percentage of solved properties as a function of runtime. Baselines are represented by dotted lines.}
	\end{subfigure}
	\caption{ Performance of different complete verification algorithms, on the OVAL dataset. }
	\label{fig:bab}
\end{figure}

Figure \ref{fig:bab} reports results for the OVAL dataset. MIP, which relies on a black-box MIP solver, is the slowest verification method in all three cases, highlighting the importance of specialized algorithms for neural network verification. Gurobi BaBSR improves upon MIP, thus demonstrating the benefits of a customized branch-and-bound framework. 
However, as seen in $\S$\ref{sec:complete-exp-lastlayer}, the use of Gurobi as bounding algorithm severely hinders scalability.
Proximal BaBSR showcases the benefits of specialized solvers for the Planet relaxation: it significantly outperforms Gurobi BaBSR and is faster than ERAN on the larger networks.  
Furthermore, BaDNB manages to further cut verification times, yielding speed-ups to Proximal BaBSR that reach an order of magnitude on the Deep network.
This testifies the advantages of Filtered Smart Branching and of the other design choices presented in $\S$\ref{sec:badnb}.
In addition to the results presented in $\S$\ref{sec:complete-exp-branching}, and $\S$\ref{sec:complete-exp-ib}, a comparison of BaDNB in table \ref{table:bab} with WK + CROWN @root in table \ref{table:ib} shows that, by automatically adjusting the number of dual iterations ($\S$\ref{sec:badnb-bt}), we can obtain a further $40\%$ reduction in average verification time on the Wide~network.

\subsubsection{COLT Dataset}

\begin{figure}[t!]
	\sisetup{detect-weight=true,detect-inline-weight=math,detect-mode=true}
	\begin{subtable}{\textwidth}
		\centering
		\scriptsize
		\setlength{\tabcolsep}{4pt}
		\aboverulesep = 0.1mm  
		\belowrulesep = 0.2mm  
		\renewcommand{\bfseries}{\fontseries{b}\selectfont}
		\newrobustcmd{\B}{\bfseries}
		\newcommand{\boldentry}[2]{%
			\multicolumn{1}{S[table-format=#1,
				mode=text,
				text-rm=\fontseries{b}\selectfont
				]}{#2}}
		\begin{adjustbox}{width=0.7\textwidth, center}
			\begin{tabular}{
					l
					S[table-format=4.3]
					S[table-format=4.3]
					S[table-format=4.3]
					S[table-format=4.3]
					S[table-format=4.3]
					S[table-format=4.3]
				}
				& \multicolumn{3}{ c }{2/255} & \multicolumn{3}{ c }{8/255} \\
				\toprule
				
				\multicolumn{1}{ c }{Method} &
				\multicolumn{1}{ c }{time [s]} &
				\multicolumn{1}{ c }{subproblems} &
				\multicolumn{1}{ c }{$\%$Timeout} &
				\multicolumn{1}{ c }{time [s]} &
				\multicolumn{1}{ c }{subproblems} &
				\multicolumn{1}{ c }{$\%$Timeout} \\
				
				\cmidrule(lr){1-1} \cmidrule(lr){2-4} \cmidrule(lr){5-7} 
				
				\multicolumn{1}{ c }{\textsc{BaDNB}}
				&  \B 211.06 &      2064.02 &    \B  4.88 &   \B 667.29 &     17254.61 &   \B 14.29 \\
				
				\multicolumn{1}{ c }{\textsc{Proximal BaBSR}}
				&   842.00 &     17979.17 &    21.95 &  1249.02 &     55653.04 &    33.93 \\
				
				\multicolumn{1}{ c }{\textsc{MIP}}
				&   1353.07 &       103.11 &    25.61 &  1611.46 &       461.70 &    37.50 \\
				
				\multicolumn{1}{ c }{\textsc{Gurobi BaBSR}}
				& 1313.88 &      32.44 &    25.61 &  1461.48 &      310.18 &    35.71 \\
				
				\multicolumn{1}{ c }{\textsc{ERAN$^*$}} 
				& 1136.76 &        {-}&    28.05 &  1541.98 &        {-} &    41.07 \\
				
				\multicolumn{1}{ c }{\textsc{nnenum$^*$}}
				& 1081.02 &        {-} &    29.27 &  1431.46 &        {-} &    39.29 \\
				
				\multicolumn{1}{ c }{\textsc{VeriNet$^*$}}
				& 845.57 &        {-} &    23.17 &  1225.24 &        {-} &    33.93 \\
				
				\bottomrule
				
			\end{tabular}
		\end{adjustbox} 
		\subcaption{\label{table:bab-eth}\small Comparison of average runtime, average number of solved subproblems and the percentage of timed out properties. The best performing method is highlighted in bold. $^*$indicates that the method was run with a $5$-minute time limit, within \citet{vnn-comp}.}
		\vspace{5pt}
	\end{subtable}
	\begin{subfigure}{\textwidth}
		\centering
		\begin{subfigure}{.32\textwidth}
			\centering
			\includegraphics[width=\textwidth]{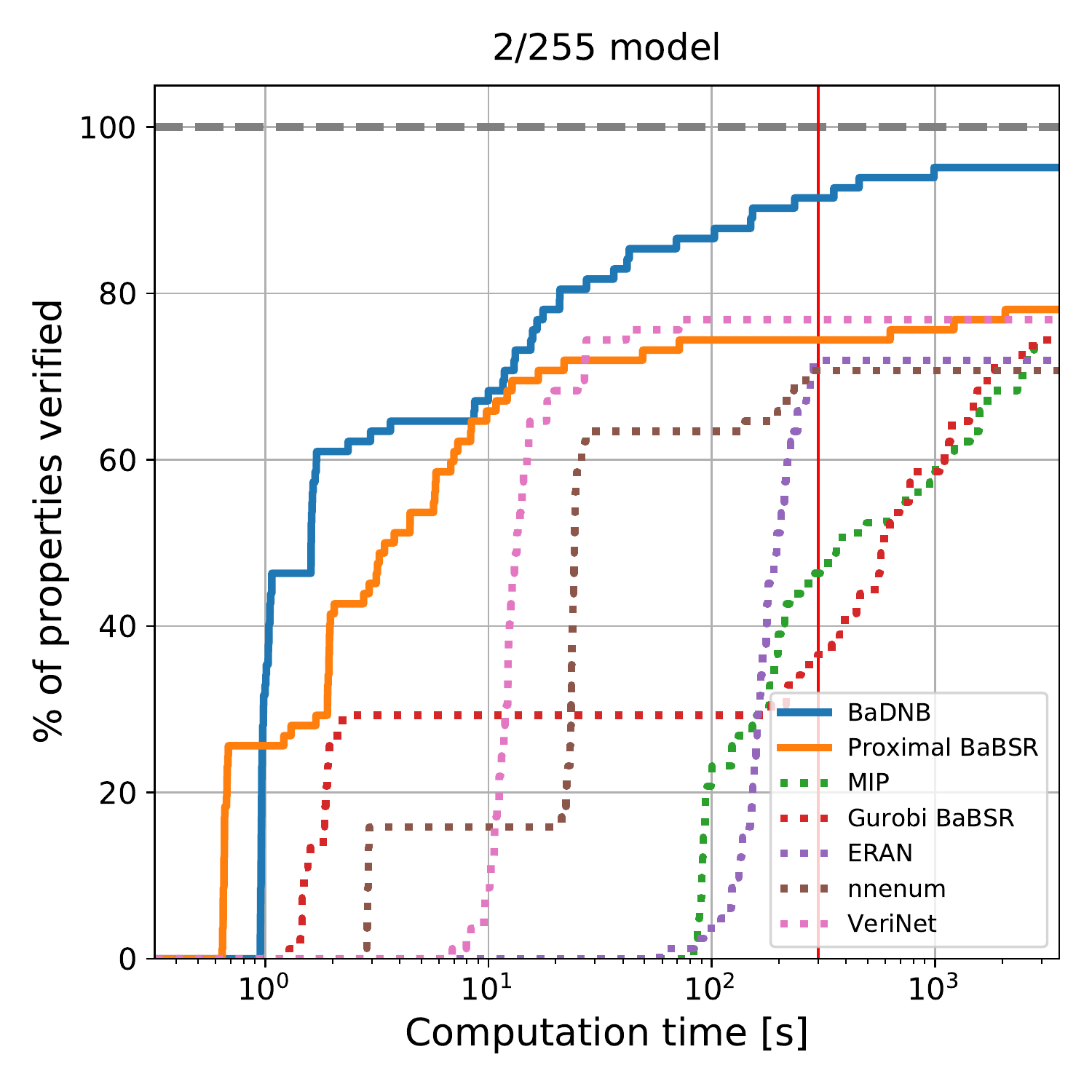}
		\end{subfigure}
		\begin{subfigure}{.32\textwidth}
			\centering
			\includegraphics[width=\textwidth]{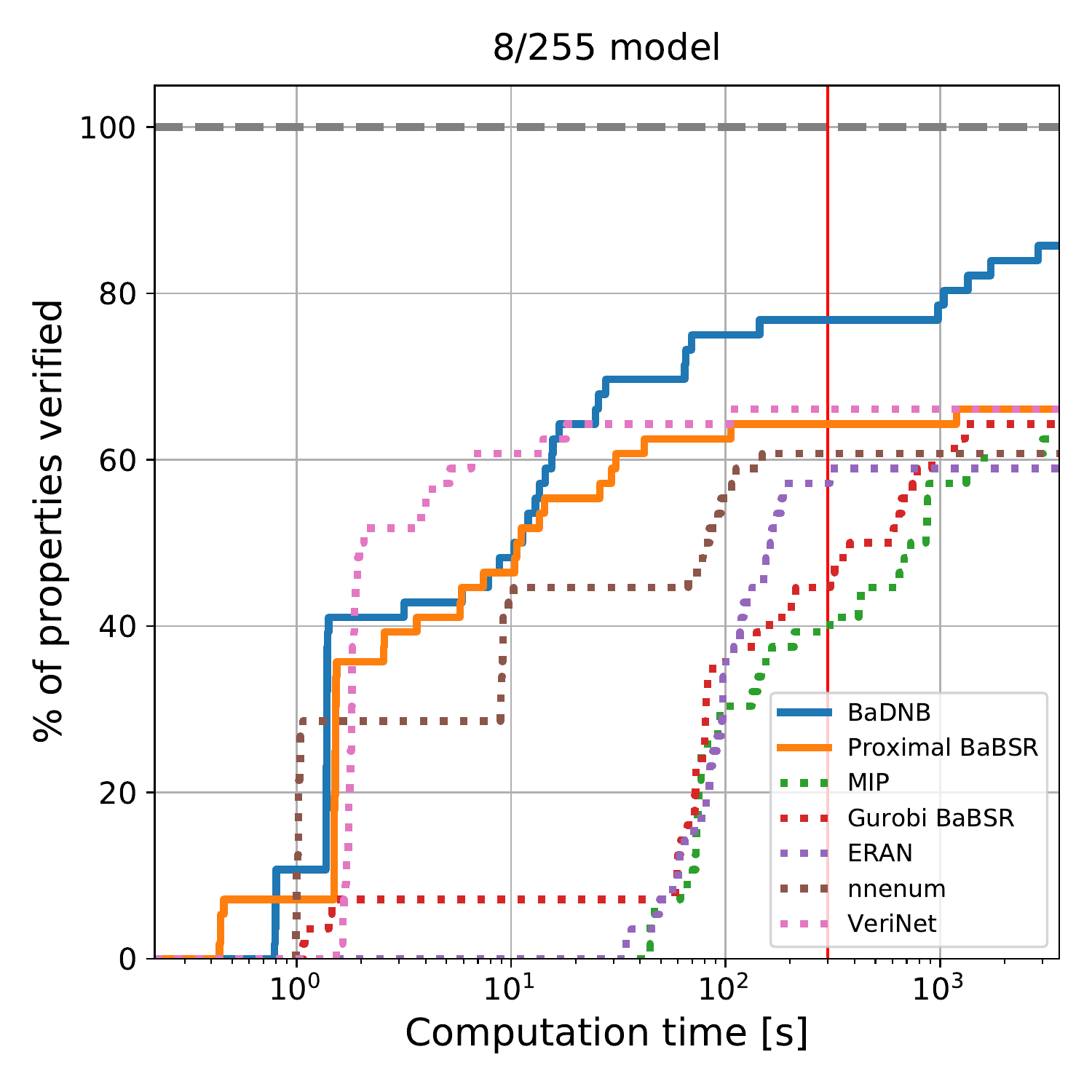}
		\end{subfigure}
		\vspace{-5pt}
		\subcaption{Cactus plots: percentage of solved properties as a function of runtime. Baselines are represented by dotted lines. The red vertical line marks the time limit for the ERAN, nnenum and VeriNet experiments from VNN-COMP~\citep{vnn-comp}.}
	\end{subfigure}
	\caption{ Performance of different complete verifiers on adversarial robustness verification of larger, COLT-trained convolutional, networks~\citep{Balunovic2020}. }
	\label{fig:bab-eth}
\end{figure}

We now report results for the COLT dataset, which offer insight into the scalability of the various complete verifiers on larger networks. 
Owing to the wider participation to this benchmark within VNN-COMP~\citep{vnn-comp}, we additionally report results for the two best-performing algorithms within the competition:
\begin{itemize}
	\item \textbf{nnenum} by \citet{Bak2020}, which pairs propagation-based methods and LP solvers with path enumeration techniques~\citep{Tran2019}.
	nenum was executed on an Amazon EC2 \texttt{m4.10xlarge} cloud instance, with a $40$-core 2.4 GHz Intel Xeon E5-2676 v3 CPU.
	\item \textbf{VeriNet} from \citet{Henriksen20}, which presents modifications to the branch and bound algorithm from Neurify~\citep{Wang2018b}. In particular, both the subproblem upper bounding strategy and the activation-based branching schemes are improved upon. VeriNet was run on a Ryzen 3700X 3.6 GHz 8-core CPU.
\end{itemize}
Due to the specifications of VNN-COMP the experiments for ERAN, nnenum and VeriNet were run with a $5$-minute time limit on this dataset. In line with the time-limit for the other methods, Table \ref{table:bab-eth} reports a runtime of 1 hour for all timed-out properties.

The performance of MIP in figure \ref{fig:bab-eth} demonstrates that black-box MIP solvers are unsuitable for larger networks. In fact MIP can rarely verify any property in less than hundreds of seconds. Furthermore, Gurobi's scaling problems are further evidenced by the results for Gurobi BaBSR, which hardly improves upon MIP's performance.
As seen on the OVAL dataset (figure \ref{fig:bab-eth}), the use of our dual bounding algorithm greatly improves upon the original BaBSR implementation~\citep{Bunel2020}: Proximal BaBSR consistently outperforms ERAN and nnenum, and it is competitive with VeriNet.
However, Proximal BaBSR still times out on a relatively large share of the considered properties, underscoring the limitations of BaBSR's design.
The superior performance of BaDNB demonstrates that a scrupulous deployment of fast dual bounds throughout the branch and bound procedure is key to effective neural network verification.

\section{Discussion}

We have presented BaDNB, a novel branch and bound framework for neural network verification, and empirically demonstrated its advantages to state-of-the-art verification systems.

As part of BaDNB, we proposed a novel dual approach to neural network bounding, based on Lagrangian Decomposition. 
Our bounding algorithms provide significant benefits compared to off-the-shelf solvers and improve on both looser relaxations and on a previous method based on Lagrangian relaxation. 
While we have focused on the convex hull of element-wise activation functions, our dual approach is far more general. In fact, we believe that Lagrangian Decomposition has the potential to scale up tighter relaxations from the Mixed Integer Linear Programming literature~\citep{Sherali1994}. 
We have furthermore shown that inexpensive dual algorithms can significantly speed up verification if employed to select branching decisions and to tighten intermediate bounds. We decided to keep costs low for these branch and bound components in order to obtain a well-rounded complete verifier. 
However, we are convinced that further verification improvements can be obtained by selectively employing solvers for Lagrangian Decomposition in this context.




\clearpage

\renewcommand{\theHsection}{A\arabic{section}}
\appendix

\section{Proof of theorem \ref{theorem}}
\label{sec:dual_comp_proof}

We will show the relation between the dual problem by \citet{Dvijotham2018} and the one proposed in this paper. 
We will show that, in the context of ReLU activation functions, from any dual point of their dual providing a given bound, our dual provides a bound at least as tight. 
Moreover, our dual coincides with the one by \citet{Dvijotham2018} if the latter is modified to include additional equality constraints.

\vspace{10pt}
Let us write $\sigma(\cdot)=\max(0, \cdot)$.
We start from problem\footnote{Our activation function $\sigma$ is denoted $h$ in their paper.
	The equivalent of $\zb$ in their paper is $\xb$ in ours, while the equivalent
	of their $\xb$ is $\xbhat$. Also note that their paper use the computation of
	upper bounds as examples while ours use the computation of lower bounds.} \eqref{eq:dj_prob-main} by \citet{Dvijotham2018}.
Decomposing it, in the same way that \citet{Dvijotham2018} do it in order to
obtain their equation (7), and with the convention that $\mub_n = -1$) we obtain:
\begin{equation}
\begin{split}
& b_n - \sum_{k=1}^{n-1} \mub_k^T \mbf{b}_k
+ \sum_{k=1}^{n-1}\min_{\xbhat_k \in [\hat{\lb}_k, \hat{\ub}_k]}\left( \mub_k^T \xbhat_k - \lambdab_k^T\sigma(\xbhat_k) \right)\\
&\qquad + \sum_{k=1}^{n-1}\min_{\xb_k \in [\sig{\hat{\lb}_k}, \sig{\hat{\ub}_k}]}\left( -\mub_{k+1}^TW_{k+1} \xb_k + \lambdab_k^T\xb_k \right)
+ \min_{\xb_0 \in \mathcal{C}} -\mub_1^T W_1 \xb_0.
\end{split}
\label{eq:dj_dual_decomp}
\end{equation}

With the convention that $\lambdabhat_n = -1$, our dual \eqref{eq:nonprox_problem} can be decomposed as:
\begin{equation}
\begin{split}
q\left( \lambdabhat \right) = \sum_{k=1}^{n-1}&\left(
\min_{\begin{array}{c}\xb_{k}, \xbhat_{B, k}, \xbhat_{A, k+1} \\ \in \mathcal{P}_k\left( \cdot, \cdot, \cdot \right)\end{array}}
\lambdabhat_k^T \xbhat_{B,k}\ -\ \lambdabhat_{k+1}^T \xbhat_{A, k+1}
\right)
+ \min_{\begin{array}{c}\xb_0, \xbhat_{A,1}\\\in \mathcal{P}_0\left( \cdot, \cdot \right)\end{array}} -\lambdabhat_1^T \xbhat_{A, 1},
\end{split}
\label{eq:dual_decomp_vs_dj}
\end{equation}
We will show that when we chose the dual variables such that
\begin{equation}
\lambdabhat_k = \mub_k,
\label{eq:vs_dj_replacement}
\end{equation}
we obtain a tighter bound than the ones given by \eqref{eq:dj_dual_decomp}.

We will start by showing that the term being optimized over $\mathcal{P}_0$ is
equivalent to some of the terms in \eqref{eq:dj_dual_decomp}:
\begin{equation}
\min_{\begin{array}{c}\xb_0, \xbhat_{A,1}\\\in \mathcal{P}_0\left( \cdot, \cdot \right)\end{array}}
-\lambdabhat_1^T \xbhat_{A, 1}
= \min_{\begin{array}{c}\xb_0, \xbhat_{A,1}\\\in \mathcal{P}_0\left( \cdot, \cdot \right)\end{array}}
- \mub_1^T \xbhat_{A, 1}
= -\mub_1^T \mbf{b}_1 + \min_{\xb_0 \in \mathcal{C}} - \mub_1^T W_1 \xb_0
\label{eq:pb0_lb}
\end{equation}
The first equality is given by the replacement
formula~\eqref{eq:vs_dj_replacement}, while the second one is given by
performing the replacement of $\xbhat_{A, 1}$ with his expression in $\mathcal{P}_0$.

We will now obtain a lower bound of the term being optimized over
$\mathcal{P}_{k}$. Let's start by plugging in the values using the
formula~\eqref{eq:vs_dj_replacement} and replace the value of $\xbhat_{A, k+1}$
using the equality constraint.
\begin{equation}
\begin{aligned}
&\min_{\begin{array}{c}\xb_k, \xbhat_{B, k}, \xbhat_{A, k+1} \\ \in \mathcal{P}_k\left( \cdot, \cdot, \cdot \right)\end{array}}
\left( \lambdabhat_k^T \xbhat_{B,k}\ -\ \lambdabhat_{k+1}^T \xbhat_{A, k+1} \right) = \\
& \qquad= -\mub_{k+1}^T \mbf{b}_{k+1} 
+  \min_{\begin{array}{l}
	\qquad \xbhat_{B, k} \in [\hat{\lb}_k, \hat{\ub}_k]\\
	(\xb_k, \xbhat_{B, k}) \in \text{Conv}(\sigma, \hat{\lb}_k, \hat{\ub}_k)
	\end{array}}\quad
\left( \mub_k^T \xbhat_{B,k}\ -\ \mub_{k+1}^T W_{k+1} \xb_{k} \right) 
\end{aligned}
\label{eq:pbk_lb}
\end{equation}
Focusing on the second term that contains the minimization over the convex hull,
we will obtain a lower bound.
It is important, at this stage, to recall that, as seen in section \ref{sec:innerminpk}, the minimum of the second term can either be one of the three vertices of the triangle in Figure~\ref{fig:relucvx} (ambiguous ReLU), the $\xbhat_{B, k}=\xb_{k}$ line (passing ReLU), or the $(\xbhat_{B, k}=\mathbf{0},\xb_{k}=\mathbf{0})$ triangle vertex (blocking ReLU). We will write $(\xbhat_{B, k}, \xb_{k}) \in \texttt{ReLU\_sol}(\xbhat_{B, k}, \xb_k,\hat{\lb}_k, \hat{\ub}_k)$.

We can add a term $\lambdab_k^T(\sig{\xbhat_k} -
\sig{\xbhat_k}) = 0$ and obtain:
\begin{equation}
\begin{split}
& \min_{\begin{array}{c}
	\xbhat_{B, k} \in [\hat{\lb}_k, \hat{\ub}_k]\\
	(\xb_k, \xbhat_{B, k}) \in \text{Conv}(\sigma, \hat{\lb}_k, \hat{\ub}_k)
	\end{array}}
\left( \mub_k^T \xbhat_{B,k}\ -\ \mub_{k+1}^T W_{k+1} \xb_{k} \right)\\[10pt]
= & \min_{\begin{array}{c}
	\xbhat_{B, k} \in [\hat{\lb}_k, \hat{\ub}_k]\\
	\texttt{ReLU\_sol}(\xbhat_{B, k}, \xb_k,\hat{\lb}_k, \hat{\ub}_k)
	\end{array}}
\left( \mub_k^T \xbhat_{B,k}\ -\ \mub_{k+1}^T W_{k+1} \xb_{k} \right)\\[10pt]
= & \min_{\begin{array}{c}
	\xbhat_{B, k} \in [\hat{\lb}_k, \hat{\ub}_k]\\
	\texttt{ReLU\_sol}(\xbhat_{B, k}, \xb_k,\hat{\lb}_k, \hat{\ub}_k)
	\end{array}}
\left( \mub_k^T \xbhat_{B,k}\ -\ \mub_{k+1}^T W_{k+1} \xb_{k} + \lambdab_k^T(\sig{\xbhat_{B,k}} - \sig{\xbhat_{B,k}})\right)\\[10pt]
\geq & \min_{\begin{array}{c}
	\xbhat_{B, k} \in [\hat{\lb}_k, \hat{\ub}_k]\\
	\texttt{ReLU\_sol}(\xbhat_{B, k}, \xb_k,\hat{\lb}_k, \hat{\ub}_k)
	\end{array}} \left( \mub_k^T \xbhat_{B, k} - \lambdab_k^T\sig{\xbhat_{B,k}}\right) \\
&\quad +\min_{\begin{array}{c}
	\xbhat_{B, k} \in [\hat{\lb}_k, \hat{\ub}_k]\\
	\texttt{ReLU\_sol}(\xbhat_{B, k}, \xb_k,\hat{\lb}_k, \hat{\ub}_k)
	\end{array}} \left(- \mub_{k+1}^TW_{k+1}\xb_k  + \lambdab_k^T \sig{\xbhat_{B,k}}\right)\\[10pt]
\geq & \min_{\xbhat_{B,k} \in [\hat{\lb}_k, \hat{\ub}_k]} \left( \mub_k^T \xbhat_{B, k} - \lambdab_k^T\sig{\xbhat_{B,k}}\right)
+ \min_{\xb_k \in [\sig{\hat{\lb}_k}, \sig{\hat{\ub}_k}]} \left( -\mub_{k+1}^TW_{k+1}\xb_k  + \lambdab_k^T \xb_k\right).
\end{split}
\label{eq:lambpos_lb}
\end{equation}
Equality between the second line and the third comes from the fact that we
are adding a term equal to zero. The inequality between the third line and the
fourth is due to the fact that the sum of minimum is going to be lower than the
minimum of the sum. 
For what concerns obtaining the final line, the first term comes from projecting $\xb_{k}$ out of the feasible domain and taking the convex hull of the resulting domain. We need to look more carefully at the second term. Plugging in the ReLU formula:
\begin{equation*}
\begin{split}
&\min_{\begin{array}{c}
	\xbhat_{B, k} \in [\hat{\lb}_k, \hat{\ub}_k]\\
	\texttt{ReLU\_sol}(\xbhat_{B, k}, \xb_k,\hat{\lb}_k, \hat{\ub}_k)
	\end{array}} \left(- \mub_{k+1}^TW_{k+1}\xb_k  + \lambdab_k^T \max\{0, \xbhat_{B,k}\}\right) \\
= &\min_{\begin{array}{c}
	\xbhat_{B, k} \in [\sigma(\hat{\lb}_k), \sigma(\hat{\ub}_k)]\\
	\texttt{ReLU\_sol}(\xbhat_{B, k}, \xb_k,\hat{\lb}_k, \hat{\ub}_k)
	\end{array}} \left(- \mub_{k+1}^TW_{k+1}\xb_k  + \lambdab_k^T \xbhat_{B,k}\right) \\
\geq &\min_{\xb_{k} \in [\sigma(\hat{\lb}_k), \sigma(\hat{\ub}_k)]} \left(- \mub_{k+1}^TW_{k+1}\xb_k  + \lambdab_k^T \xb_{k}\right),
\end{split}
\end{equation*}
as (keeping in mind the shape of $\texttt{ReLU\_sol}$ and for the purposes of this specific problem) excluding the negative part of the $\xbhat_{B,k}$ domain does not alter the minimal value. 
The final line then follows by observing that forcing $\xbhat_{B, k}=\xb$ is a convex relaxation of the positive part of $\texttt{ReLU\_sol}$.
Summing up the lower bounds for all the terms in \eqref{eq:dual_decomp_vs_dj},
as given by equations \eqref{eq:pb0_lb} and \eqref{eq:lambpos_lb},
we obtain the formulation of Problem~\eqref{eq:dj_dual_decomp}.
We conclude that the bound obtained by the original dual by \citet{Dvijotham2018} is necessarily
no larger than the bound derived using our dual. Given that we are computing lower
bounds, this means that their bound is~looser.

\vspace{10pt}
Finally, we now prove that our dual \eqref{eq:nonprox_problem} coincides with the following modified version of problem \eqref{eq:dj_prob-main}:
\begin{equation*}
\begin{aligned}
& \max_{\mub,\lambdab}\enskip  d_{O}(\mub,\lambdab) \\
\text{s.t. }\quad& \xb_0 \in \mathcal{C},\\
& (\xb_k, \xbhat_k) \in [\sigma(\hat{\lb}_k),\sigma(\hat{\ub}_k)]  \times  [\hat{\lb}_k, \hat{\ub}_k] \qquad && k \in \left\llbracket1, n-1\right\rrbracket, \\
&  \lambdab'_{n-1} = -W_n^T \mbf{1}, \quad \lambdab'_{k-1} = W_k^T \mub_k  && k \in \left\llbracket2,n-1\right\rrbracket,
\end{aligned}
\end{equation*}
whose objective, keeping the convention that $\mub_n = -1$, can be rewritten as:
\begin{equation}
\begin{aligned}
d_{O}'(\mub) &=  \min_{\xbhat} \
\sum_{k=1}^{n-1} \mub_k^T \left( \xbhat_k - \mbf{b}_k \right) 
-\sum_{k=1}^{n-1} \mub_k^T W_k   \sigma(\xbhat_k) 
-\mub_1^T W_1 \xb_0 + \mbf{b}_n   \\
\text{s.t. }\quad& \xb_0 \in \mathcal{C},\qquad \xbhat_k \in [\hat{\lb}_k, \hat{\ub}_k] \quad k \in \left\llbracket1, n-1\right\rrbracket.
\end{aligned}
\label{eq:dual-dj-modified}
\end{equation}
Exploiting equations \eqref{eq:pb0_lb} and \eqref{eq:pbk_lb}, and re-using the notation for $\texttt{ReLU\_sol}(\xbhat_{B, k}, \xb_k,\hat{\lb}_k, \hat{\ub}_k)$ and $\mub_n = -1$, we obtain the following reformulation of $q(\mub)$: 
\begin{equation*}
\begin{aligned}
q(\mub) &=  \min_{\xb, \xbhat} \
- \sum_{k=1}^{n} \mub_k^T \mbf{b}_k 
+ \sum_{k=1}^{n-1} \left( \mub_k^T \xbhat_{B,k}\ -\ \mub_{k+1}^T W_{k+1} \xb_{k} \right) 
-\mub_1^T W_1 \xb_0  \\
\text{s.t. }\quad& \xb_0 \in \mathcal{C}, \\ 
& \xbhat_{B, k} \in [\hat{\lb}_k, \hat{\ub}_k] \span  k \in \left\llbracket1, n-1\right\rrbracket, \\
& (\xbhat_{B, k}, \xb_{k}) \in \texttt{ReLU\_sol}(\xbhat_{B, k}, \xb_k,\hat{\lb}_k, \hat{\ub}_k) \span k \in \left\llbracket1, n-1\right\rrbracket.
\end{aligned}
\end{equation*}
In order for the above equation to be equal to \eqref{eq:dual-dj-modified}, noting that we can substitute $\xbhat_k=\xbhat_{B,k}$, we only need to prove the following for $ k \in \left\llbracket1, n-1\right\rrbracket$:
\begin{equation*}
	\hspace{-50pt}\min_{
		\begin{aligned} 
		\hspace{50pt}\xbhat_{k}  &\in [\hat{\lb}_k, \hat{\ub}_k] \\
		\hspace{50pt}(\xbhat_{k}, \xb_k) & \in \texttt{ReLU\_sol}(\xbhat_{k}, \xb_k,\hat{\lb}_k, \hat{\ub}_k)
		\end{aligned}
		} \hspace{-70pt}
	\left( \mub_k^T \xbhat_{k}\ -\ \mub_{k+1}^T W_{k+1} \xb_{k} \right) = \min_{		\begin{aligned} \xbhat_{k} \in [\hat{\lb}_k, \hat{\ub}_k] \end{aligned}} \left( \mub_k^T \xbhat_{k}\ - \mub_{k+1}^T W_{k+1} \sigma(\xbhat_{k}) \right),
\end{equation*}
which holds trivially for blocking or passing ReLUs. In the case of ambiguous ReLUs, instead, it suffices to point out that, due to the piecewise-linearity of the objective, the right hand side corresponds to: $\min \{\mub_k^T[i] \hat{\lb}_k[i], 0, \mub_k^T[i] \hat{\ub}_k[i] - (\mub_{k+1}^T W_{k+1})\hspace{-1pt}[i]\hat{\ub}_k[i] \}$. Looking at equation \eqref{eq:min-pk}, we can see that this minimization is hence identical to the to the right hand side, proving the second part of the theorem.
\section{Proof of proposition \ref{prop:init}}  \label{sec:dual-propagation}
We will show that, for ReLU activations, the solution generated by propagation-based methods 
(including amongst others, CROWN~\citep{Zhang2018}, and the algorithm by \citet{Wong2018}, see $\S$\ref{sec:propagation}),
can be used to provide initialization to both our dual~\eqref{eq:nonprox_problem} and dual~\eqref{eq:dj_prob-main} by \citet{Dvijotham2018}.
This holds in spite of the differences between the three dual derivations, which consist of an application of Lagrangian Decomposition, and of the Lagrangian relaxations of two different~problems. 

\vspace{10pt}
Recall that $\lsigmab_{k}(\xbhat_{k}) = \ubar{\mbf{a}}_k \xbhat_{k} + \ubar{\mbf{b}}_k$ and $\usigmab_{k}(\xbhat_{k}) = \bar{\mbf{a}}_k \xbhat_{k} + \bar{\mbf{b}}_k$ are two linear functions that bound $\sigma_k(\xbhat_{k})$ respectively from below and above.
Before proceeding with the proof, we point out that, relying on the definition of the convex hull of the ReLU in \eqref{eq:planet-relu-hull}, the following is true for any $\lsigmab_{k}(\xbhat_{k})$ and $\usigmab_{k}(\xbhat_{k})$ which are not unnecessarily loose (see Figure \ref{fig:relucvx}):
\begin{equation}
\begin{aligned}
0 \leq \bar{\mbf{a}}_k=\frac{\hat{\ub}_k}{\hat{\ub}_k - \hat{\lb}_k}\leq 1, \quad \bar{\mbf{b}}_k=\frac{- \hat{\lb}_k \odot \hat{\ub}_k}{\hat{\ub}_k - \hat{\lb}_k} \quad 0 \leq \ubar{\mbf{a}}_k \leq 1 & \qquad \text{if } \hat{\lb}_k \leq  0 \text{ and } \hat{\ub}_k \geq 0, \\[5pt]
\bar{\mbf{a}}_k=\ubar{\mbf{a}}_k=0 & \qquad  \text{if }  \hat{\ub}_k \leq 0, \\[5pt]
\bar{\mbf{a}}_k=\ubar{\mbf{a}}_k=1 & \qquad   \text{if } \hat{\lb}_k \geq 0, \\[5pt]
\ubar{\mbf{b}}_k=0 & \qquad   \text{in all cases}.
\end{aligned}
\label{eq:slopes-biases}
\end{equation}
We start by proving that, by taking as our solution of~\eqref{eq:nonprox_problem}:
\begin{equation}
\lambdabhat = \bar{\mub}, \qquad \text{where $\bar{\mub}$ is defined as per equation \eqref{eq:salman-assignment}},
\label{eq:kw_to_ours_dual}
\end{equation}
we obtain exactly the same bound given by plugging equation \eqref{eq:salman-assignment} into equation~\eqref{eq:bounds-propagation}.

Recall that, given a choice of $\lambdabhat$, the bound that we generate is:
\begin{equation*}
\begin{split}
\min_{\xb, \xbhat}\ & \hat{x}_{A, n}
+ \sum_{k=1}^{n-1} \lambdabhat_k^T \left( \xbhat_{B, k} - \xbhat_{A, k} \right)\\
\text{s.t. }\quad & \mathcal{P}_0(\xb_0, \xbhat_{A,1}), \\
&\mathcal{P}_k(\xb_k, \xbhat_{B, k}, \xbhat_{A, k+1}) \qquad k \in \left\llbracket1,n-1\right\rrbracket,
\end{split}
\end{equation*}
which, using the dummy variable $\lambdabhat_n = - \mbf{1}$ for ease of notation, can be decomposed into several subproblems:
\begin{equation}
\sum_{k=1}^{n-1}\left(
\min_{\begin{array}{c}\xb_k, \xbhat_{B, k}, \xbhat_{A, k+1} \\ \in \mathcal{P}_k\left( \xb_k, \xbhat_{B, k}, \xbhat_{A, k+1}\right)\end{array}}
\lambdabhat_k^T \xbhat_{B,k}\ -\ \lambdabhat_{k+1}^T \xbhat_{A, k+1}
\right)
+ \min_{\begin{array}{c}\xb_0, \xbhat_{A,1}\\\in \mathcal{P}_0\left( \xb_0, \xbhat_{A,1} \right)\end{array}} -\lambdabhat_1^T \xbhat_{A, 1}.
\label{eq:dual_decomp_for_kw}
\end{equation}

Let's start by replacing $\lambdabhat_1$ by $\bar{\mub}_1$, in accordance with~\eqref{eq:kw_to_ours_dual}. The problem over $\mathcal{P}_0$ becomes:
\begin{equation*}
\begin{split}
\min_{\xb_0, \xbhat_{A, 1}}\quad& - \bar{\mub}_1^T \xbhat_{A, 1}\\
\text{s.t.}\quad&  \xbhat_{1} = W_{1} \xb_{0} + \mathbf{b}_{1}\\
& \xb_{0} \in \mathcal{C}.
\end{split}
\end{equation*}
Substituting the equality into the objective of the subproblem, we get equation \eqref{eq:pb0_lb}:
\begin{equation*}
\min_{\xb_0 \in \mathcal{C}}\quad - \bar{\mub}_1^T W_{1} \xb_{0} - \bar{\mub}_1^T\mathbf{b}_{1}.
\end{equation*}
We will now evaluate the values of the problem over $\mathcal{P}_k$. Recall that these subproblems take the following form:
\begin{equation*}
\begin{aligned}
\hspace{-7pt}[\xbhat_{B, k}^*&, \xbhat_{A, k+1}^*] = \smash{\argmin_{\xbhat_{B, k}, \xbhat_{A, k+1}}} \quad
\lambdabhat_k^T \xbhat_{B, k} - \lambdabhat_{k+1}^T \xbhat_{A, k+1} \\[8pt]
\text{s.t } & \quad (\xb_k, \xbhat_{B, k}) \in \text{Conv}(\sigma_k, \hat{\lb}_k, \hat{\ub}_k),\\
& \quad (\xb_k, \xbhat_{B,k}) \in [\sigma_k(\hat{\lb}_k),\sigma_k(\hat{\ub}_k)]  \times  [\hat{\lb}_k, \hat{\ub}_k] ,\\
& \quad \xbhat_{A, k+1} = W_{k+1} \xb_{k} + \mathbf{b}_{k+1}.
\end{aligned}
\end{equation*}
Merging the last equality constraint into the objective function:
\begin{equation*}
\begin{split}
\min_{\xbhat_{B, k}, \xb_{k}}& \quad
\lambdabhat_{k}^T \xbhat_{B, k} - \lambdabhat_{k+1}^T W_{k+1} \xb_{k} - \lambdabhat_{k+1}^T \mathbf{b}_{k+1}\\
\text{s.t } & \quad (\xb_k, \xbhat_{B, k}) \in \text{Conv}(\sigma_k, \hat{\lb}_k, \hat{\ub}_k),\\
& \quad (\xb_k, \xbhat_{B,k}) \in [\sigma_k(\hat{\lb}_k),\sigma_k(\hat{\ub}_k)]  \times  [\hat{\lb}_k, \hat{\ub}_k].
\end{split}
\end{equation*}
We now distinguish the different cases of ReLU: passing ($\mathcal{I}^+$), blocking $\mathcal{I}^-$, and ambiguous ($\mathcal{I}$), recalling that the problem above decomposes over the ReLUs of layer $k$, the $i$-th ReLU being associated to the $i$-th entries of $\xb_k$ and $\xbhat_k$. 

\vspace{3pt}
If $i \in \mathcal{I}^+$, we have
$\text{Conv}(\sigma_k[i], \hat{\lb}_k[i], \hat{\ub}_k[i])\equiv \left\{(\xb_k[i], \xbhat_{B, k}[i]) | \xbhat_{B, k}[i] = \xb_k[i]\right\}$.
Then, the objective function term for $j$ becomes:
\begin{equation*}
\begin{split}
\min_{\xbhat_{B, k}[i]}& \quad
\left( \lambdabhat_{k}^T[i] - \lambdabhat_{k+1}^T[i] W_{k+1}\right)\xbhat_{B,k}[i] - \lambdabhat_{k+1}^T[i] \mathbf{b}_{k+1}[i]\\
\text{s.t } & \quad \xbhat_{B,k}[i] \in [\hat{\lb}_k[i], \hat{\ub}_k[i]].
\end{split}
\end{equation*}
Using equations \eqref{eq:salman-assignment} and \eqref{eq:kw_to_ours_dual}, and pointing out that $\bar{\mbf{a}}_k[i]=\ubar{\mbf{a}}_k[i]=1$ for $j \in \mathcal{I}^+$ due to equation \eqref{eq:slopes-biases}, the minimum of the subproblem above simplifies to: $- \bar{\mub}_{k+1}^T[i] \mathbf{b}_{k+1}[i]$.

\vspace{3pt}
If $i \in \mathcal{I}^-$, we have $\text{Conv}(\sigma_k[i], \hat{\lb}_k[i], \hat{\ub}_k[i])\equiv \left\{(\xb_k[i], \xbhat_{B, k}[i]) | \xb_k[i]=0 \right\}$.
Replacing $\xb_k[i]=0$:
\begin{equation*}
\begin{split}
\min_{\xbhat_{B, k}[i]}& \quad
\lambdabhat_{k}^T[i] \xbhat_{B,k}[i] - \lambdabhat_{k+1}^T[i] \mathbf{b}_{k+1}[i]\\
\text{s.t } & \quad \xbhat_{B,k}[i] \in [\hat{\lb}_k[i], \hat{\ub}_k[i]].
\end{split}
\end{equation*}
Using equations \eqref{eq:salman-assignment} and \eqref{eq:kw_to_ours_dual}, and pointing out that $\bar{\mbf{a}}_k[i]=\ubar{\mbf{a}}_k[i]=0$ for $j \in \mathcal{I}^-$ due to equation \eqref{eq:slopes-biases}, the minimum of the subproblem above simplifies again to: $- \bar{\mub}_{k+1}^T[i] \mathbf{b}_{k+1}[i]$.


\vspace{3pt}
The case for $i \in \mathcal{I}$ is more complex. We apply the same
strategy that led to equation~\eqref{eq:min-pk}, keeping all constant terms in the objective. 
This leads us to the following subproblem:
\begin{equation*}
\begin{split}
\argmin_{(\xbhat_{B,k}[i], \xb_{k}[i]) \in \{( \hat{\lb}_k[i], 0 ), \left( 0, 0 \right), \left( \hat{\ub}_k[i], \hat{\ub}_k[i] \right)\}} \enskip \bigg(\lambdabhat_k[i] \xbhat_{B, k}[i] - (\lambdabhat_{k+1}^T W_{k+1} )\hspace{-1pt}[i]\xb_{k}[i]\bigg) - \lambdabhat_{k+1}[i] \mathbf{b}_{k+1}[i],
\end{split}
\end{equation*}
which amounts to:
\begin{equation*}
\begin{aligned}
\min \left\{ \lambdabhat_k[i]\hat{\lb}_{k}[i], \left(\lambdabhat_k^T - \lambdabhat_{k+1}^T W_{k+1} \right) \hspace{-2pt}[i]\hat{\ub}_{k}[i], 0 \right\} - \lambdabhat_{k+1}[i] \mathbf{b}_{k+1}[i].
\end{aligned}
\end{equation*}
Using equation \eqref{eq:kw_to_ours_dual}, equation \eqref{eq:salman-assignment} and $\bar{\mub}_n = -1$, we obtain:
\begin{equation*}
\begin{aligned}
&\min \left\{ \begin{array}{l}
 \left( \bar{\mbf{a}}_k \odot [\bar{\lambdab}_k]_+ +  \ubar{\mbf{a}}_k \odot [\bar{\lambdab}_k]_- \right)[i]\hat{\lb}_{k}[i], \\
\left(\left( \bar{\mbf{a}}_k \odot [\bar{\lambdab}_k]_+ +  \ubar{\mbf{a}}_k \odot [\bar{\lambdab}_k]_- \right) - \bar{\lambdab}_k \right) \hspace{-2pt}[i]\hat{\ub}_{k}[i], \\
0,
\end{array} \right\} - \bar{\mub}_{k+1}^T[i] \mathbf{b}_{k+1}[i] \\
\overset{\text{using \eqref{eq:slopes-biases}}}{=}& \min 
\left\{ \begin{array}{l}
 -\bar{\mbf{b}}_k[i] [\bar{\lambdab}_k]_+[i] +  \hat{\lb}_{k}[i] \ubar{\mbf{a}}_k[i] [\bar{\lambdab}_k]_-[i] , \\
\left( (\bar{\mbf{a}}_k - 1) \odot [\bar{\lambdab}_k]_+ +  (\ubar{\mbf{a}}_k-1) \odot [\bar{\lambdab}_k]_- \right) \hspace{-2pt}[i]\hat{\ub}_{k}[i], \\
0,
\end{array} \right\} - \bar{\mub}_{k+1}^T[i] \mathbf{b}_{k+1}[i] 
\end{aligned}
\end{equation*}
Let us make the following observations: (i) due to equation \eqref{eq:slopes-biases} and $\hat{\ub}_k[i]>0$ (as $i \in \mathcal{I}$), the second argument of the minimum is always non-negative, (ii) $-\bar{\mbf{b}}_k[i] [\bar{\lambdab}_k]_+[i] \leq 0$, (iii)  $\hat{\lb}_{k}[i] \ubar{\mbf{a}}_k[i] [\bar{\lambdab}_k]_-[i] \geq 0$ due to $\hat{\lb}_{k}[i] \leq 0$ (as $i \in \mathcal{I}$).
Therefore, the minimum evaluates to $0$ if $\bar{\lambdab}_k \leq 0$, to $-\bar{\mbf{b}}_k[i] [\bar{\lambdab}_k]_+[i]$ otherwise. Hence, for $k \in \left\llbracket1,n-1\right\rrbracket$:
\begin{equation*}
\begin{aligned}
\min \left\{ \lambdabhat_k[i]\hat{\lb}_{k}[i], \left(\lambdabhat_k^T - \lambdabhat_{k+1}^T W_{k+1} \right) \hspace{-2pt}[i]\hat{\ub}_{k}[i], 0 \right\} \overset{\text{using \eqref{eq:kw_to_ours_dual}}}{\to} -\bar{\mbf{b}}_k[i] [\bar{\lambdab}_k]_+[i].
\end{aligned}
\end{equation*}
Plugging all the optimisation result into Equation~\eqref{eq:dual_decomp_for_kw}, we
obtain:
\begin{equation*}
\begin{aligned}
\min_{\xb_0 \in \mathcal{C}} 
\left(- \bar{\mub}_{1}^T W_0 \xb_{0}\right) - \sum_{k=1}^{n} \bar{\mub}_k^T\mbf{b}_k
- \sum_{k=1}^{n-1}  [\bar{\lambdab}_k]_+^T\bar{\mbf{b}}_k. 
\end{aligned}
\end{equation*}
which is exactly equation \eqref{eq:bounds-propagation}, recalling that $\ubar{\mbf{b}}_k=0$ from \eqref{eq:slopes-biases} and that we had set $\bar{\mub}_n = -1$.

\vspace{10pt}
Having proved that \eqref{eq:nonprox_problem} can be initialized with propagation-based methods, the result for dual \eqref{eq:dj_prob-main} then trivially follows from theorem \ref{theorem}. In fact, equation \eqref{eq:salman-assignment} satisfies the conditions (namely, $\bar{\lambdab}_{n-1} = -W_n^T \mbf{1}$ and $\bar{\lambdab}_{k-1} = W_k^T \bar{\mub}_k$ for $k \in \left\llbracket2,n-1\right\rrbracket$) under which the two duals yield the same bounds.

\vspace{30pt}
\section{Sigmoid Activation function}
\label{sec:sigmoid_act}
This section describes the computation highlighted in the paper in the context
where the activation function $\sig{x}$ is the sigmoid function:
\begin{equation}
  \sig{x} = \frac{1}{1+e^{-x}}
\end{equation}
A similar methodology to the one described in this section could be used to
adapt the method to work with other activation function such as hyperbolic
tangent.

We will start with a reminder about some properties of the sigmoid activation
function. It takes values between 0 and 1, with $\sig{0}=0.5$. We can easily
compute its derivatives:
\begin{equation}
  \begin{split}
    \sigma^\prime\left( x \right) &= \sig{x} \times (1 - \sig{x})\\
    \sigma^{\prime\prime}\left( x \right) &= \sig{x} \times (1 - \sig{x}) \times \left( 1 - 2 \sig{x} \right)\\
  \end{split}
\end{equation}

If we limit the domain of study to the negative inputs ($\left[ -\infty, 0
\right]$), then the function $x \mapsto \sig{x}$ is a convex function, as can
be seen by looking at the sign of the second derivative over that domain.
Similarly, if we limit the domain of study to the positive inputs ($\left[ 0,
  \infty \right]$), the function is concave.

\begin{figure}
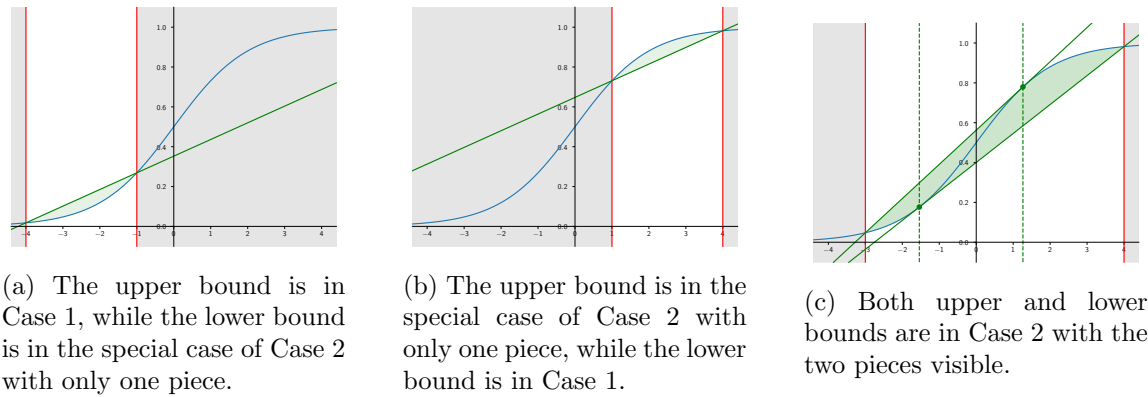

  \begin{subfigure}{.3\textwidth}
    \resizebox{\textwidth}{!}{\input{./sigmoid_fig/blocking.pgf}}
    \caption{\label{fig:blocked_sigmoid}The upper bound is in Case 1, while the
      lower bound is in the special case of Case 2 with only one piece.}
  \end{subfigure}%
  \hfill%
  \begin{subfigure}{.3\textwidth}
    \resizebox{\textwidth}{!}{\input{./sigmoid_fig/passing.pgf}}
    \caption{\label{fig:passing_sigmoid}The upper bound is in the special case
      of Case 2 with only one piece, while the lower bound is in Case 1.}
  \end{subfigure}%
  \hfill%
  \begin{subfigure}{.3\textwidth}
    \resizebox{\textwidth}{!}{\input{./sigmoid_fig/ambiguous.pgf}}
    \caption{\label{fig:amb_sigmoid}Both upper and lower bounds are in Case 2
      with the two pieces visible.}
  \end{subfigure}
  \caption{\label{fig:sigmoid_cvx_hull}Convex hull of the Sigmoid activation
    function for different input bound configurations.}
\vspace{20pt}
\end{figure}

\subsection{Convex hull computation}
In the context of ReLU, the convex hull of the activation function is given by
equation~\eqref{eq:planet-relu-hull}, as introduced by \citet{Ehlers2017}. We will
now derive it for sigmoid functions. Upper and lower bounds will be dealt in the
same way, so our description will only focus on how to obtain the concave upper
bound, limiting the activation function convex hull by above. The computation
to derive the convex lower bound is equivalent.

Depending on the range of inputs over which the convex hull is taken, the form
of the concave upper bound will change. We distinguish two cases.

\paragraph{Case 1: $\sigma^\prime\left(\hat{\ub}_k\right) \geq \ulslope$.}
We will prove that in this case, the upper bound will be the line passing through the points $(\hat{\lb}_k,
\sig{\hat{\lb}_k})$ and $(\hat{\ub}_k, \sig{\hat{\lb}_k})$. The equation of it is given by:
\begin{equation}
  \phiub(\xb) = \ulslope \left( \xb - \hat{\lb}_k \right) + \sig{\hat{\lb}_k}
  \label{eq:sig_lin_upper_bound}
\end{equation}

Consider the function $d(\xb) = \phiub(\xb) - \sig{\xb}$. To show that $\phiub$
is a valid upper bound, we need to prove that $\forall \xb \in \left[ \hat{\lb}_k,
  \hat{\ub}_k \right],\ d(\xb) \geq 0$. We know that $d(\hat{\lb}_k) = 0$ and $d(\hat{\ub}_k) =
0$, and that $d$ is a continuous function. Its derivative is given by:
\begin{equation}
  d^\prime(\xb) = \ulslope - \sig{\xb} (1 - \sig{\xb}).
\end{equation}
To find the roots of $d^\prime$, we can solve $d^\prime(\xb) = 0$ for the value
of $\sig{\xb}$ and then use the logit function to recover the value of $\xb$. In
that case, this is only a second order polynomial, so it can admit at most two
roots.

We know that $\lim_{\xb \to \infty} d^\prime(\xb) \geq 0$, and our hypothesis
tells us that $d^\prime(\hat{\ub}_k) \leq 0$. This means that at least one of the root
lies necessarily beyond $\hat{\ub}_k$ and therefore, the derivative of $d$ change
signs at most once on the $\left[ \hat{\lb}_k, \hat{\ub}_k \right]$ interval. If it never
changes sign, it is monotonous. Given that the values taken at both extreme
points are the same, $d$ being monotonous would imply that $d$ is constant,
which is impossible. We therefore conclude that this means that the derivative
change its sign exactly once on the interval, and is therefore unimodal. As we know
that $d^\prime(\hat{\ub}_k) \leq 0$, this indicates that $d$ is first increasing and
then decreasing. As both extreme points have a value of zero, this means that
$\forall \xb \in \left[ \hat{\lb}_k, \hat{\ub}_k \right], \ d(x) \geq 0$.

From this result, we deduce that $\phiub$ is a valid upper bound of $\sigma$. As a
linear function, it is concave by definition. Given that it constitutes the line
between two points on the curve, all of its points necessarily belong to the convex
hull. Therefore, it is not possible for a concave upper bound of the activation
function to have lower values. This means that $\phiub$ defines the upper
bounding part of the convex hull of the activation function.

\paragraph{Case 2: $\sigma^\prime \left( \hat{\ub}_k \right) \leq \ulslope$.}
In this case, we will have to decompose the upper bound into two parts, defined
as follows:
\begin{subnumcases}{\label{eq:sig_pw_upper_bound}\phiub(\xb)}
  \tlslope\left( \xb - \hat{\lb}_k \right) +
  \sig{\hat{\lb}_k}  & if $\xb \in \left[ \hat{\lb}_k, \tb_k \right]$\\
  \sig{\xb} & if $\xb \in \left[ \tb_k, \hat{\ub}_k \right]$\label{eq:ub_match_fun},
\end{subnumcases}
where $\tb_k$ is defined as the point such that $\sigma^\prime\left( \tb_k
\right) = \tlslope$ and $\tb_k > 0$. The value of $\tb_k$ can be computed by
solving the equation $\sig{\tb_k}(1 - \sig{\tb_k}) = \tlslope$, which can be
done using the Newton-Raphson method or a binary search. Note that this needs to
be done only when defining the problem, and not at every iteration of the
solver. In addition, the value of $\tb_k$ is dependant only on $\hat{\lb}_k$ so it's
possible to start by building a table of the results at the desired accuracy and
cache them.

Evaluating both pieces of the function of equation\eqref{eq:sig_pw_upper_bound}
at $\tb_k$ show that $\phiub$ is continuous. Both pieces are concave (for $\xb
\geq \tb_k \geq 0$, $\sigma$ is concave) and they share a supergradient (the
linear function of slope $\tlslope$) in $\tb_k$, so $\phiub$ is a concave
function. The proof we did for {\bf Case 1} can be duplicated to show that the
linear component is the best concave upper bound that can be achieved over the
interval $\left[ \hat{\lb}_k, \tb_k \right]$. On the interval $\left[ \tb_k, \hat{\ub}_k
\right]$, $\phiub$ is equal to the activation function, so it is also an upper
bound which can't be improved upon. Therefore, $\phiub$ is the upper bounding
part of the convex hull of the activation function.

Note that a special case of this happens when $\hat{\lb}_k \geq \tb_k$. In which case,
$\phiub$ consists of only equation~\eqref{eq:ub_match_fun}.

All cases are illustrated in Figure~\ref{fig:sigmoid_cvx_hull}. Case 1 is shown
in \ref{fig:blocked_sigmoid}, where the upper bound contains only the linear
upper bound. Case 2 with both segments is visible in
Figure\ref{fig:amb_sigmoid}, with the cutoff points $tb_k$ highlighted by a
green dot, and the special case with $\hat{\lb}_k \geq \tb_k$ is demonstrated in
Figure~\ref{fig:passing_sigmoid}.

\subsection{Solving the $\mathcal{P}_k$ subproblems over sigmoid activation}

As a reminder, the problem that needs to be solved is the following (where $\mbf{c}_{k+1} = -\lambdabhat_{k+1}$, $\mbf{c}_{k} = \lambdabhat_{k}$):
\begin{equation}
  \begin{split}
    [\xbhat_{B, k}, \xbhat_{A, k+1}] = \argmin_{\xbhat_{B, k}, \xbhat_{A, k+1}}& \quad
    \mbf{c}_k^T \xbhat_{B, k} + \mbf{c}_{k+1}^T \xbhat_{A, k+1}\\
    \text{s.t } & \quad \hat{\lb}_{k} \leq \xbhat_{B, k} \leq \hat{\ub}_{k}\\
    & \quad (\xb_k, \xbhat_{B, k}) \in \text{Conv}(\sigma, \hat{\lb}_k, \hat{\ub}_k) \\
    & \quad \xbhat_{A, k+1} = W_{k+1} \xb_{k} + \mathbf{b}_{k+1},
  \end{split}
  \label{eq:pk_sub_for_sig}
\end{equation}
where $\texttt{cvx\_hull}$ is defined either by
equations~\eqref{eq:sig_lin_upper_bound} or \eqref{eq:sig_pw_upper_bound}.
In this case, we will still be able to compute a closed form solution.

We will denote $\phiub$ and $\psiub$ the upper and lower bound functions
defining the convex hull of the sigmoid function, which can be obtained as
described in the previous subsection. If we use the last equality constraint of
Problem~\eqref{eq:pk_sub_for_sig} to replace the $\xbhat_{A, k+1}$ term in the objective
function, we obtain  \mbox{$\mbf{c}_k^T \xbhat_{B, k} + \mbf{c}_{k+1}^T  W_{k+1}
  \xb_{k}$}. Depending on the sign of $\mbf{c}_{k+1}^T  W_{k+1}$, $\xb_k$ will
either take the value $\phiub$ or $\psiub$, resulting in the following problem:

\begin{equation}
  \begin{split}
    \min_{\xbhat_{B,k}} \quad & \mbf{c}^T_{k} \xbhat_{B,k} + \left[ \mbf{c}^T_{k+1}W_{k+1} \right]_{-} \phiub\left(\xbhat_{B,k}\right) + \left[ \mbf{c}^T_{k+1}W_{k+1} \right]_{+} \psiub\left( \xbhat_{B, k}\right)\\
    \text{s.t } & \quad \hat{\lb}_{k} \leq \xbhat_{B, k} \leq \hat{\ub}_{k}.
    \label{eq:siglinopt}
  \end{split}
\end{equation}

To solve this problem, several observations can be made: First of all, at that
point, the optimisation decomposes completely over the components of
$\xbhat_{B,k}$ so all problems can be solved independently from each other.
The second observation is that to solve this problem, we can decompose the minimisation over the
whole range $\left[ \hat{\lb}_k, \hat{\ub}_k \right]$ into a set of minimisation over
the separate pieces, and then returning the minimum corresponding to the piece
producing the lowest value.

The minimisation over the pieces can be of two forms. Both bounds can be linear
(such as between the green dotted lines in Figure~\ref{fig:amb_sigmoid}), in
which case the problem is easy to solve by looking at the signs of the
coefficient of the objective function. The other option is that one of the
bound will be equal to the activation function (such as in
Figures~\ref{fig:blocked_sigmoid} or \ref{fig:passing_sigmoid}, or in the outer
sections of Figure~\ref{fig:amb_sigmoid}), leaving us with a problem of the
form:
\begin{equation}
  \begin{split}
    \min_{\xbhat_{B, k}} \quad & \mbf{c}_\text{lin}^T \xbhat_{B, k} + \mbf{c}_{\sigma}^T \sig{\xbhat_{B,k}}\\
    & \hat{\lb} \leq \xbhat_{B,k} \leq \hat{\ub},
  \end{split}
  \label{eq:with_sig_subp}
\end{equation}
where $\hat{\lb}$, $\hat{\ub}$, $\mbf{c}_\text{lin}$ and $\mbf{c}_\sigma$ will depend on
what part of the problem we are trying to solve.

This is a convex problem so the value will be reached either at the extremum of
the considered domain ($\hat{\lb}$ or $\hat{\ub}$), or it will be reached at the points
where the derivative of the objective functions cancels. This corresponds to
the roots of $\mbf{c}_\text{lin} + \mbf{c}_{\sigma}^T \sig{\xbhat_{B,k}}(1 - \sig{\xbhat_{B,k}})$.
Provided that $1 + \frac{4 \mbf{c}_\text{lin}}{\mbf{c}_\sigma} \geq 0$, the
possible roots will be given by \mbox{$\sigma^{-1}\left( \frac{1 \pm \sqrt{1 + \frac{4
        \mbf{c}_\text{lin}}{\mbf{c}_\sigma}}}{2} \right)$}, with $\sigma^{-1}$
being the logit function, the inverse of the sigmoid function $\left(
  \sigma^{-1}(\xb) = \log\left( \frac{\xb}{1-\xb} \right) \right)$. To solve
problem~\eqref{eq:with_sig_subp}, we evaluate its objective function at the
extreme points ($\hat{\lb}$ and $\hat{\ub}$) and at those roots if they are in the feasible
domain, and return the point corresponding to the minimum score achieved. With
this method, we can solve the $\mathcal{P}_k$ subproblems even when the
activation function is a sigmoid.

\section{Momentum for the Proximal solver} \label{sec:prox-appendix}

Supergradient methods for both our dual \eqref{eq:nonprox_problem} and for \eqref{eq:dj_prob-main} by \citet{Dvijotham2018} can easily rely on acceleration techniques such as Adam~\citep{Kingma2015} to speed-up convergence. 
Therefore, inspired by its presence in Adam, and by the work on accelerating proximal methods \citep{Lin2017,Salzo12}, we apply momentum on the proximal updates.

By closely looking at equation \eqref{eq:lambda_updates}, we can draw a similarity between dual updates in the proximal algorithm and supergradient ascent. The difference is that the former operation takes place after a closed-form minimization of the linear inner problem in $\xbhat_A$ and $\xbhat_B$, whereas the latter after some steps of an optimization algorithm that solves the quadratic form of the Augmented Lagrangian \eqref{eq:z_updates}. Let us denote the (approximate) argmin of the Augmented Lagrangian at the $t$-th iteration of the method of multipliers by $\xbhat^{t, \dagger}$.
Thanks to the aforementioned similarity, we can keep an exponential average of the gradient-like terms with parameter $\mu \in [0,1]$ and adopt momentum for the dual updates,~yielding:
\begin{equation}
\begin{split}
\pib_k^{t+1} &= \mu \pib_k^{t} + \frac{\xbhat^{t, \dagger}_{B,k} - \xbhat^{t, \dagger}_{A,k}}{\eta_k} \\
\lambdabhat_k^{t+1} &= \lambdabhat_k^{t} + \pib_k^{t+1} 
\end{split}
\label{eq:momentum_updates}
\end{equation}
Under the proximal interpretation of the method of multipliers, the Augmented Lagrangian can be derived by adding a (quadratic) proximal term on $\lambdabhat$ in the standard Lagrangian, and plugging the closed-form solution of the resulting quadratic problem, which corresponds to equation \eqref{eq:lambda_updates}, into the objective~\citep{Bertsekas1989}.
In other words, we make the following modification to the standard Lagrangian:
\begin{equation*}
	\hat{x}_{A, n} + \sum_{k=1}^{n-1} \left( \xbhat_{B, k} - \xbhat_{A, k} \right)^T\lambdabhat_k^{t+1} \quad \rightarrow  \quad\hat{x}_{A, n} + \sum_{k=1}^{n-1} \left( \xbhat_{B, k} - \xbhat_{A, k} \right)^T \lambdabhat_k^{t+1}  - \frac{\eta_k}{2}  \| \lambdabhat_k^{t+1}-  \lambdabhat_k^{t} \|^2,
\end{equation*}
and then plug in equation \eqref{eq:lambda_updates}, which results from setting the dual gradient of the quadratic function above to $0$.
If, rather than equation \eqref{eq:lambda_updates}, we plug in equation \eqref{eq:momentum_updates}, incorporating momentum, the primal problem of the modified method of multipliers becomes:
\begin{gather}
\begin{split}
\left[ \xb^{t}, \xbhat^{t} \right] \ = \smash{\argmin_{\xb, \xbhat}}\, \mathcal{L}\left( \xbhat, \lambdabhat^t\right)&
\quad \text{s.t. } \mathcal{P}_0(\xb_0, \xbhat_{A,1}); \quad \mathcal{P}_k(\xb_k, \xbhat_{B, k}, \xbhat_{A, k+1})\ \quad k \in \left\llbracket1,n-1\right\rrbracket\\
\text{where: } \quad \mathcal{L}\left( \xbhat, \lambdabhat^t\right)=\  &\left[\begin{array}{l}
\hat{x}_{A, n} + \sum_{k=1}^{n-1} \left( \xbhat_{B, k} - \xbhat_{A, k} \right)^T \lambdabhat_k^t + \sum_{k=1}^{n-1} \frac{1}{2\eta_k}  \|\xbhat_{B,k} - \xbhat_{A,k} \|^2\\
 - \sum_{k=1}^{n-1} \quad \frac{\eta_k}{2}  \| \mu \pib_k \|^2. \end{array}\right.
\end{split}
\raisetag{10pt}
\label{eq:new_ag}
\end{gather}
We point out that the Augmented Lagrangian's gradients \mbox{$\nabla_{\xbhat_{B,k}} \mathcal{L}\left( \xbhat,\lambdabhat\right)$} and \mbox{$\nabla_{\xbhat_{A,k}} \mathcal{L}\left( \xbhat,\lambdabhat\right)$} remain unvaried with respect to the momentum-less solver. Therefore, in practice, the only algorithmic change for the solver lies in the dual update formula \eqref{eq:momentum_updates}. 

\newpage
\begin{figure*}[b!]
	\centering
	\begin{subfigure}{\textwidth}
		\includegraphics[width=1\textwidth]{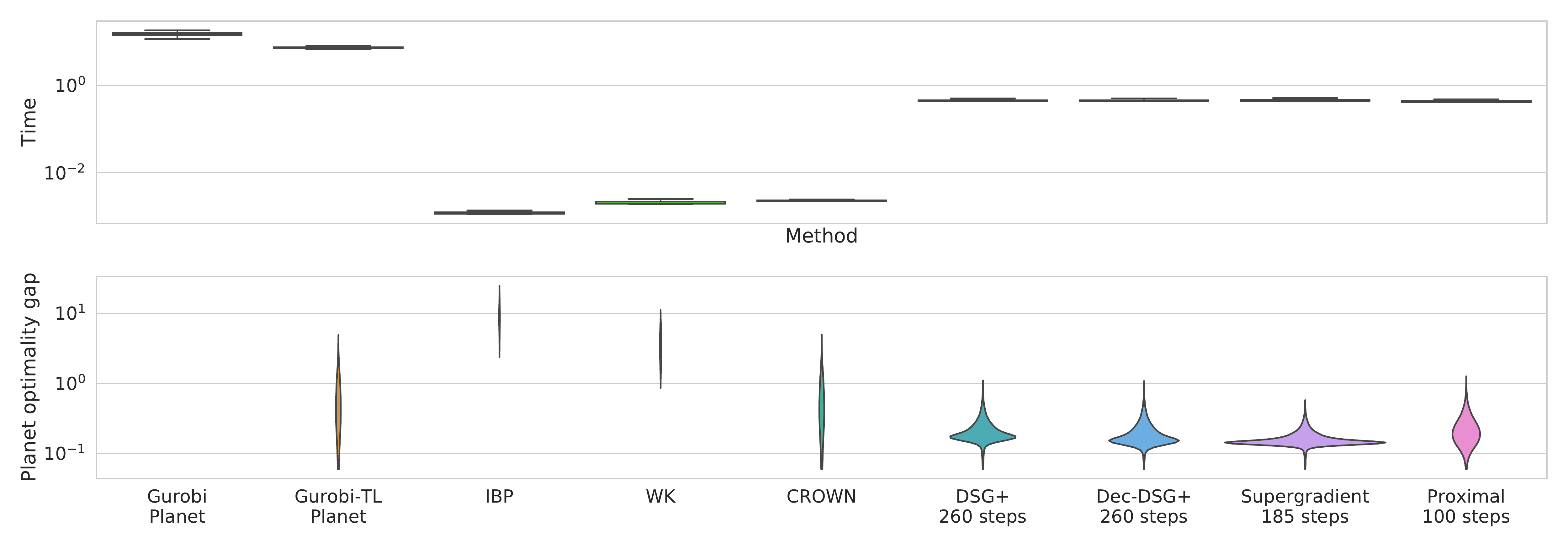}
		\vspace{-15pt}
	\end{subfigure}
	\begin{subfigure}{\textwidth}
		\includegraphics[width=1\textwidth]{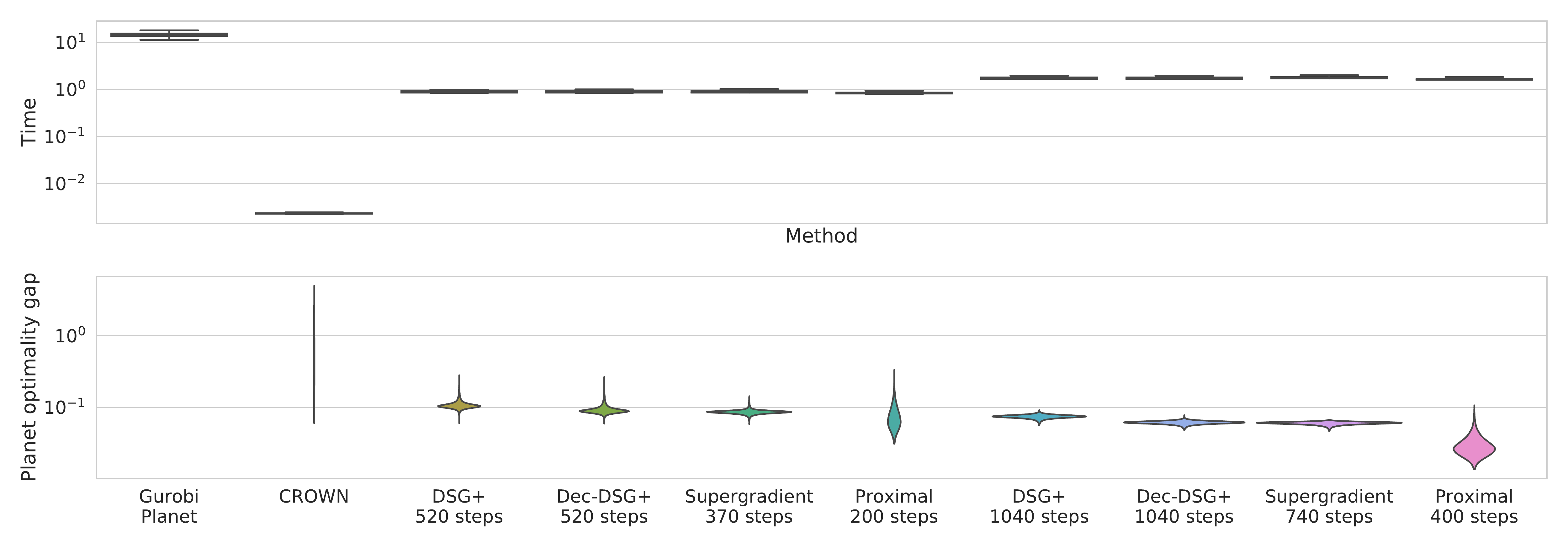}
	\end{subfigure}
	\caption{\label{fig:sgd8} Comparison of the distribution of runtime
		and gap to optimality on an SGD-trained network. In both
		cases, lower is better. The width at a given value represents the
		proportion of problems for which this is the result. Gurobi Planet always returns the optimal solution to problem \eqref{eq:planet-relax}, at a large computational cost.}
\end{figure*}            

\section{Supplementary Incomplete Verification Experiments} \label{sec:experiments-incomplete-appendix}
\begin{figure*}[t!]
	\centering
	\begin{minipage}{.48\textwidth}
		\centering
		\includegraphics[width=\textwidth]{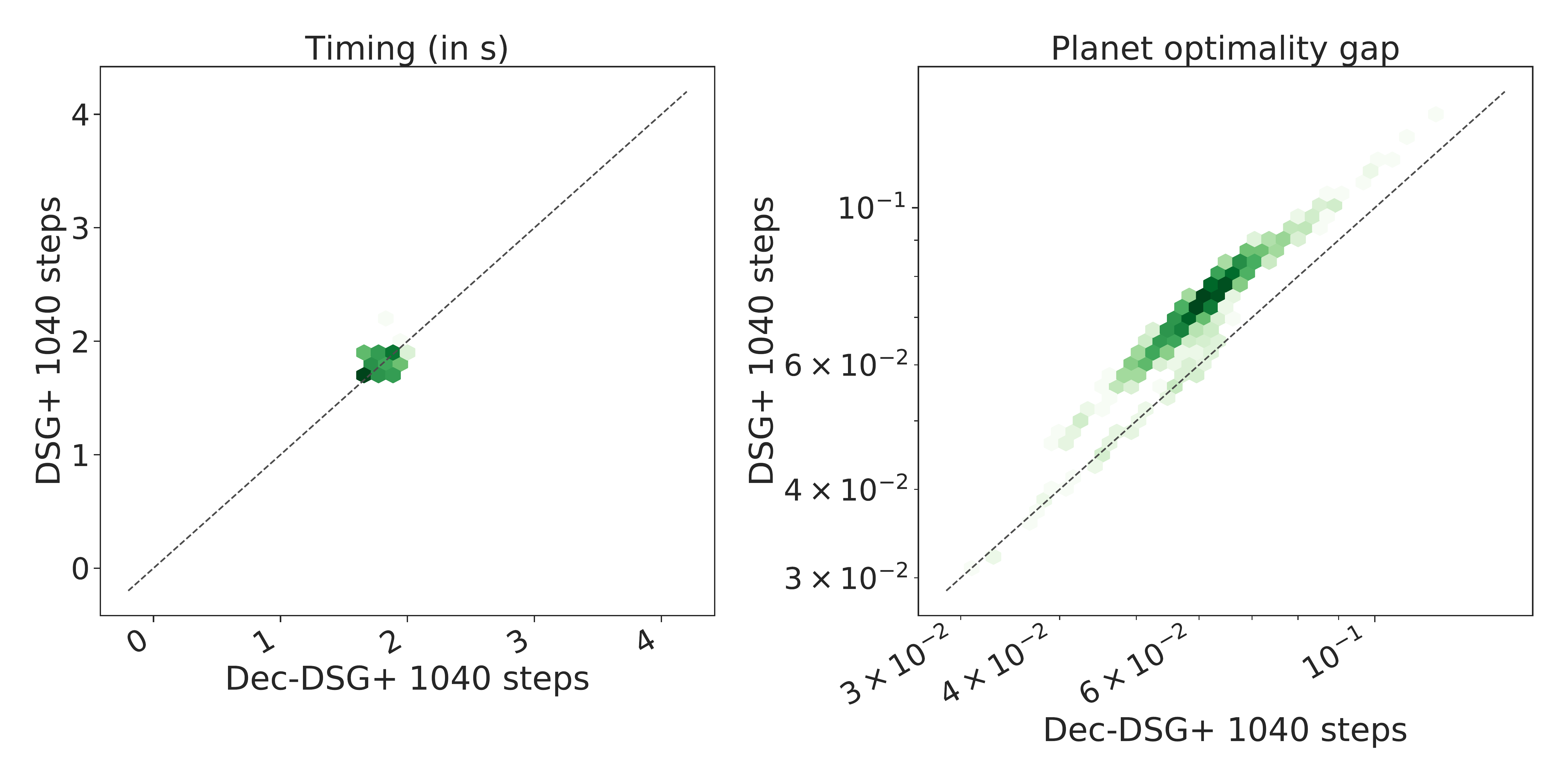}
	\end{minipage}
	\hspace{5pt}
	\begin{minipage}{.48\textwidth}
		\centering
		\includegraphics[width=\textwidth]{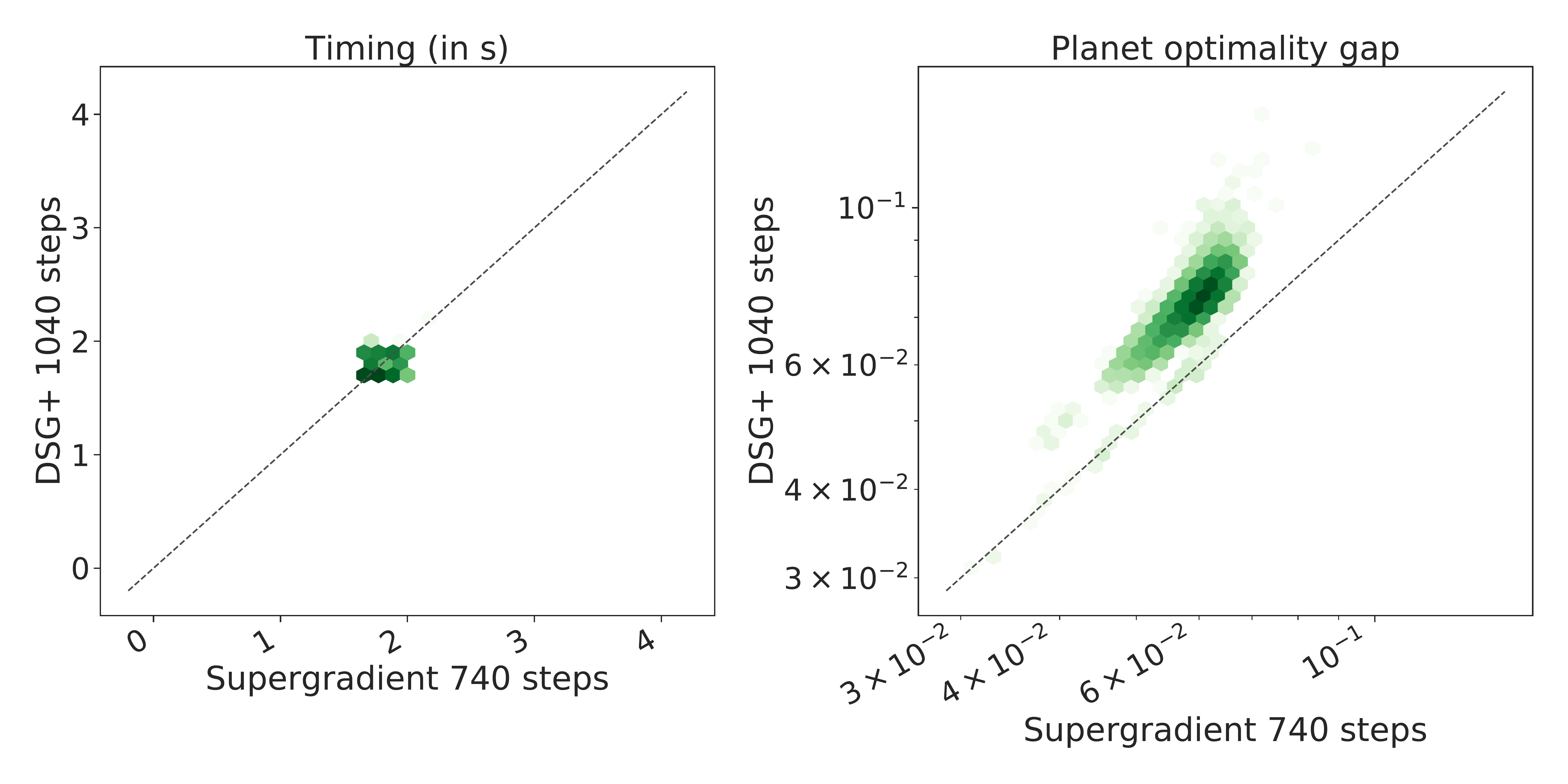}
	\end{minipage}
	\vspace{5pt}
	\begin{minipage}{.48\textwidth}
		\centering
		\includegraphics[width=\textwidth]{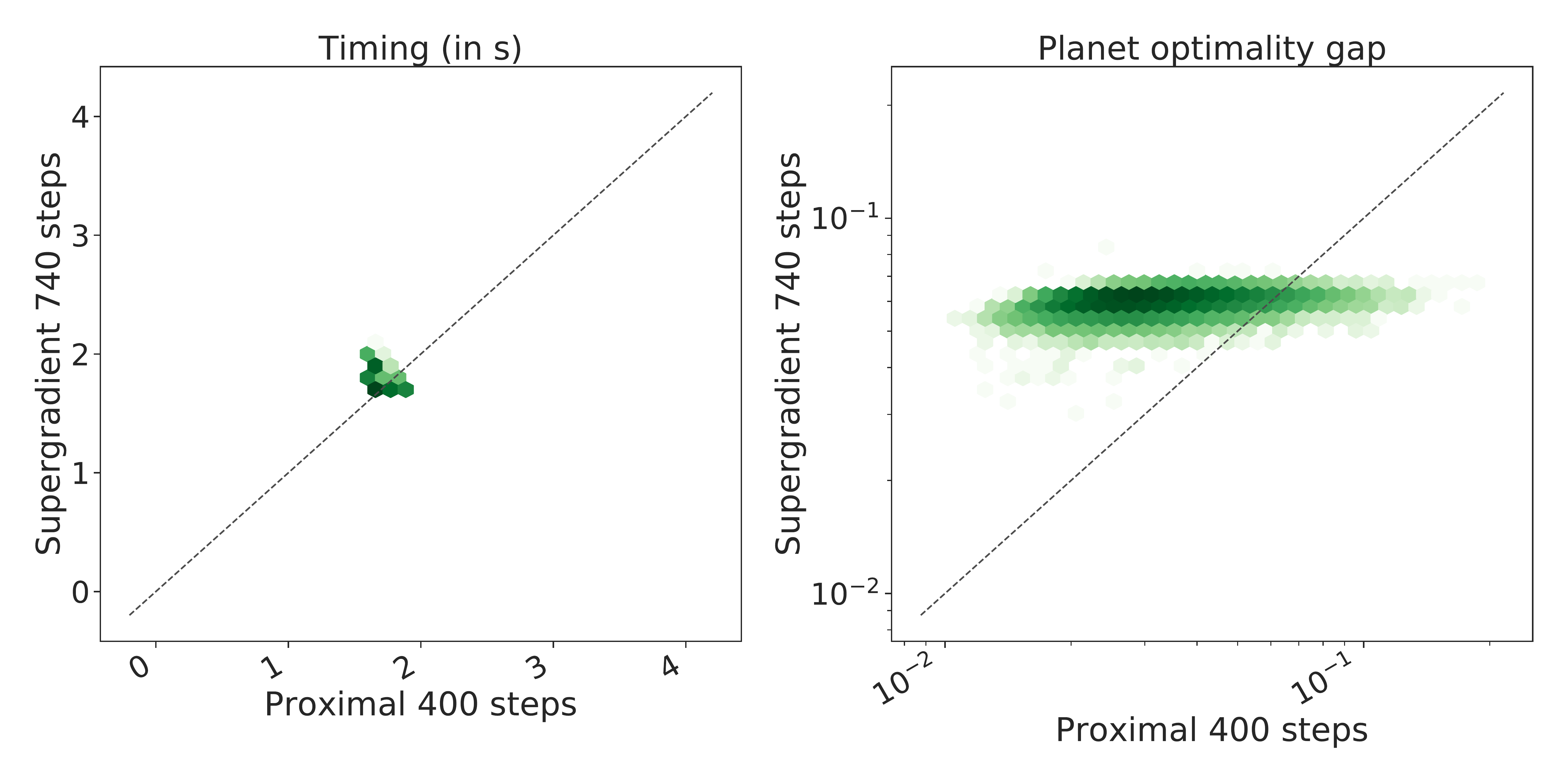}
	\end{minipage}
	\hspace{5pt}
	\begin{minipage}{.48\textwidth}
		\centering
		\includegraphics[width=\textwidth]{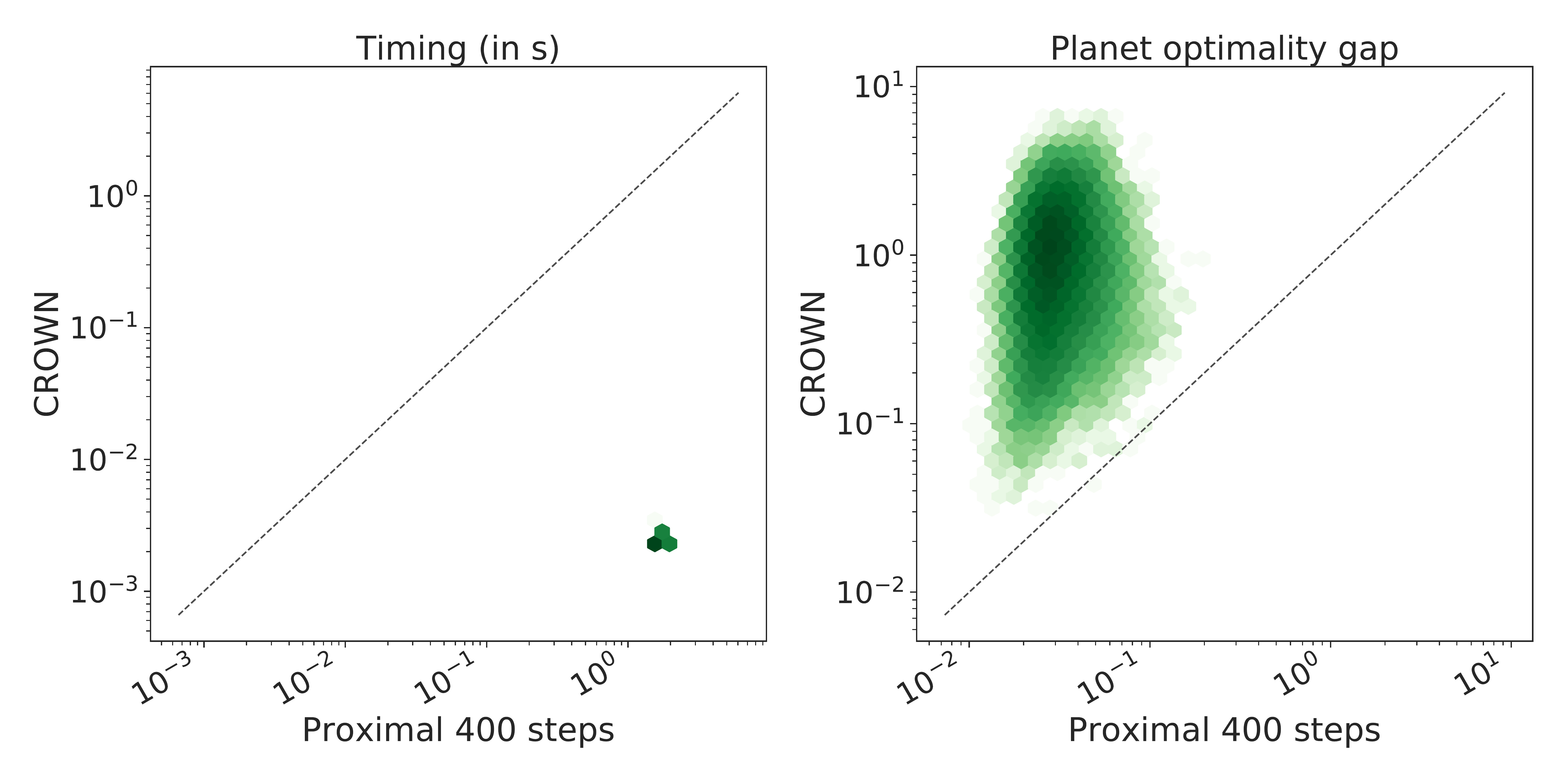}
	\end{minipage}                                                                                                 
	\vspace{-15pt}                                                                                                 
	\caption{Pointwise comparison for a subset of the methods on the data presented in Figure \ref{fig:sgd8}. Each datapoint corresponds to a CIFAR image, darker colour shades mean higher point density in a logarithmic scale. The dotted line corresponds to the equality and in both graphs, lower is better along both axes.}
	\label{fig:sgd-pointwise}
\end{figure*}

We now complement the incomplete verification results presented in the main paper.
Figures \ref{fig:sgd8} and \ref{fig:sgd-pointwise} show experiments (see section \ref{sec:experiments-incomplete}) for a network trained using standard stochastic gradient descent and cross entropy, with no robustness objective. We employ $\epsilon_{\text{verif}} = 5/255$, which is smaller than commonly employed verification radii due to the network's adversary-agnostic training. Most observations done in the main paper for the adversarially trained network (Figures \ref{fig:madry8} and \ref{fig:madry8-pointwise}) still hold true: in particular, the advantage of the Lagrangian Decomposition based dual compared to the dual by \citet{Dvijotham2018} is even more marked than in Figure \ref{fig:madry8-pointwise}. In fact, for a subset of the properties, DSG+ returns rather loose bounds even after $1040$ iterations. 
The Proximal returns better bounds than the Supergradient in this case as well, with Proximal displaying a much larger support of the optimality gap distribution.

\clearpage

\bibliography{standardstrings,oval,explp}

\end{document}